\documentclass{article}

\usepackage{microtype}
\usepackage{graphicx}
\usepackage{subfigure}
\usepackage{booktabs}
\usepackage{rotating}
\usepackage{hyperref}

\usepackage[accepted]{icml2024}

\usepackage{amsmath}
\usepackage{amssymb}
\usepackage{mathtools}
\usepackage{amsthm}
\usepackage{makecell}
\usepackage{cancel}
\usepackage{dsfont}

\usepackage[capitalize,noabbrev]{cleveref}
\usepackage{autonum}
\usepackage{listings}
\usepackage{enumitem}
\usepackage[most]{tcolorbox}
\usepackage{bm}
\usepackage{xspace}
\usepackage{nccmath}
\usepackage{arydshln}

\usepackage{etoc}

\allowdisplaybreaks

\newtcolorbox{emphasis}[1][]{%
  colback=gray!25, colframe=gray!25,
  coltitle=black,
  sharp corners,
  left=1pt,
  right=1pt,
  top=1pt,
  bottom=1pt,
  title=#1
}

\theoremstyle{plain}
\newtheorem{theorem}{Theorem}[section]
\newtheorem{proposition}[theorem]{Proposition}

\theoremstyle{definition}

\theoremstyle{remark}

\newcommand{\x}{x}
\newcommand{\pdata}{p_{\mathrm{data}}}

\newcommand{\pnoise}{p_{\mathrm{noise}}}
\newcommand{\E}{\mathbb{E}}

\newcommand{\relu}{\mathrm{ReLU}}
\newcommand{\oneDd}{1:D \backslash d}
\newcommand{\relurate}{R^*}
\newcommand{\dbrate}{R^{\mathrm{DB}}}

\newcommand{\dd}{\mathrm{d}}

\newcommand{\norm}[1]{\left \lvert \left\lvert #1 \right\rvert \right\rvert}

\newcommand{\dt}{\mathrm{d}t}

\newcommand{\allframes}{\mathbf{T}}
\newcommand{\tx}{\tilde{\x}}
\newcommand{\statespace}{S}
\newcommand{\SO}{\mathrm{SO}(3)}
\newcommand{\SE}{\mathrm{SE}(3)}
\newcommand{\Dt}{\Delta t}

\newcommand{\kdelta}[2]{\delta \left\{ #1, #2 \right\} }

\newcommand{\ptdt}{p_{t+\mathrm{d} t | t}}

\newcommand{\noisemarg}{p_{t|1}}
\newcommand{\denoise}{p_{1|t}}

\newcommand{\noise}{\eta}

\Crefname{equation}{Eq.}{Eqs.}
\Crefname{figure}{Fig.}{Figs.}
\Crefname{table}{Table.}{Tables.}
\Crefname{section}{Sec.}{Secs.}
\Crefname{appendix}{App.}{Apps.}
\Crefname{algorithm}{Alg.}{Algs.}
\Crefname{proposition}{Prop.}{Props.}

\newcommand{\unif}{\mathcal{U}}
\newcommand{\vf}{\nu}
\newcommand{\protmodel}{Multiflow\xspace}
\newcommand{\method}{DFM\xspace}
\newcommand{\trans}{x}
\newcommand{\probflow}{p_t}
\newcommand{\rots}{r}
\newcommand{\amino}{a}
\newcommand{\residue}{T}
\newcommand{\allresidue}{\mathbf{T}}

\newcommand{\real}{\mathbb{R}}

\newenvironment{talign}
 {\let\displaystyle\textstyle\align}
 {\endalign}

 \lstdefinestyle{pythonstyle}{
  language=Python,
  basicstyle=\small\ttfamily,
  commentstyle=\color{green!40!black},
  keywordstyle=\color{blue},
  numberstyle=\tiny\color{gray},
  numbers=left,
  frame=single,
  breaklines=true,
  breakatwhitespace=true,
  tabsize=4
}

\newcommand{\appendixhead}{
  \centerline{\textbf{\LARGE Appendix to:}\vspace{0.15in}}
  \centerline{\textbf{\LARGE Generative Flows on Discrete State-Spaces:}\vspace{0.1in}}
  \centerline{\textbf{\LARGE Enabling Multimodal Flows with Applications to Protein Co-Design}\vspace{0.1in}}
}

\usepackage[textsize=tiny]{todonotes}

\icmltitlerunning{Discrete Flow Models}

\usepackage{xpatch}
\xpatchcmd{\NCC@ignorepar}{%
\abovedisplayskip\abovedisplayshortskip}
{%
\abovedisplayskip\abovedisplayshortskip%
\belowdisplayskip\belowdisplayshortskip}
{}{}

\begin{document}

\twocolumn[
\icmltitle{Generative Flows on Discrete State-Spaces: \\ Enabling Multimodal Flows with Applications to Protein Co-Design}

\icmlsetsymbol{equal}{*}

\begin{icmlauthorlist}
\icmlauthor{Andrew Campbell}{equal,ox}
\icmlauthor{Jason Yim}{equal,mit}
\icmlauthor{Regina Barzilay}{mit}
\icmlauthor{Tom Rainforth}{ox}
\icmlauthor{Tommi Jaakkola}{mit}
\end{icmlauthorlist}

\icmlaffiliation{ox}{Department of Statistics, University of Oxford, UK}
\icmlaffiliation{mit}{Department of Electrical Engineering and Computer Science, Massachusetts Institute of Technology, Massachusetts, USA}

\icmlcorrespondingauthor{Andrew Campbell}{campbell@stats.ox.ac.uk}
\icmlcorrespondingauthor{Jason Yim}{jyim@csail.mit.edu}

\icmlkeywords{Discrete Flow Models}

\vskip 0.3in
]

\printAffiliationsAndNotice{\icmlEqualContribution} 

\begin{abstract}
Combining discrete and continuous data is an important capability for generative models.
We present Discrete Flow Models (DFMs), a new flow-based model of discrete data that provides the missing link in enabling flow-based generative models to be applied to multimodal continuous and discrete data problems.
Our key insight is that the discrete equivalent of continuous space flow matching can be realized using Continuous Time Markov Chains.
DFMs benefit from a simple derivation that includes discrete diffusion models as a specific instance while allowing improved performance over existing diffusion-based approaches.
We utilize our DFMs method to build a multimodal flow-based modeling framework.
We apply this capability to the task of protein co-design, wherein we learn a model for jointly generating protein structure and sequence.
Our approach achieves state-of-the-art co-design performance while allowing the same multimodal model to be used for flexible generation of the sequence or structure.

\end{abstract}

\section{Introduction}
\label{sec:introduction}
Expanding the capabilities of generative models to handle discrete and continuous data, which we refer to as \emph{multimodal}, is a fundamental problem to enable their widespread adoption in scientific applications \citep{wang2023scientific}.
One such application requiring a multimodal generative model is protein co-design where the aim is to jointly generate continuous protein structures alongside corresponding discrete amino acid sequences \citep{shi2022protein}.
Proteins have been well-studied: the function of the protein is endowed through its structure while the sequence is the blueprint of how the structure is made.
This interplay motivates jointly generating the structure and sequence rather than in isolation.
To this end, the focus of our work is to develop a multimodal generative framework capable of co-design.

Diffusion models \citep{sohl2015deep, ho2020denoising, song2020score} have achieved state-of-the-art performance across multiple applications.
They have potential as a multimodal framework because they can be defined on both continuous and discrete spaces \citep{hoogeboom2021argmax, austin2021structured}.
However, their sample time inflexibility makes them unsuitable for multimodal problems.
On even just a single modality, finding optimal sampling parameters requires extensive re-training and evaluations \citep{karras2022elucidating}.
This problem is exacerbated for multiple modalities.
On the other hand, flow-based models \citep{liu2022flow, albergo2022building, lipman2022flow} improve over diffusion models with a simpler framework that allows for superior performance through sampling flexibility \citep{ma2024sit}.
Unfortunately, our current inability to define a flow-based model on discrete spaces holds us back from a multimodal flow model.

\begin{figure*}[t]
    \vspace{-0.1cm}
    \centering
    \includegraphics[width=0.95\textwidth]{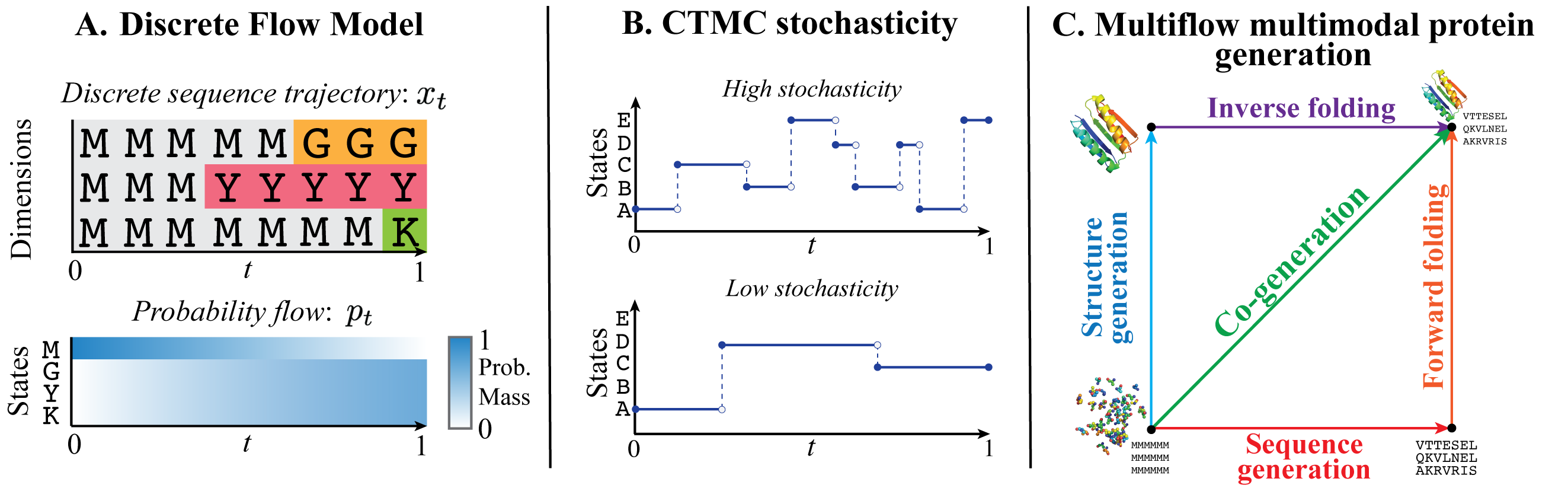}
    \vspace{-0.3cm}
    \caption{
    \textbf{Overview.}
    \textbf{(A.)} A \method trajectory with masking over a 3-dim. sequence with 4 possible states.
    \textbf{(B.)} CTMC stochasticity controls the number of transitions in a sequence trajectory \emph{while  respecting the flow $p_t$}. Shown is a 1-dim. sequence with 5 states.
    \textbf{(C.)} Sampling with \protmodel can start from noise (bottom left) or with either the structure or sequence given (top left and bottom right).
    Any sampling tasks (structure/sequence generation, forward/inverse folding, co-generation) can be achieved with a single \protmodel model.
    }
    \label{fig:overview}
    \vspace{-0.3cm}
\end{figure*}

We address this by introducing a novel flow-based model for discrete data named \textbf{Discrete Flow Models (DFMs)} and thereby unlock a complete framework for flow-based multimodal generative modeling.
Our key insight comes from seeing that a discrete flow-based model can be realized using Continuous Time Markov Chains (CTMCs).
DFMs are a new discrete generative modeling paradigm: less restrictive than diffusion, allows for sampling flexibility without re-training and enables simple combination with continuous state space flows to form multimodal flow models.

\cref{fig:overview}A provides an overview of DFMs.
We first define a probability flow $p_t$ that linearly interpolates from noise to data.
We then generate new data by simulating a sequence trajectory $\x_t$ that follows $p_t$ across time which requires training a denoising neural network with cross-entropy.
The sequence trajectory could have many transitions or few, a property we term CTMC Stochasticity (\cref{fig:overview}B).
Prior discrete diffusion models are equivalent to picking a specific stochasticity at training time, whereas we can adjust it at inference: enhancing sample quality and exerting control over sample distributional properties.

Using DFMs, we are then able to create a multimodal flow model by defining factorized flows for each data modality.
We apply this capability to the task of protein co-design by developing a novel continuous structure and discrete sequence generative model named \textbf{\protmodel}.
We combine a DFM for sequence generation and a flow-based structure generation method developed in \citet{yim2023fast}.
Previous multimodal approaches either generated only the sequence or only the structure and then used a prediction model to infer the remaining modality (see \cref{sec:related}).
Our \emph{single} model can jointly generate sequence and structure while being able to condition on either modality.

In our experiments (\cref{sec:experiments}), we first verify on small scale text data that DFMs outperform the discrete diffusion alternative, D3PM \citep{austin2021structured} through their expanded sample time flexibility.
We then move to our main focus, assessing \protmodel's performance on the co-design task of jointly generating protein structure and sequence.
\protmodel achieves state-of-the-art co-design performance while data distillation allows for obtaining state-of-the-art structure generation.
We find CTMC stochasticity enables controlling sample properties such as secondary structure composition and diversity.
Preliminary results on inverse and forward folding show \protmodel is a promising path towards a general-purpose protein generative model.

Our contributions are summarized as follows:
\begin{itemize}[
    leftmargin=4mm,
    partopsep=0pt,
]
\vspace{-0.35cm}
    \item We present Discrete Flow Models (DFMs), a novel discrete generative modeling method built through a CTMC simulating a probability flow.
    \vspace{-0.2cm} 
    \item We combine DFMs with continuous flow-based methods to create a multimodal generative modeling framework.
    \vspace{-0.2cm}
    \item We use our multimodal framework to develop \protmodel, a state-of-the-art generative protein co-design model with the flexibility of multimodal protein generation.
    \vspace{-0.3cm}
\end{itemize}

\section{Background}
\label{sec:background}
We aim to model discrete data where a sequence $x \in \{1,\dots,\statespace\}^D$ has $D$ dimensions, each taking on one of $\statespace$ states.
For ease of exposition, we will assume $D=1$; all results hold for $D>1$ as discussed in \cref{sec:factorization}.
We first explain a class of continuous time discrete stochastic processes called Continuous Time Markov Chains (CTMCs) \cite{norris1998markov} and then describe the link to probability flows.
\subsection{Continuous Time Markov Chains.}
A sequence trajectory $x_t$ over time $t \in [0, 1]$ that follows a CTMC alternates between resting in its current state and periodically jumping to another randomly chosen state. We show example trajectories in \cref{fig:overview}B.
The frequency and destination of the jumps are determined by the rate matrix $R_t \in \mathbb{R}^{\statespace \times \statespace}$ with the constraint its off-diagonal elements are non-negative.
The probability $\x_t$ will jump to a different state $j$ is $R_t(\x_t, j) \dt$ for the next infinitesimal time step $\dt$ .
We can write the transition probability as
\begin{align}
    \label{eq:transition_prob}
    \ptdt (j | \x_t) &= \begin{cases}
        R_t(\x_t, j) \dt & \text{for} \,\, j \neq \x_t \\
        1 + R_t(\x_t, \x_t)\dt & \text{for} \,\, j=\x_t
        \end{cases} \\
        \label{eq:shorthand_marginal}
        &= \kdelta{\x_t}{j}+ R_t(\x_t, j) \dt
\end{align}
where $\kdelta{i}{j}$ is the Kronecker delta which is $1$ when $i=j$ and is otherwise $0$ and $R_t(\x_t, \x_t) \vcentcolon= - \sum_{k \neq x} R_t(\x_t, k)$ in order for $\ptdt(\cdot | i)$ to sum to $1$.
We use compact notation \cref{eq:shorthand_marginal} in place of \cref{eq:transition_prob}.
Therefore, $\ptdt$ is a Categorical distribution with probabilities $\kdelta{\x_t}{\cdot}+ R_t(\x_t, \cdot) \dd t$ that we denote as $\mathrm{Cat}(\kdelta{\x_t}{j}+ R_t(\x_t, j) \dd t)$:
\begin{equation}
    j \sim \ptdt(j | \x_t) \ \Longleftrightarrow \ j \sim \mathrm{Cat}(\kdelta{\x_t}{j}+ R_t(\x_t, j) \dd t).
\end{equation}
In practice, we need to simulate the sequence trajectory with finite time intervals $\Dt$.
A sequence trajectory can be simulated with Euler steps \citep{sun2022score}
\begin{equation}
    \label{eq:sample_euler_step}
    \x_{t+ \Dt} \sim \mathrm{Cat}(\kdelta{\x_t}{\x_{t+ \Dt}} + R_t(\x_t, \x_{t+ \Dt})  \Dt),
\end{equation}
where the sequence starts from an initial sample $\x_0 \sim p_0$ at time $t=0$.
The rate matrix $R_t$ along with an initial distribution $p_0$ together define the CTMC.

\subsection{Kolmogorov equation}
For a sequence trajectory following the dynamics of a CTMC, we write its marginal distribution at time $t$ as $p_t(\x_t)$.
The Kolmogorov equation allows us to relate the rate matrix $R_t$ to the change in $p_t(\x_t)$. It has the form:
\begin{equation}
    \hspace*{-0.2cm} \partial_t p_t(\x_t) = \underbrace{ \sum_{j \neq \x_t} R_t(j, \x_t) p_t(j)}_{\text{incoming}} - \underbrace{ \sum_{j \neq \x_t} R_t(\x_t, j) p_t(\x_t)}_{\text{outgoing}}
    \label{eq:continuity}
    \vspace{-0.1cm}
\end{equation}
The difference between the incoming and outgoing probability mass is the time derivative of the marginal $\partial_t p_t(\x_t)$.
Using our definition of $R_t(\x_t, \x_t)$, \cref{eq:continuity} can be succinctly written as $\partial_t p_t = R_t^\top p_t$ where the marginals are treated as probability mass vectors: $p_t \in [0,1]^\statespace$.
This defines an Ordinary Differential Equation (ODE) in a vector space.
We refer to the series of distributions $p_t \ \forall t \in [0,1]$ satisfying the ODE as a \emph{probability flow}. \looseness=-1
\begin{emphasis}
\underline{\textbf{Key terms:}}
A \textbf{CTMC} is defined by an initial distribution $p_0$ and rate matrix $R_t$.
Samples along CTMC dynamics are called a \textbf{sequence trajectory} $\x_t$.
The \textbf{probability flow} $p_t$ is the marginal distribution of $x_t$ at every time $t$.
We say $R_t$ \textbf{generates} $p_t$ if $\partial_t p_t = R_t^\top p_t \ \forall t \in [0,1]$.
\end{emphasis}

\section{Discrete Flow Models}
\label{sec:dmf}

A Discrete Flow Model (\method) is a \textbf{Discrete} data generative model built around a probability \textbf{Flow} that interpolates from noise to data.
To sample new datapoints, we simulate a sequence trajectory that matches the noise to data probability flow.
The flow construction allows us to combine \method with continuous data flow models to define a multimodal generative model.
Proofs for all propositions are in \cref{sec:apdx_proofs}.

\subsection{A Flow Model for Sampling Discrete Data}
\label{sec:sec:discrete_flow}
We start by constructing the data generating probability flow referred to as the \emph{generative flow}, $p_t$, that we will later sample from using a CTMC.
The generative flow interpolates from noise to data where $p_{0}(x_0) = \pnoise(x_0)$ and $p_{1}(x_1) = \pdata(x_1)$.
Since $\probflow$ is complex to consider directly, the insight of flow matching is to define $p_t$ using a simpler datapoint conditional flow, $\noisemarg(\cdot | \x_1)$ that we will be able to write down explicitly.
We can then define $p_t$ as
\begin{equation}
\label{eq:uncond_flow}
p_t(\x_t) \vcentcolon = \E_{\pdata(\x_1)} \left[ \noisemarg(\x_t | \x_1) \right].
\end{equation}
The conditional flow, $\noisemarg(\cdot | \x_1)$ interpolates from noise to the datapoint $\x_1$.
The conditioning allows us to write the flow down in closed form.
We are free to define $\noisemarg(\cdot | \x_1)$ as needed for the specific application.
The conditional flows we use in this paper linearly interpolate towards $\x_1$ from a uniform prior or an artificially introduced mask state, $M$:
\begin{talign}
\label{eq:corruption_flow}
\noisemarg^{\mathrm{unif}}(\x_t | \x_1) &= \mathrm{Cat}(t \kdelta{\x_1}{\x_t} + (1-t) \frac{1}{\statespace}),
\\
\noisemarg^{\mathrm{mask}}(\x_t | \x_1) &= \mathrm{Cat}(t \kdelta{\x_1}{\x_t} + (1-t) \kdelta{M}{\x_t}).
\end{talign}
We require our conditional flow to converge on the datapoint $\x_1$ at $t=1$, i.e. $\noisemarg(\x_t | \x_1) = \kdelta{\x_1}{\x_t}$.
We also require that the conditional flow starts from noise at $t=0$, i.e. $\noisemarg(\x_t | \x_1) = \pnoise(\x_t)$.
In our examples, $\pnoise^{\mathrm{unif}}(\x_t) = \frac{1}{S}$ and $\pnoise^{\mathrm{mask}}(\x_t) = \kdelta{M}{\x_t}$.
These two requirements ensure our generative flow, $p_t$, defined in \cref{eq:uncond_flow} interpolates from $\pnoise$ at $t=0$ towards $\pdata$ at $t=1$ as desired.
Next, we will show how to sample from the generative flow by exploiting $p_t$'s decomposition into conditional flows.

\renewcommand{\arraystretch}{1.3}
\begin{table*}[t]
\caption{
Comparison between continuous space linear interpolant flow models and DFMs with masking.
Both start with a conditional flow $\noisemarg(\x_t | \x_1)$ interpolating between data and noise.
For continuous, $\noisemarg(\x_t | \x_1) = \mathcal{N}(t \x_1, (1-t)^2 I)$ and for discrete we use $\noisemarg^{\mathrm{mask}}$.
Solving the Fokker-Planck or Kolmogorov equations with $\noisemarg(\x_t | \x_1)$ gives a data conditioned process, specified either by the velocity field ($\nu_t$) or the rate matrix ($R_t$).
We train a model to learn the unconditional process -- written analytically as the expected value of the conditional quantity -- which is then used for sampling.
The side-by-side comparison reveals the similar forms of each quantity.
}
\label{tab:cont_disc_comparison}
\begin{center}
\begin{small}
\begin{sc}
\begin{tabular}{lll}
\toprule
Quantity & Continuous & Discrete \\
\midrule
Fokker-Planck-Kolmogorov & $\partial_t p_t = - \nabla \cdot \left( v_t p_t \right)$ & $ \partial_t p_t = R_t^\top p_t$\\
Conditional process &  $\nu_t(\x_t | \x_1) = \frac{\x_1 - \x_t}{1-t}$ &  $R_t(\x_t, j | \x_1) = \frac{\kdelta{j}{\x_1}}{1-t} \kdelta{\x_t}{M} $ \\
Generative process & $\nu_t(\x_t) = \E_{\denoise(\x_1 | \x_t)} \left[ \nu_t(\x_t | \x_1) \right]$ & $R_t(\x_t, j) =  \E_{\denoise(\x_1 | \x_t)} \left[ R(\x_t, j | \x_1) \right]$ \\
Generative sampling & $\x_{t+\Dt} = \x_t + v_t(\x_t) \Dt$ & $\x_{t+\Dt} \sim \mathrm{Cat}(\kdelta{\x_t}{\x_{t+\Dt}}+ R_t(\x_t, \x_{t+\Dt})\Dt)$\\
\bottomrule
\end{tabular}
\end{sc}
\end{small}
\end{center}
\vskip -0.2in
\end{table*}
\renewcommand{\arraystretch}{1.0}

\subsubsection{Sampling}
\label{sec:sec:sampling}
To sample from $p_{\mathrm{data}}$ using the generative flow, $p_t$, we need access to a rate matrix $R_t(\x_t, j)$ that generates $p_t$.
Given a $R_t(\x_t, j)$, we could use \cref{eq:sample_euler_step} to simulate a sequence trajectory that begins with marginal distribution $\pnoise$ at $t=0$ and ends with marginal distribution $\pdata$ at $t=1$.
The definition of $p_t$ in \cref{eq:uncond_flow} suggests $R_t(\x_t, j)$ can also be derived as an expectation over a simpler conditional rate matrix.
Define $R_t(\x_t, j | \x_1)$ as a datapoint conditional rate matrix that generates $\noisemarg(\x_t | \x_1)$. We now show $R_t(\x_t, j)$ can indeed be defined as an expectation over $R_t(\x_t, j | \x_1)$.

\begin{proposition}
    \label{prop:unconditional_generative_rate}
    If $R_t(\x_t, j | \x_1)$ is a rate matrix that generates the conditional flow $\noisemarg(\x_t | \x_1)$, then 
    \vspace{-0.2cm}
    \begin{talign}
        R_t(\x_t, j) \vcentcolon = \E_{\denoise(\x_1 | \x_t)} \left[ R_t(\x_t, j | \x_1) \right]
        \label{eq:uncond_rate}
    \end{talign} 
    is a rate matrix that generates $p_t$ defined in \cref{eq:uncond_flow}.
    The expectation is taken over $\denoise(\x_1 | \x_t) = \frac{\noisemarg(\x_t | \x_1) \pdata(\x_1)}{p_t(\x_t)}$.
\end{proposition}

\vspace{-0.1cm}
Our aim now is to calculate $R_t(\x_t, j | \x_1)$ and $\denoise(\x_1 | \x_t)$ to plug into \cref{eq:uncond_rate}.
$\denoise(\x_1 | \x_t)$ is the distribution predicting clean data $\x_1$ from noisy data $\x_t$ and in \cref{sec:sec:training}, we will train a neural network $\denoise^\theta(\x_1 | \x_t)$ to approximate it.
In \cref{sec:rate_matrices}, we will show how to derive $R_t(\x_t, j|\x_1)$ in closed form.
Sampling pseudo-code is provided in \cref{alg:sampling}.
\begin{algorithm}[h]
    \caption{DFM Sampling}
    \label{alg:sampling}
    \begin{algorithmic}[1]
        \STATE \textbf{init} $t = 0, \x_0 \sim p_0$, choice of $R_t(\x_t, \cdot | \x_1)$ (\cref{sec:rate_matrices})
        \WHILE{$t < 1$}
        \STATE $R_t^\theta(\x_t, \cdot) \gets \mathbb{E}_{p^\theta_{1|t}(\x_1 | \x_t) } \left[ R_t(\x_t, \cdot|\x_1) \right] $
        \STATE $\x_{t+ \Dt} \sim \mathrm{Cat}\left(\kdelta{\x_t}{\x_{t+ \Dt}} + R_t^\theta(\x_t, \x_{t+ \Dt})  \Dt\right) $
        \STATE $t \gets t + \Dt$
        \ENDWHILE
        \STATE \textbf{return} $\x_1$
    \end{algorithmic}
\end{algorithm}

We discuss further CTMC sampling methods in \cref{sec:apdx_ctmc_sampling}.
Our construction of the generative flow from conditional flows is analogous to the construction of generative probability paths from conditional probability paths in \citet{lipman2022flow}, where instead of a continuous vector field generating the probability path, we have a rate matrix generating the probability flow.
We expand on these links in \cref{tab:cont_disc_comparison}.

\subsubsection{Training}

\label{sec:sec:training}
We train a neural network with parameters $\theta$, $\denoise^\theta(\x_t | \x_1)$, to approximate the true denoising distribution using the standard cross-entropy i.e. learning to predict the clean datapoint $\x_1$ when given noisy data $\x_t \sim \noisemarg(\x_t | \x_1)$.
\begin{equation}
    \label{eq:ce_loss}
    \mathcal{L}_{\mathrm{ce}} = \E_{\pdata(\x_1) \mathcal{U}(t; 0, 1) \noisemarg(\x_t | \x_1)} \left[ \log \denoise^\theta(\x_1 | \x_t) \right]
\end{equation}
where $\mathcal{U}(t; 0, 1)$ is a uniform distribution on $[0, 1]$.
$\x_t$ can be sampled from $\noisemarg(\x_t | \x_1)$ in a simulation-free manner by using the explicit form we wrote down for $\noisemarg$ e.g. \cref{eq:corruption_flow}.
In \cref{sec:apdx_analysis_of_training_objective}, we analyse how $\mathcal{L}_{\mathrm{ce}}$ relates to the model log-likelihood and its relation to the Evidence Lower Bound (ELBO) used to train diffusion models.
We stress that $\mathcal{L}_{\mathrm{ce}}$ does not depend on $R_t(\x_t, j | \x_1)$ and so we can postpone the choice of $R_t(\x_t, j | \x_1)$ until after training.
This enables inference time flexibility in how our discrete data is sampled.

\subsection{Choice of Rate Matrix}
\label{sec:rate_matrices}

The missing piece in \cref{eq:uncond_rate} is a conditional rate matrix $R_t(\x_t, j | \x_1)$ that generates the conditional flow $\noisemarg(\x_t | \x_1)$.
There are many choices for $R_t(\x_t, j | \x_1)$ that all generate the same $\noisemarg(\x_t | \x_1)$ as we later show in \cref{prop:db_rate}.
In order to proceed, we start by giving one valid choice of rate matrix and from this, build a set of rate matrices that all generate $\noisemarg$.
At inference time, we can then pick the rate matrix from this set that performs the best.
Our starting choice for a rate matrix that generates $\noisemarg$ is defined for $\x_t \neq j$ as,
\begin{align}
    \relurate_t(\x_t, j | \x_1) \vcentcolon = \frac{\relu \left( \partial_t \noisemarg(j | \x_1) - \partial_t \noisemarg(\x_t | \x_1) \right)}{ \statespace \cdot \noisemarg(\x_t | \x_1) }
    \label{eq:relu_rate}
\end{align}
where $\relu(a) = \text{max}(a, 0)$ and $\partial_t \noisemarg$ can be found by differentiating our explicit form for $\noisemarg$. This assumes $\noisemarg(\x_t | \x_1) > 0$, see \cref{sec:rate_match_marginal_proof} for the full form.
We first heuristically justify $\relurate_t$ and then prove it generates $\noisemarg(\x_t | \x_1)$ in \cref{prop:relu_rate}.
$\relurate_t$ can be understood as distributing probability mass to states that require it.
If $\partial_t \noisemarg(j | \x_1) > \partial_t \noisemarg(\x_t | \x_1)$ then state $j$ needs to gain more probability mass than the current state $\x_t$ resulting in a positive rate.
If $\partial_t \noisemarg(j | \x_1) \leq \partial_t \noisemarg(i | \x_1)$ then state $\x_t$ should give no mass to state $j$ hence the $\relu$.
This rate should then be normalized by the probability mass in the current state.
The $\relu$ ensures off-diagonal elements of $\relurate_t$ are positive and is inspired by \citet{zhang2023formulating}.

\begin{proposition}
\label{prop:relu_rate}
Assuming zero mass states, $\noisemarg(j | \x_1)=0$, have $\partial_t \noisemarg(j | \x_1) = 0$, then $\relurate_t$ generates $\noisemarg(\x_t | \x_1)$.
\end{proposition}

The proof is easy to derive by substituting $\relurate_t$ along with $\noisemarg(\x_t | \x_1)$ into the Kolmogorov equation \cref{eq:continuity}.
The forms for $\relurate_t(\x_t, j | \x_1)$ under $\noisemarg^{\mathrm{unif}}$ or $\noisemarg^{\mathrm{mask}}$ are simple
\begin{equation}
\textstyle
     R_t^{* \mathrm{unif}} = \frac{\kdelta{\x_1}{j} (1 - \kdelta{\x_1}{\x_t})}{1-t}, \quad R_t^{* \mathrm{mask}}= \frac{\kdelta{\x_1}{j} \kdelta{\x_t}{M}}{1-t}
\end{equation}
as we derive in \cref{sec:apdx_implementation_details}.
Using $\relurate_t$ as a starting point, we now build out a set of rate matrices that all generate $\noisemarg$. 
We can accomplish this by adding on a second rate matrix that is in detailed balance with $\noisemarg$.
\begin{proposition}
\label{prop:db_rate}
    Let $\dbrate_t$ be a rate matrix that satisfies the detailed balance condition for $\noisemarg$,
    \vspace{-0.1cm}
    \begin{equation}
        \noisemarg(i | \x_1) \dbrate_t(i, j | \x_1) = \noisemarg(j | \x_1) \dbrate_t(j, i | \x_1),
        \label{eq:detailed_balance}.
    \end{equation}
    Let $R_t^\noise$ be defined by $\relurate_t$, $\dbrate_t$ and parameter $\noise \in \mathbb{R}^{\geq 0}$,
    \vspace{-0.1cm}
    \begin{equation}
    R_t^\noise \vcentcolon = \relurate_t + \noise \dbrate_t.
    \label{eq:db_rate}
    \end{equation} 
    Then we have $R_t^\noise$ generates $\noisemarg(\x_t | \x_1)$, $\forall \noise \in \mathbb{R}^{\geq 0}$.
\end{proposition}
The detailed balance condition intuitively enforces the incoming probability mass, $\noisemarg(j | \x_1) \dbrate_t(j, i | \x_1)$ to equal the outgoing probability mass, $\noisemarg(i | \x_1) \dbrate_t(i, j | \x_1)$.
Therefore, $\dbrate_t$ has no overall effect on the probability flow and can be added on to $\relurate_t$ with the combined rate still generating $\noisemarg$.
In many cases, \cref{eq:detailed_balance} is easy to solve for $\dbrate_t$ due to the explicit relation between elements of $\dbrate_t$ as we exemplify in \cref{sec:apdx_implementation_details}.
Detailed balance has been used previously in CTMC generative models \citep{campbell2022continuous} to make post-hoc inference adjustments.

\paragraph{CTMC stochasticity.} 
We now have a set of rate matrices, $\{ R_t^\noise : \noise \geq 0 \}$, that all generate $\noisemarg$.
We can plug any one of these into our definition for $R_t(\x_t, j)$ (\cref{eq:uncond_rate}) and sample novel datapoints using \cref{alg:sampling}.
The chosen value for $\noise$ will influence the dynamics of the CTMC we are simulating.
For large values of $\noise$, the increased influence of $\dbrate_t$ will cause large exchanges of probability mass between states.
This manifests as increasing the frequency of jumps occurring in the sequence trajectory.
This leads to a short auto-correlation time for the CTMC and a high level of unpredictability of future states given the current state.
We refer to the behaviour that $\noise$ controls as \textit{CTMC stochasticity}. \cref{fig:overview}B shows examples of high and low $\noise$.

On a given task, we expect there to be an optimal stochasticity level.
Additional stochasticity improves performance in continuous diffusion models \cite{cao2023exploring, xu2023restart}, but too much stochasticity can result in a poorly performing degenerate CTMC.
In some cases, setting $\noise=0$, i.e. using $\relurate_t$, results in the minimum possible number of jumps because the $\relu$ within $\relurate_t$ removes state pairs that needlessly exchange mass \citep{zhang2023formulating}.
\begin{proposition}
\label{prop:Roptimality}
    For $\noisemarg^{\mathrm{unif}}$ and $\noisemarg^{\mathrm{mask}}$,
   $\relurate_t$ generates $\noisemarg$ whilst minimizing the expected number of jumps during the sequence trajectory. This assumes multi-dimensional data under the factorization assumptions listed in \cref{sec:factorization}.
\end{proposition}

\vspace{-0.3cm}
\subsection{DFMs Recipe}
\label{sec:sec:recipe}
We now summarize the key steps of a DFM. PyTorch code for a minimal \method implementaton is provided in \cref{sec:apdx_implementation_details}.
\vspace{-8pt}
\begin{enumerate}[leftmargin=5mm,noitemsep,topsep=5pt,partopsep=0pt]
    \item Define the desired noise schedule $\noisemarg(\x_t | \x_1)$ (\cref{sec:sec:discrete_flow}).
    \item Train denoising model $\denoise^\theta(\x_1 | \x_t)$ (\cref{sec:sec:training}).
    \item Choose rate matrix $R_t^\noise$ (\cref{sec:rate_matrices}).
    \item Run sampling (\cref{alg:sampling}).
\end{enumerate}
\vspace{-0.3cm}

\section{Multimodal Protein Generative Model}
\label{sec:cogen}
Using our flow formulation on discrete state spaces, we can now combine a \method with a flow on a continuous space to define a multimodal generative flow. We use this to perform protein joint structure-sequence generation.
A protein can be modeled as a linear chain of residues, each with an assigned amino acid and 3D atomic coordinates.
Protein co-design aims to jointly generate the amino acids (sequence) and coordinates (structure).
Prior works have used a generative model on one modality (sequence or structure) with a separate model to predict the other (see \cref{sec:sec:protein_related}).
Instead, our approach uses a single generative model to jointly sample both modalities: a \method for the sequence and a flow model, FrameFlow \citep{yim2023fast}, for the structure.
We refer to our multimodal flow model as \textbf{\protmodel}.

\subsection{Multimodal Flows}
Following FrameFlow, we refer to the protein structure as the \emph{backbone} atomic coordinates of each residue.
We leave modeling side-chain atoms as a follow-up work.
The structure is represented as elements of $\SE$ to capture the rigidity of the local frames along the backbone \citep{yim2023se}.
A protein of length $D$ residues can then be represented as $\{(\trans^d, \rots^d, \amino^d)\}_{d=1}^D $ where $\trans \in \real^3$ is the translation of the residue's Carbon-$\alpha$ atom, $\rots \in \SO$ is a rotation matrix of the residue's local frame with respect to global reference frame, and $\amino \in \{1, \dots, 20 \} \cup \{M\}$ is one of 20 amino acids or the mask state $M$. For brevity, we refer to the residue state as $\residue^d = (\trans^d, \rots^d, \amino^d)$ and let the full protein's structure and sequence as $\allresidue = \{\residue^d\}_{d=1}^D$. We define the multimodal conditional flow as $p_{t|1}(\allresidue_t | \allresidue_1)$ which is a shorthand for a probability density over the continuous variables and a probability mass function over the discrete variables. We define $p_{t|1}(\allresidue_t | \allresidue_1)$ to factorize over both dimensions and modality.
\begin{equation}
    \scalebox{0.95}{$\displaystyle p_{t|1}(\allresidue_t | \allresidue_1) \vcentcolon = \prod_{d=1}^D p_{t|1}(\trans_t^d | \trans_1^d) p_{t|1}(\rots_t^d | \rots_1^d) p_{t|1}(\amino_t^d | \amino_1^d)$}
    \label{eq:total_multimodal_conditional_flow}
\end{equation}
Following \citet{yim2023fast}, $p_{t|1}(\trans_t^d | \trans_1^d)$ and $p_{t|1}(\rots_t^d | \rots_1^d)$ are defined implicitly through specifying how samples $\trans_t^d$, $\rots_t^d$ are generated from $p_{t|1}(\trans_t^d | \trans_1^d)$, $p_{t|1}(\rots_t^d | \rots_1^d)$,
\begin{talign}
    \trans_t^d &= t \trans_1^d + (1-t) \trans_0^d , \  \trans_0^d \sim \mathcal{N}(0, I) \label{eq:trans_interp} \\ 
    \rots_t^d &= \mathrm{exp}_{\rots_0^d}\left( t \mathrm{log}_{\rots_0^d} (\rots_1^d)\right), \  \rots_0^d \sim \mathcal{U}_{\text{SO}(3)}, \label{eq:rots_interp}
\end{talign}
where $\mathrm{exp}$ and $\log$ are the exponential and logarithmic maps. 
$\mathcal{U}_{\SO}$ is the uniform distribution on $\SO$.
Following \cref{sec:dmf}, $p_{t|1}(\amino_t^d | \amino_1^d)$ is defined explicitly.
\begin{equation}
    \scalebox{0.95}{$\displaystyle p_{t|1}(\amino_t^d | \amino_1^d) = \mathrm{Cat}\big( t \kdelta{\amino_1^d}{\amino_{t}^d} + (1-t) \kdelta{M}{\amino_{t}^d} \big)$} \label{eq:amino_interp}
\end{equation}
Our conditional trajectory that follows this conditional flow will be an ODE on the continuous modalities with a CTMC for the amino acids. The conditional ODE on translations and rotations is parameterized through conditional velocities $v_\trans^d(\trans_t^d | \trans_1^d) \in \mathbb{R}^3$, $v_\rots^d(\rots_t^d | \rots_1^d) \in \text{Tan}_{r_t^d} \SO$ \cite{yim2023fast}. $v_\trans^d$ is a standard Euclidean vector field whereas $v_\rots^d$ is a vector field on the Riemannian Manifold $\SO$ \cite{chen2023riemannian}. The trajectory can be simulated using Euler steps with step size $\Delta t$,
\begin{talign}
    \trans_{t+\Delta t}^d &= \trans_t^d + v_\trans^d(\trans_t^d | \trans_1^d) \Delta t \\
    \rots_{t+\Delta t}^d &= \mathrm{exp}_{\rots_t^d} ( \Delta t \cdot v_{\rots}^d(\rots_t^d | \rots_1^d )) \label{eq:conditional_multimodal_update} \\
    \amino_{t+\Delta t}^d &\sim \text{Cat} \big( \kdelta{\amino_t^d}{\amino_{t+\Delta t}^d} + R_t^d(\amino_t^d, \amino_{t+\Delta t}^d | \amino_1^d) \Delta t \big).
\end{talign}
We choose $v_\trans^d$ such that it individually generates the $p_{t|1}(\trans_t^d | \trans_1^d)$ given by \cref{eq:trans_interp} if it were simulated by itself in $\mathbb{R}^3$. Similarly, for $v_\rots^d$ and $R_t^d$, they are chosen such that they individually generate $p_{t|1}(\rots_t^d | \rots_1^d)$ (\cref{eq:rots_interp}) and $p_{t|1}(\amino_t^d | \amino_1^d)$ (\cref{eq:amino_interp}) respectively. The explicit forms for $v_\trans^d$, $v_\rots^d$ and $R_t^d$ are as follows,
\begin{align}
    v_\trans^d(\trans_t^d | \trans_1^d) &= ( \trans_1^d - \trans_t^d) / (1 - t)\\
    v_\rots^d(\rots_t^d | \rots_1^d) &= \mathrm{log}_{\rots_t^d} ( \rots_1^d) / (1 - t) \label{eq:multimodal_conditional_velocities}\\
    R_t^d(\amino_t^d, j^d | \amino_1^d) &= \kdelta{j^d}{\amino_1^d} \kdelta{ \amino_t^d}{M} / (1-t).
\end{align}
wth velocities following \citet{yim2023fast} and rate matrix derived in \cref{sec:apdx_masking_example} assuming $\eta=0$. The following proposition verifies these choices are consistent with our initial definition of $p_{t|1}(\allresidue_t | \allresidue_1)$.
\begin{proposition}
 The multimodal process defined by \cref{eq:multimodal_conditional_velocities} has the flow $p_{t|1}(\allresidue_t | \allresidue_1)$ given by \cref{eq:total_multimodal_conditional_flow}.
\label{prop:multimodal_consistency}
\end{proposition}
We would now like to be able to sample a trajectory that follows the unconditional flow. Mirroring \cref{prop:unconditional_generative_rate}, we again find that the desired unconditional velocities and rate matrix are expectations of their respective conditional quantities.
\begin{proposition}
\label{prop:multimodal_unconditional_flow}
The following velocities and rate matrix together generate $p_t(\allresidue_t) = \E_{\pdata(\allresidue_1)}\left[ p_{t|1}(\allresidue_t | \allresidue_1) \right]$,
\begin{align}
    v_\trans^d(\allresidue_t) &= \E_{p_{1|t}(\trans_1^d | \allresidue_t)} \left[ v_\trans^d(\trans_t^d | \trans_1^d) \right]\\
    v_\rots^d(\allresidue_t) &= \E_{p_{1|t}(\rots_1^d | \allresidue_t)} \left[ v_\rots^d(\rots_t^d | \rots_1^d) \right]\\
    R_t^d(\allresidue_t, j^d) &= \E_{p_{1|t}(\amino_1^d | \allresidue_t)} \left[ R_t^d(\amino_t^d, j^d | \amino_1^d) \right].
\end{align}

\end{proposition}
We note that even though the conditional flow is defined to factorize over modality and dimension, the unconditional generative flow has coupled modalities and dimensions because each velocity and rate matrix depends on the entire corrupted protein state $\allresidue_t$.

Thus far, we have assumed the same noise level in all modalities. To enable flexible sampling options, we can use a noise level for the structure, $t$, that is independent to the noise level of the sequence, $\tilde{t}$ \citep{albergo2023multimarginal}. We let $\allresidue_{t, \tilde{t}} = ( \trans_t^{1:D}, \rots_t^{1:D}, \amino_{\tilde{t}}^{1:D})$ and use a conditional flow of 
\begin{equation}
    p_{t, \tilde{t}|1}(\allresidue_{t, \tilde{t}} | \allresidue_1) = \prod_{d=1}^D p_{t|1}(\trans_t^d | \trans_1^d) p_{t|1}(\rots_t^d | \rots_1^d) p_{\tilde{t}|1}(\amino_{\tilde{t}}^d | \amino_1^d).
\end{equation}
The unconditional flow then becomes $p_{t, \tilde{t}}(\allresidue_{t, \tilde{t}}) = \E_{\pdata(\allresidue_1)} [p_{t, \tilde{t}|1}(\allresidue_{t, \tilde{t}} | \allresidue_1) ]$, with the expectations for the unconditional velocities and rate matrix in \cref{prop:multimodal_unconditional_flow} now computed using $p_{1|t, \tilde{t}}(\cdot | \allresidue_{t, \tilde{t}})$ instead of $p_{1|t}(\cdot | \allresidue_t)$.

\subsection{Training}
During training, our network will take as input the noised protein $\allresidue_{t,\tilde{t}}$ and predict the denoised translations $\hat{\trans}_1(\allresidue_{t,\tilde{t}})$, rotations $\hat{\rots}_1(\allresidue_{t,\tilde{t}})$, and amino acid distribution $p_\theta(\amino_1 | \allresidue_{t,\tilde{t}})$.
We then parameterize the unconditional velocities and rate matrix in terms of these predicted quantities.
\begin{talign}
    &\vf_{\trans}^d(\allresidue_{t, \tilde{t}}) = \frac{\hat{\trans}_1^d(\allresidue_{t, \tilde{t}}) - \trans_t^d}{1-t}, \,\, \vf_{\rots}^d(\allresidue_{t, \tilde{t}}) = \frac{\mathrm{log}_{\rots_{t}^d} (\hat{\rots}_1^d(\allresidue_{t,\tilde{t}}))}{1-t}, \\
    &R_{\tilde{t}}^{\theta d}(\allresidue_{t, \tilde{t}}, j) = \frac{p_\theta(\amino_1^d = j | \allresidue_{t, \tilde{t}})}{1-\tilde{t}} \kdelta{\amino_{\tilde{t}}^d}{M}. \label{eq:residue_vf}
\end{talign}
In order for these to match their optimum values given in \cref{prop:multimodal_unconditional_flow}, we minimize the following loss
\begin{talign}
    \E \Big[ \sum_{d=1}^D & \frac{\norm{ \hat{\trans}_1^d(\allresidue_{t,\tilde{t}} ) - \trans_1^d  }^2}{1-t} - \log p_\theta(\amino_1^d | \allresidue_{t,\tilde{t}}) \label{eq:cogen_loss} \\ + &\frac{\norm{ \mathrm{log}_{\rots_t^d}\left(\hat{\rots}_1^d(\allresidue_t) \right) - \mathrm{log}_{\rots_t^d} \left( \rots_1^d \right)  }^2}{1-t}\Big].
\end{talign}
where the expectation is over $t, \tilde{t} \sim \unif(0,1)$, $\allframes_{1,1} \sim \pdata$ and $\allframes_{t,\tilde{t}} \sim p_{t, \tilde{t} | 1}(\allframes_{t, \tilde{t}} | \allframes_{1,1})$.
Our independent $t$, $\tilde{t}$ objective enables the model to learn over different relative levels of corruption between the sequence and structure.
\cref{eq:cogen_loss} corresponds to the flow matching loss for continuous data and the DFMs loss \cref{eq:ce_loss} for discrete amino acids.
The neural network architecture is modified from FrameFlow with a larger transformer, smaller Invariant Point Attention, and extra multi-layer perception head to predict the amino acid logits.

\subsection{Sampling}
To sample the generative model, we use the update equations from \cref{eq:conditional_multimodal_update} but with the learned unconditional velocities and rate matrix. Furthermore, we find sample quality can be improved by using the exponential rate scheduler for rotations from \citet{bose2023se}. In practice, this means $v_\rots^d$ has the following form,
\begin{equation}
    \vf_{\rots}^d(\allresidue_{t, \tilde{t}}) = c\cdot\mathrm{log}_{\rots_{t}^d} (\hat{\rots}_1^d(\allresidue_{t,\tilde{t}})).
\end{equation}
We use $c=10$ following \cite{yim2023fast}.
When sampling the amino acids, we also found it beneficial to utilize purity \citep{tang2022improved} to choose which indices to unmask at each step.
The advantage of training with decoupled time schedules is that we have freedom to arbitrarily sample with any combination of $(t, \tilde{t})$.
We use this to perform conditional inpainting where one of the modalities is fixed by setting $t$ or $\tilde{t}$ equal to 1.
For example, setting $t=1$ then using Euler steps to update $\tilde{t}$ from $0 \rightarrow 1$ performs sequence generation conditioned on the structure.
We summarize the capabilities in \cref{fig:overview}C and in \cref{tab:flexible_t_examples}.
\vspace{-0.5cm}
\begin{table}[h]
  \centering
  \caption{Flexible multimodal sampling.}
  \begin{tabular}{l|c|c|c}
    \toprule
    & Codesign & Inverse folding & Forward folding \\
    \midrule
    $\trans_{t},\rots_{t}$ & $t: 0 \rightarrow 1$ & $t=1$ & $t: 0 \rightarrow 1$   \\
    $\amino_{\tilde{t}}$ & $\tilde{t}: 0 \rightarrow 1$ & $\tilde{t}: 0 \rightarrow 1$ & $\tilde{t}=1$  \\
    \bottomrule
  \end{tabular}
  \label{tab:flexible_t_examples}
\end{table}
\vspace{-0.5cm}

\section{Related Work}
\label{sec:related}
\textbf{Discrete Diffusion Models.} \
Our continuous time flow builds on work that extends discrete diffusion \cite{hoogeboom2021argmax, austin2021structured} to continuous time \cite{campbell2022continuous, sun2022score, santos2023blackout, lou2023discrete} but we simplify and extend the framework.
We are not restricted to noising processes that can be defined by a matrix exponential as we just write $\noisemarg$ down directly and we have the freedom to choose $R_t(\x_t, j | \x_1)$ at inference time rather than being restricted to the time reversal.
We show how DFMs encompasses prior discrete diffusion models in \cref{sec:apdx_comparison_diff}.
For molecular retrosynthesis, \citet{igashov2023retrobridge} also considered a data conditional process, but did not build a modeling framework around it.
\citet{zhang2023formulating} constructed low-stochasticity rate matrices and their derivation provides the building blocks of \cref{prop:relu_rate}.
Some works have built a multimodal diffusion model for molecule generation \citep{peng2023moldiff, vignac2023midi, hua2023mudiff} whereas we focus on protein co-design using flows.
We discuss further related work in \cref{sec:apdx_related_work}.

\textbf{Protein Generation.} \
\label{sec:sec:protein_related}
Diffusion and flow models have risen in popularity for generating novel and diverse protein backbones \citep{yim2023se,yim2023fast,bose2023se,lin2023generating,ingraham2023illuminating}.
RFDiffusion achieved notable success by generating proteins validated in wet-lab experiments \citep{watson2023novo}.
However, these methods required a separate model for sequence generation.
Some works have focused only on sequence generation with diffusion models \citep{alamdari2023protein,gruver2023protein,yang2023fast,yi2023graph}.
We focus on co-design which aims to jointly generate the structure and sequence.

Prior works have attempted co-design.
ProteinGenerator \citep{lisanza2023joint} performs Euclidean diffusion over one-hot amino acids while predicting the structure at each step with RosettaFold \citep{baek2021accurate}.
Conversely, Protpardelle \citep{chu2023all} performs Euclidean diffusion over structure while iteratively predicting the sequence.
\protmodel instead uses a generative model over \emph{both} the structure and sequence which allows for flexibility in conditioning at inference time (see \cref{sec:sec:codesign}).
\citet{luo2022antigen,shi2022protein} are co-design methods, but are limited to generating CDR loops on antibodies.
Lastly, \citet{anand2022protein} presented diffusion on structure and sequence, but did not report standard evaluation metrics nor is code available.

\section{Experiments}
\label{sec:experiments}
We first show that tuning stochasticity at sample time improves pure discrete generative modeling performance by modeling text data.
We then evaluate \protmodel, the first flow model on discrete and continuous state spaces.
We show \protmodel provides state-of-the-art-performance on protein generation compared to prior approaches that do not generate using a true multimodal generative model.
Finally, we investigate \protmodel's crossmodal properties of how varying the sequence sampling affects the structure.
\subsection{Text Modeling}
\label{sec:text}
\begin{figure}[t]
    \centering
    \hspace{-0.5cm}
    \includegraphics[trim={0 0.4cm 0 0}, clip, width=8.5cm]{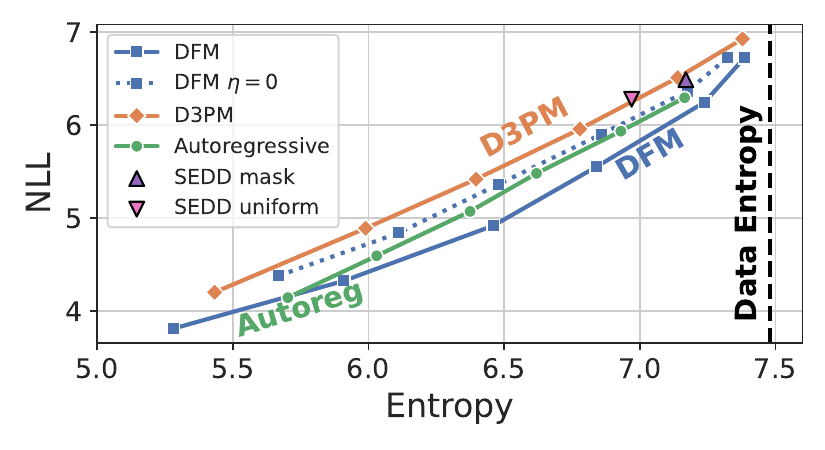}
    \vspace{-0.3cm}
    \caption{Negative log-likelihood as measured by GPT-J-6B versus sample entropy. For \method, D3PM and autoregressive, we sweep the logit temperature for $\denoise^\theta(\x_1 | \x_t)$ over $\{0.5, 0.6, , \dots, 1 \}$. We aim to minimize NLL whilst staying close to the dataset entropy.}
    \label{fig:text_figure}
    \vspace{-0.5cm}
\end{figure}
\paragraph{Set-up.}
We model the text dataset, text8 \cite{mahoney2006large}, which is $100$MB of text from English Wikipedia.
We model at the character level, following \cite{austin2021structured}, with $\statespace=28$ categories for $26$ lowercase letters, a white-space and a mask token. We split the text into chunks of length $D=256$. We train a DFM using $\noisemarg^{\mathrm{mask}}$ and parameterize the denoising network using a transformer with $86$M non-embedding parameters, full details are in \cref{sec:apdx_text_experiment_details}.

\textbf{Results.} Text samples are evaluated following \citet{strudel2022self}.
A much larger text model, we use GPT-J-6B \cite{gpt-j}, is used to evaluate the negative log-likelihood (NLL) of the generated samples.
The NLL metric alone can be gamed by repeating similar sequences, so the token distribution entropy is also measured.
Good samples should have both low NLL and entropy close to the data distribution.
For a given value of $\noise$, we create a Pareto-frontier in NLL vs entropy space by varying the temperature applied to the $\denoise^\theta(\x_1 | \x_t)$ logits during the softmax operation.
\cref{fig:text_figure} plots the results for varying levels of $\noise$ and sampling temperature.
For comparison, we also include results for the discrete diffusion D3PM method with absorbing state corruption \cite{austin2021structured} as well as the Score Entropy Discrete Diffusion (SEDD) method of \citet{lou2023discrete} using both uniform and absorbing style corruption.
SEDD does not have logits that can be temperature scaled and so only single points in NLL vs entropy space are shown.
We find the \method performs better than D3PM and SEDD due to our additional sample time flexibility. We are able to choose the value of $\noise$ that optimizes the Pareto-frontier at sample time (here $\noise=15$) whereas D3PM and SEDD do not have this flexibility. We show the full $\noise$ sweep in \cref{sec:apdx_text_experiment_details} and show the frontier for $\noise=0$ in \cref{fig:text_figure}. 
When $\noise=0$, performance is similar to D3PM due to DFMs being a continuous time generalization of D3PM at this setting, see \cref{sec:apdx_absorbing_state_link_to_d3pm}.
We also include results for an autoregressive model in \cref{fig:text_figure} for reference; however, we note this is not a complete like-for-like comparison as autoregressive models require much less compute to train than diffusion based models \cite{gulrajani2023likelihood}.

\subsection{Protein generation}
\label{sec:codesign}

\begin{table*}[!ht]
\vskip -0.2in
\caption{Co-design results. Abbreviations: Designability (\textbf{DES.}), Diversity (\textbf{DIV.}), Novelty (\textbf{NOV.}). For Protpardelle, we report Co-design 1 as same numbers as PMPNN 1 since their co-design approach employs PMPNN. We note this is not co-generation since PMPNN is used while \protmodel explicitly learns co-generation without using PMPNN.}
\vskip 3pt
\begin{center}
\label{table:designability}
\begin{small}
\begin{sc}
\begin{tabular}{l||ccc|ccc|ccc}
\toprule
\textbf{Method} & \multicolumn{3}{c|}{\textbf{Co-design 1}} & \multicolumn{3}{c|}{\textbf{PMPNN 8}} & \multicolumn{3}{c}{\textbf{PMPNN 1}} \\
 & Des. ($\uparrow$) & Div. ($\uparrow$)  & Nov. ($\downarrow$) & Des. & Div. & Nov. & Des. & Div. & Nov. \\
 \midrule
Protpardelle & 0.63* & 38* & 0.60* & 0.90 & 47 & \textbf{0.59} & 0.63 & 38 & \textbf{0.60} \\
ProteinGenerator & 0.37 & 35 & 0.69 & 0.89 & 75 & 0.65 & 0.78 & 64 & 0.66 \\
RFdiffusion & & N/A & & 0.87 & \textbf{158} & 0.63 & 0.66 & 111 & 0.64  \\
\hline
\protmodel & \textbf{0.86} & \textbf{143} & \textbf{0.61} & \textbf{0.99} & 156 & 0.61 & \textbf{0.88} & \textbf{143} & 0.61 \\
\protmodel w/o distillation & 0.42 & 72 & 0.62 & 0.89 & 126 & 0.62 & 0.71 & 101 & 0.63  \\
\protmodel w/o sequence &  & N/A &  & \textbf{0.99} & 116 & 0.63 & 0.86 & 97 & 0.62 \\
\bottomrule
\end{tabular}
\vspace{-0.55cm}
\end{sc}
\end{small}
\end{center}
\end{table*}

\textbf{Metrics.} \
Evaluating the quality of structure-sequence samples is performed with \emph{self-consistency} which measures how consistent a generated sequence is with a generated structure by testing how accurately a protein folding network can predict the structure from the sequence.
Specifically, either AlphaFold2 \citep{jumper2021highly} or ESMFold \citep{lin2023evolutionary}, is first used to predict a structure given only the generated sequence.
Our results will use ESMFold but we show results with AlphaFold2 in \cref{sec:apdx_protein_codesign_details}.
Then, we calculate scRMSD: the Root Mean Squared Deviation between the generated and predicted structure's backbone atoms.
The generated structure is called \emph{designable} if $\text{scRMSD} < 2\text{\AA{}}$.

Structure-only generative models such as RFdiffusion first use ProteinMPNN (PMPNN) \citep{dauparas2022robust} to predict a sequence given the generated structure in order to then be able to use the self-consistency metric.
We present three variants of self-consistency:
\vspace{-0.3cm}
\begin{itemize}[leftmargin=4mm,
    noitemsep,
    topsep=5pt,
    partopsep=0pt]
    \item \emph{Co-design 1}: use the sampled (structure, sequence) pair.
    \item \emph{PMPNN 8}: take only the sampled structure and predict 8 sequences with PMPNN. Then use ESMFold to predict a new structure for each sequence.
    The final structure-sequence pair is the original sampled structure along with the PMPNN sequence with minimum scRMSD.
    \item \emph{PMPNN 1}: same as PMPNN 8 except PMPNN only generates one sequence.
\end{itemize}
\vspace{-0.3cm}
PMPNN 8 and PMPNN 1 evaluate only the quality of a model's generated structures whereas, for co-design models, Co-design 1 evaluates the quality of a model's generated (structure, sequence) pairs.
The comparison between PMPNN 1 and Co-design 1 allows for evaluating the quality of co-designed sequences.
PMPNN 8 is the procedure used in prior structure-only works.
As our main metric of sample quality, we report \emph{designability} as the percentage of designable samples.
As a further sanity check, designable samples are then evaluated on \emph{diversity} and \emph{novelty}.
We use FoldSeek \citep{van2022foldseek} to report diversity as the number of unique clusters while novelty is the average TM-score \citep{zhang2005tm} of each sample to its most similar protein in PDB.

\textbf{Training.}
Our training data consisted of length 60-384 proteins from the Protein Data Bank (PDB) \citep{berman2000protein} that were curated in \citet{yim2023se} for a total of 18684 proteins.
Training took 200 epochs over 3 days on 4 A6000 Nvidia GPUs using the AdamW optimizer \citep{loshchilov2017decoupled} with learning rate 0.0001.

\textbf{Distillation.}
\protmodel with PDB training generated highly designable structures.
However, the co-designed sequences suffered from lower designability than PMPNN.
Our analysis revealed the original PDB sequences achieved worse designability than PMPNN.
We sought to improve performance by distilling knowledge from other models.
To accomplish this, we first replaced the original sequence of each structure in the training dataset with the lowest scRMSD sequence out of 8 generated by PMPNN conditioned on the structure.
Second, we generated synthetic structures of random lengths between 60-384 using an initial \protmodel model and added those that passed PMPNN 8 designability into the training dataset with the lowest scRMSD PMPNN sequence.
We found that we needed to add only an extra 4179 examples to the original set of 18684 proteins to see a dramatic improvement.
This procedure can be seen as a single step of reinforced self training (ReST) \cite{gulcehre2023reinforced}.

\subsubsection{Co-design results.}
\label{sec:sec:codesign}
Following RFdiffusion's benchmark, we sample 100 proteins for each length 70, 100, 200, and 300.
We sample \protmodel with 500 timesteps using a temperature of 0.1 (PMPNN also uses 0.1) and stochasticity level $\noise=20$.
We compare our structure quality to state-of-the-art structure generation method RFdiffusion.
For co-design, we compare to Protpardelle and ProteinGenerator.
All methods were ran using their publicly released code and evaluated identically.

Our results are presented in \Cref{table:designability} where report the average of three seeds for each metric -- see \Cref{tab:metrics_with_err} for results with standard error.
We find that \protmodel's co-design capabilities surpass previous co-design methods, none of which use a joint multimodal generation process.
\protmodel generates sequences that are consistent with the generated structure at a comparable level to PMPNN which we see through comparing the Co-design 1 and PMPNN 1 designability.
On pure structure generation, we find that \protmodel outperforms all baselines in terms of structure quality measured by PMPNN 8 designability.
\protmodel also attains comparable diversity and novelty to previous approaches.
We ablate our use of distillation and find that distillation results in overall designability improvements while also improving diversity.
Finally, we train our exact same architecture except only modeling the structure on the distilled dataset using the loss presented in \citet{yim2023fast}.
We find our joint structure-sequence model achieves the same structural quality as the structure-only version, however, additionally including the \textit{sequence} in our generative process induces extra \textit{structural} diversity.

\begin{figure}[t]
    \centering
    \vspace{-0.05cm}
    \includegraphics[trim={0 0.7cm 0 0}, width=7.5cm]{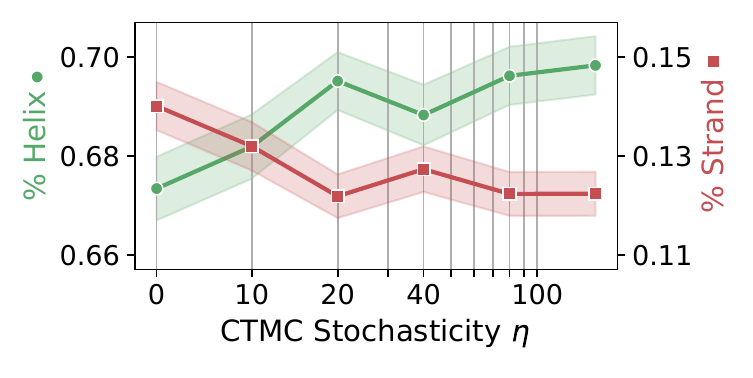}
    \vspace{-0.3cm}
    \caption{\textbf{\protmodel structural properties.} Average proportion of residues that are part of an alpha helix or beta strand versus the CTMC stochasticity level.
    Proportions of helices or strands can be desirable based on the family of proteins to generate \citep{vinothkumar2010structures}.
    Error bars show the standard error.
    }\label{fig:noise_vs_helixstrand}
    \vspace{-0.45cm}
\end{figure}

\textbf{Crossmodal modulation.}
We next investigate how modulating the CTMC stochasticity of the sequence affects the structural properties of sampled proteins.
\cref{fig:noise_vs_helixstrand} shows that varying the stochasticity level $\noise$ results in a change of the secondary structure composition \citep{kabsch1983dictionary} of the sampled proteins.
This is an example of the flexibility our multimodal framework provides to tune properties between data modalities at inference time.

\subsubsection{Forward and Inverse Folding}
\protmodel can achieve state-of-the-art codesign performance, but can accomplish more tasks as described in \cref{fig:overview}B and \cref{tab:sampling}.
Expanding \protmodel to achieve competitive performance on all tasks is a future work.
Here, we take the same model weights for co-design and evaluate forward and inverse folding \emph{without additional training}.
We compare performance to ESMFold and ProteinPMNN which are specialized models for forward and inverse folding.
We curated a clustered test-out set of 449 monomeric proteins with length $<400$ from the PDB using a date split of our training set. 
Details of forward/inverse folding and these experiments can be found in \cref{sec:apdx_protein_codesign_details}.
We find \protmodel can achieve very close performance with ProteinMPNN while it achieves poor results compared to ESMFold.
This highlights a limitation that \protmodel cannot perform competitively at every generation task, but leaves exciting future work for a potential general-purpose generative model.
\vspace{-0.5cm}
\begin{table}[h]
  \centering
  \caption{Forward and inverse folding: mean $\pm$ std.}
  \begin{tabular}{l|c|c}
    \toprule
     & Inverse folding & Forward folding \\
    Method & scRMSD ($\downarrow$) & RMSD ($\downarrow$) \\
    \midrule
    ProteinMPNN & 1.9 $\pm$ 2.7 & N/A \\
    ESMFold & N/A  & 2.7 $\pm$ 3.9 \\
    \protmodel & 2.2 $\pm$ 2.6 & 15.3 $\pm$ 4.5\\
    \bottomrule
  \end{tabular}
  \label{tab:sampling}
\end{table}
\vspace{-0.5cm}

\section{Discussion}
\label{sec:discussion}
We presented Discrete Flow Models (DFMs), a flow based generative model framework by making analogy to continuous state space flow models.
Our formulation is simple to implement, removes limitations in defining corruption processes, and provides more sampling flexibility for improved performance compared to previous discrete diffusion models.
Our framework enables easy application to multimodal generative problems which we apply to protein co-design.
The combination of a \method and FrameFlow enables state-of-the-art co-design with \protmodel.
Future work includes to develop more domain specific models with DFMs and improve \protmodel's performance on all protein generation tasks including sidechain modeling.

\section*{Acknowledgments}
The authors would like to thank Ricardo Baptista, Mathieu Le Provost, George Deligiannidis, Joe Benton, Bowen Jing, Hannes Stärk, Emile Mathieu, Luhuan Wu, Timur Garipov, Rachel Wu, Mingyu Choi, Sidney Lisanza, and Woody Ahern for helpful discussions.
AC acknowledges support from the EPSRC CDT in Modern Statistics and Statistical Machine Learning (EP/S023151/1)
JY was supported in part by an NSF-GRFP.
JY, RB, and TJ acknowledge support from NSF Expeditions grant (award 1918839: Collaborative Research: Understanding the World Through Code), Machine Learning for Pharmaceutical Discovery and Synthesis (MLPDS) consortium, the Abdul Latif Jameel
Clinic for Machine Learning in Health, the DTRA Discovery of Medical Countermeasures Against New and Emerging (DOMANE) threats program, the DARPA Accelerated Molecular Discovery program and the Sanofi Computational Antibody Design grant.  IF is  supported by the Office of Naval Research, the Howard Hughes Medical Institute (HHMI), and NIH (NIMH-MH129046).
The authors would like to acknowledge the use of the University of Oxford Advanced Research Computing (ARC) facility in carrying out this work. \url{http://dx.doi.org/10.5281/zenodo.22558}.

\section*{Impact statement}
In this paper we work to advance general purpose generative modeling techniques, specifically those used for modeling discrete and multimodal data. We apply these techniques to the task of protein generation. Improving protein modeling capabilities can have wide ranging societal impacts and care must be taken to ensure these impacts are positive. For example, improved modeling capabilities can help design better enzymes and drug candidates that can then go on to improve the lives of many people. Conversely, these general purpose techniques could also be misused to design toxic substances. To mitigate these risks, we do not present any specific methods to apply \protmodel to tasks that could be easily adjusted to the design of harmful substances without expert knowledge.

\bibliography{references}
\bibliographystyle{icml2024}

\newpage
\appendix
\onecolumn
\appendixhead

\section{Organization of Appendix}

The Appendix is organized as follows.
\cref{sec:apdx_proofs} provides proofs for all propositions in the main text.
\cref{sec:apdx_analysis_of_training_objective} analyses the cross entropy objective used to train \method and links controlling the cross entropy to controlling the model log-likelihood.
\cref{sec:apdx_related_work} discusses further related work.
\cref{sec:factorization} shows how \method can be applied to multidimensional data through applying factorization assumptions to $\noisemarg$.
\cref{sec:apdx_implementation_details} gives concrete realizations with PyTorch code for \method using the masking or uniform forms for $\noisemarg$.
\cref{sec:apdx_ctmc_sampling} discusses methods for sampling from CTMCs and discusses their relation to our sampling method.
\cref{sec:apdx_comparison_diff} compares \method to classical discrete diffusion models in discrete and continuous time finding that they can be fit within the \method framework.
\cref{sec:apdx_text_experiment_details} gives further details and results for our text experiment.
\cref{sec:apdx_protein_codesign_details} gives further details and results for our protein co-design experiments.

\section{Proofs}
\label{sec:apdx_proofs}

\paragraph{Notation}
When writing rate matrices, $R_t(i, j)$, we will assume $i \neq j$ unless otherwise explicitly stated.

We write $R_t(i) \vcentcolon = \sum_{j \neq i} R_t(i, j)$.

\subsection{Proof of Proposition \ref{prop:unconditional_generative_rate}}
We simply take the expectation with respect to $\pdata$ of both sides of the Kolmogorov equation for $\noisemarg(\x_t | \x_1)$ and $R_t(\x_t, j | \x_1)$. Note we use the fact that $R_t(i, i) = - \sum_{j \neq i} R_t(i, j)$ for compactness.  
    \begin{align}
    \partial_t \noisemarg(\x_t | \x_1) &= \sum_j R_t (j, \x_t | \x_1) \noisemarg(j | \x_1) \\
    \E_{\pdata(\x_1)} \left[ \partial_t \noisemarg( \x_t | \x_1) \right] &= \E_{\pdata(\x_1)} \left[ \sum_j R_t(j, \x_t | \x_1) \noisemarg(j | \x_1) \right]\\
    \partial_t \E_{\pdata(\x_1)} \left[ \noisemarg(\x_t | \x_1) \right] &= \sum_j \sum_{\x_1} \pdata(\x_1) \noisemarg(j | \x_1) R_t(j, \x_t | \x_1)\\
    \partial_t p_t(\x_t) &= \sum_j \sum_{x_1} p_t(j) \denoise(\x_1 | j) R_t(j, \x_t | \x_1)\\
    \partial_t p_t(\x_t) &= \sum_j   \E_{\denoise(\x_1 | j)} \left[R_t(j, \x_t | \x_1)\right]  p_t(j)
\end{align}
Where we notice that the final line is the Kolmogorov equation for a CTMC with marginals $p_t(\x_t)$ and rate $\E_{\denoise(\x_1 | \x_t)}\left[ R_t(\x_t, j | \x_t) \right]$. Therefore we have shown that $\E_{\denoise(\x_1 | \x_t)}\left[ R_t(\x_t, j | \x_t) \right]$ generate $p_t(\x_t)$.

\subsection{Proof of Proposition \ref{prop:relu_rate}}
\label{sec:rate_match_marginal_proof}
In the main text we provided the form for $\relurate_t$ under the assumption that $\noisemarg(j | \x_1) > 0$ for all $j$. Before proving \cref{prop:relu_rate}, we first give the full form for $\relurate_t$. First, assuming $\x_t \neq j$ and $\noisemarg(\x_t | \x_1) > 0$ we have,
\begin{align}
    \relurate_t(\x_t, j | \x_1) \vcentcolon = \frac{\relu \left( \partial_t \noisemarg(j | \x_1) - \partial_t \noisemarg(\x_t | \x_1) \right)}{ \mathcal{Z}_t \noisemarg(\x_t | \x_1) }
    \label{eq:relu_rate}
\end{align}
where $\relu(a) = \text{max}(a, 0)$ and $\mathcal{Z}_t$ is the number of states that have non-zero mass, $\mathcal{Z}_t = | \{ \x_t : \noisemarg(\x_t | \x_1) > 0 \} |$. $\relurate_t(\x_t, j | \x_1) = 0$ when $\noisemarg(\x_t | \x_1) = 0$ or $\noisemarg(j | \x_1) = 0$.
When $\x_t = j$, $\relurate_t(\x_t, \x_t | \x_1) = - \sum_{j \neq \x_t} \relurate_t(\x_t, j | \x_1)$ as we have defined before.

For our proof, we assume that $\noisemarg(j | \x_1) = 0 \implies \partial_t \noisemarg(j | \x_1)=0$. This assumption means that when we have dead states with zero probability mass, they cannot be resurrected and gain probability mass in the future.
We begin the proof with the Kolmogorov equation for processes conditioned on $\x_1$,
\begin{equation}
    \partial_t \noisemarg(\x_t | \x_1) = \sum_{j \neq \x_t} R_t(j, \x_t | \x_1) \noisemarg(j | \x_1) - \sum_{j \neq \x_t} R_t(\x_t, j | \x_1) \noisemarg(\x_t | \x_1)
    \label{eq:conditional_continuity}
\end{equation}
We will now verify that $\relurate_t$ satisfies this Kolmogorov equation and thus generates the desired $\noisemarg(\x_t | \x_1)$ conditional flow. We will first check that the Kolmogorov equation is satisfied when $\noisemarg(\x_t | \x_1) > 0$. With this form of rate matrix, the RHS of equation \eqref{eq:conditional_continuity} becomes
\begin{align}
    \text{RHS} &= \sum_{j \neq \x_t, \noisemarg(j | \x_1) > 0} \frac{ \relu \left( \partial_t \noisemarg(\x_t | \x_1) - \partial_t \noisemarg(j | \x_1) \right)}{\mathcal{Z}_t \noisemarg(j | \x_1) } \noisemarg(j | \x_1) \\
    & \qquad - \sum_{j \neq \x_t, \noisemarg(j | \x_1) > 0} \frac{\relu \left( \partial_t \noisemarg(j | \x_1) - \partial_t \noisemarg(\x_t | \x_1) \right)}{\mathcal{Z}_t \noisemarg(\x_t | \x_1)} \noisemarg(\x_t | \x_1)\\
    &= \frac{1}{\mathcal{Z}_t} \sum_{j \neq \x_t, \noisemarg(j | \x_1) > 0} \relu \left( \partial_t \noisemarg(\x_t | \x_1) - \partial_t \noisemarg(j | \x_1) \right) \\
    & \qquad - \frac{1}{\mathcal{Z}_t} \sum_{j \neq \x_t, \noisemarg(j | \x_1) > 0} \relu \left( \partial_t \noisemarg(j | \x_1) - \partial_t \noisemarg(\x_t | \x_1) \right)\\
    &= \frac{1}{\mathcal{Z}_t} \sum_{j \neq \x_t, \noisemarg(j | \x_1) > 0} \left( \partial_t \noisemarg(\x_t | \x_1) - \partial_t \noisemarg(j | \x_1) \right)\\
    &= \frac{\mathcal{Z}_t - 1}{\mathcal{Z}_t} \partial_t \noisemarg(\x_t | \x_1) - \frac{1}{\mathcal{Z}_t} \sum_{j \neq \x_t, \noisemarg(j | \x_1) > 0} \partial_t \noisemarg(j | \x_1)\\
    &= \frac{\mathcal{Z}_t - 1}{\mathcal{Z}_t} \partial_t \noisemarg(\x_t | \x_1) - \frac{1}{\mathcal{Z}_t} \partial_t ( 1 - \noisemarg(\x_t | \x_1))\\
    &= \frac{\mathcal{Z}_t - 1}{\mathcal{Z}_t} \partial_t \noisemarg(\x_t | \x_1) + \frac{1}{\mathcal{Z}_t} \partial_t \noisemarg(\x_t | \x_1)\\
    &= \partial_t \noisemarg(\x_t | \x_1) \\
    &= \text{LHS}
\end{align}
In the case that $\noisemarg(\x_t | \x_1) = 0$ by assumption we have that $\partial_t \noisemarg(\x_t | \x_1) = 0$. We have both $\relurate_t(\x_t, j | \x_1) = 0$ and $\relurate_t(j, \x_t | \x_1) = 0$ because $\noisemarg(\x_t | \x_1)=0$. Therefore we have $\text{LHS} = \text{RHS} = 0$ and thus the Kolmogorov equation is satisfied.

Intuitively, we require the assumption that dead states cannot be resurrected because $\relurate_t$ is designed such that all states can equally distribute the mass flux requirements of making sure the marginal derivatives $\partial_t \noisemarg(\x_t | \x_1)$ are satisfied. If there is a state for which $\noisemarg(\x_t | \x_1) = 0$ but $\partial_t \noisemarg(\x_t | \x_1) > 0$ then this state would require mass from other states but could not provide any mass of its own since $\noisemarg(\x_t | \x_1) = 0$. This would then violate the sharing symmetry required for our form of $\relurate_t$.  We note that this assumption is not strictly satisfied for the masking interpolant at $t=0$ or $t=1$ and not satisfied for the uniform interpolant at $t=1$. However, it is satisfied for any $t \in (0, 1)$ and so we can conceptualize starting our process at $t=\epsilon$, $\epsilon \ll 1$, $\epsilon > 0$, approximating a sample from $p_\epsilon(\x_\epsilon)$ with a sample from $p_0(\x_0)$ and running the process until $t = 1 - \epsilon$ and stopping here. The approximation can be made arbitrarily accurate by taking $\epsilon \rightarrow 0$.

\subsection{Proof of Proposition \ref{prop:db_rate}}
\label{sec:detailed_balance_proof}

A rate matrix that satisfies the detailed balance condition \eqref{eq:detailed_balance} will result in $\partial_t \noisemarg(i | \x_1) = 0$ when simulating with this rate. This can be seen by substituting into the conditional Kolmogorov equation \eqref{eq:conditional_continuity}
\begin{align}
    \partial_t \noisemarg(\x_t | \x_1) = &\sum_{j \neq \x_t} \dbrate_t(j, \x_t | \x_1) \noisemarg(j | \x_1) \\
    &- \sum_{j \neq \x_t} \dbrate_t(\x_t, j | \x_1) \noisemarg(\x_t | \x_1)\\
    \partial_t \noisemarg(\x_t | \x_1) = &\sum_{j \neq \x_t} \dbrate_t(\x_t, j | \x_1) \noisemarg(\x_t | \x_1) \\
    &- \sum_{j \neq \x_t} \dbrate_t(\x_t, j | \x_1) \noisemarg(\x_t | \x_1)\\
    \partial_t \noisemarg(\x_t | \x_1) = & 0
\end{align}
Given a rate matrix $R_t(\x_t, j | \x_1)$ that generates $\noisemarg(\x_t | \x_1)$, we first prove that $R_t(\x_t, j | \x_1) + \noise \dbrate_t(\x_t, j | \x_1)$ also generates $\noisemarg(\x_t | \x_1)$ for any $\noise \in \mathbb{R}^{\geq 0}$.
We show this by verifying that the combined rate matrix satisfies the Kolmogorov equation for conditional flow $\noisemarg(\x_t | \x_1)$.
The right hand side of the Kolmogorov equation is
\begin{align}
    \text{RHS} &= \sum_{j} \left(R_t(\x_t, j | \x_1) + \noise \dbrate_t(\x_t, j | \x_1) \right) \noisemarg(j | \x_1)\\
    &= \sum_j R_t(\x_t, j | \x_1) \noisemarg(j | \x_1) + \noise  \underbrace{\sum_j \dbrate_t(\x_t, j | \x_1) \noisemarg(j | \x_1)}_{=0}\\
    &= \sum_j R_t(\x_t, j \ \x_1) \noisemarg(j | \x_1)\\
    &= \partial_t \noisemarg(\x_t | \x_1)\\
    &= \text{LHS}
\end{align}
where we have used the fact that $\dbrate$ is in detailed balance with $\noisemarg(j | \x_1)$ and that $R_t(\x_t, j | \x_1)$ generates $\noisemarg$. Since $\relurate_t$ is a matrix that generates $\noisemarg$, we also have the stated result as a specific case: $\relurate_t + \noise \dbrate_t$ generates $\noisemarg$.

\subsection{Proof of Proposition \ref{prop:Roptimality}}
\label{sec:rate_optimality_proof}
We will assume we have $D$ dimensional data $\x_1^{1:D}$ with each $\x_1^d \in \{1, \dots, \statespace \}$. We give an overview of how our method operates in the multi-dimensional case in Appendix \ref{sec:factorization}. Namely, we assume that our conditional flow factorizes as $\noisemarg(\x_t^{1:D} | \x_1^{1:D}) = \prod_{d=1}^D \noisemarg(\x_t^d | \x_1^d)$. We also assume that our rate matrix is $0$ for jumps that vary more than $1$ dimension at a time. Our optimality results are derived under these assumptions.

\subsubsection{Masking Interpolant}
We first prove that $\relurate_t$ achieves the minimum number of transitions for the masking interpolant case. We have
\begin{equation}
    \noisemarg(\x_t^{1:D} | \x_1^{1:D}) = \prod_{d=1}^D \noisemarg(\x_t^d | \x_1^{d})
\end{equation}
with
\begin{equation}
    \noisemarg(\x_t^d | \x_1^d) = t \kdelta{ \x_t^d}{\x_1^d } + (1-t) \kdelta{ \x_t^d}{M }
\end{equation}
Our rate in dimension $d$ is
\begin{equation}
    {\relurate_t}^d(\x_t^d, j^d | \x_1^d) =\begin{cases}
    \frac{\relu \left( \partial_t \noisemarg(j^d | \x_1^d) - \partial_t \noisemarg(\x_t^d | \x_1^d) \right)}{\mathcal{Z}_t^d \noisemarg(\x_t^d | \x_1^d) } & \text{ for } \noisemarg(\x_t^d | \x_1^d) > 0, \noisemarg(j^d | \x_1^d) > 0\\
    = 0 &\text{ otherwise}
    \end{cases}
\end{equation}
with $\mathcal{Z}_t^d = | \{ j^d : \noisemarg(j^d | \x_1^d) > 0 \} |$. Substituting in $\partial_t \noisemarg$ and $\noisemarg$ in the masking case gives
\begin{equation}
    {\relurate_t}^d(\x_t^d, j^d | \x_1^d) = \frac{1}{1-t} \kdelta{\x_t^d}{ M } \kdelta { j^d} {\x_1^d }
\end{equation}
We refer to Appendix \ref{sec:apdx_masking_example} for the details of this derivation. Since ${\relurate_t}^d$ depends only on $\x_t^d$, $j^d$ and $\x_1^d$ and not values in any other dimensions, each dimension propagates independently and we can consider each dimension in isolation. Consider the process for dimension $d$. The CTMC begins in state $\x_0^d = M$. We have ${\relurate_t}^d(\x_t^d = M, j^d | \x_1^d) = \frac{1}{1-t} \kdelta { j^d}{ \x_1^d }$. Therefore, the only possible next state that the process can jump to is $\x_1^d$. Once the process has jumped to $\x_1^d$, the rate then becomes ${\relurate_t}^d(\x_t^d = \x_1^d, j^d | \x_1^d) = 0$. We also know that the process must jump because $p_1(\x_t^d | \x_1^d) = \kdelta { \x_t^d}{ \x_1^d }$, $\x_1^d \neq M$ and we know our rate matrix traverses our desired marginals by Proposition \ref{prop:relu_rate}. Therefore, exactly one jump is made in dimension $d$. In total, our $D$ dimensional process will make $D$ jumps. Under our factorization assumption, during a jump no more than one dimension can change value. Therefore, the absolute minimum number of jumps for any process that starts at $\x_0^{1:D}$ with $\x_0^d = M, \forall d$ and ends at $\x_1^{1:D}$, $\x_1^d \neq M, \forall d$ is $D$. Our prior distribution is $p_0(\x_0^d) = \kdelta { \x_0^d}{ M } $ and so for any $\x_0$ sample, we will always need to make $D$ jumps. Therefore, the minimum expected number of jumps is $D$ and $\relurate_t$ achieves this minimum.

\subsubsection{Uniform Interpolant}
We now prove that $\relurate_t$ achieves the minimum number of transitions for the uniform interpolant case. The conditional flow is 
\begin{equation}
    \noisemarg(\x_t^d | \x_1^d) = t \kdelta { \x_t^d}{ \x_1^d } + (1-t) \frac{1}{\statespace}
\end{equation}
With this interpolant, our rate matrix becomes
\begin{equation}
    {\relurate_t}^d(\x_t^d, j^d | \x_1^d) = \frac{1}{1-t} \kdelta { j^d}{ \x_1^d } \left( 1 - \kdelta { \x_t^d}{\x_1^d } \right) 
\end{equation}
We refer to Appendix \ref{sec:apdx_uniform_example} for the derivation. As before, ${\relurate_t}^d$ depends only on the values in dimension $d$, $\x_t^d, j^d, \x_1^d$ and therefore each process propagates independently in each dimension and we can consider each dimension in isolation. Considering dimension $d$, the process begins in state $\x_0^d$. Both $\x_0^d = \x_1^d$ and $\x_0^d \neq \x_1^d$ are possible in the uniform interpolant case. In the case that $\x_0^d = \x_1^d$, then ${\relurate_t}^d = 0$ for all $t$ and therefore no jumps are made in this dimension. In the case that $\x_0^d \neq \x_1^d$ then before any jump is made we have ${\relurate_t}^d(\x_t^d, j^d | \x_1^d) = \frac{1}{1-t} \kdelta { j^d}{ \x_1^d }$ and so the only possible next state the process can jump to is $\x_1^d$. Once the process has jumped to $\x_1^d$, the rate then becomes ${\relurate_t}(\x_t^d = \x_1^d, j^d | \x_1^d) = 0$ and so no more jumps are made. We also know that the process must jump at some point because $p_1(\x_t^d | \x_1^d) = \kdelta { \x_t^d}{ \x_1^d }$ and we know our rate matrix traverses our desired marginals by Proposition \ref{prop:relu_rate}. Therefore, in the case that $\x_0^d \neq \x_1^d$, exactly one jump is made for the process in dimension $d$. In total, the number of jumps made in all $D$ dimensions is $d_H(\x_0, \x_1) = |\{d: \x_0^d \neq \x_1^d \} |$ which is the Hamming distance between $\x_0$ and $\x_1$. The expected number of jumps for our process with $\relurate_t$ is thus $\E_{p_0(\x_0) \pdata(\x_1)} \left[ d_H(\x_0, \x_1) \right]$.

Now consider an optimal process that makes the minimum number of jumps when starting from $\x_0$ and meets our factorization assumptions. By this assumption, during a jump only one dimension can change in value. Clearly we have that the minimum number of jumps required to get from $\x_0$ to $\x_1$ is $d_H(\x_0, \x_1)$. Therefore, for this optimal process we also have that the minimum number of expected jumps is $\E_{p_0(\x_0) \pdata(\x_1)}\left[d_H(\x_0, \x_1) \right]$. Therefore, $\relurate_t$ achieves the minimum expected number of jumps.

\subsubsection{Discussion}
We have proven $\x_1$ conditioned optimality only for the two simple conditional flows featured in the main text and we note that this result in not generally true for any conditional flow. Intuitively this is because $\relurate_t$ treats the distribution of mass symmetrically between states, considering only the local differences in $\partial_t \noisemarg$ between pairs of states. In general, the optimal rate would need to solve a global programming problem.

We also note that although we have masking and uniform optimality for $\relurate_t(\x_t, j | \x_1)$ when conditioned on $\x_1$, this is not necessarily the case when we consider the unconditional version $\E_{\denoise(\x_1 | \x_t)} \left[ \relurate_t(\x_t, j | \x_1) \right]$. There may exist rate matrices that achieve a lower number of average jumps and successfully pass through the unconditional marginals $p_t(\x_t) = \E_{\pdata(\x_1)} \left[ \noisemarg(\x_t | \x_1) \right]$. This is analogous to continuous flow-based methods which can create optimal straight-line paths when conditioned on the end point $\x_1$, but don't necessarily achieve the optimal transport when considering the unconditional vector field \cite{shaul2023kinetic}.

\subsection{Proof of Proposition \ref{prop:multimodal_consistency}}
We have that the individual velocities and rate matrices independently generate their respective flows in each dimension. Specifically, for $\trans_t^d \in \mathbb{R}^3$, we have that it satisfies the following Fokker-Planck equation,
\begin{equation}
    \partial_t p_{t|1}(\trans_t^d | \trans_1^d) = - \nabla^{(d)} \cdot \left( v_\trans^d(\trans_t^d | \trans_1^d) p_{t|1}(\trans_t^d | \trans_1^d) \right).
\end{equation}
where $\nabla^{(d)} \cdot$ is the divergence operator for elements in dimension $d$. Similarly for $\rots_t \in \SO$,
\begin{equation}
    \partial_t p_{t|1}(\rots_t^d | \rots_1^d) = - \nabla^{(d)} \cdot \left( v_\rots^d (\rots_t^d | \rots_1^d) p_{t|1}(\rots_t^d | \rots_1^d) \right),
\end{equation}
where the divergence operator now acts on elements in $\text{Tan}_{\rots_t^d} \SO$ \cite{yim2023se}. Finally, for $\amino_t^d$, we have the familiar Kolmogorov equation,
\begin{equation}
    \partial_t p_{t|1}(\amino_t^d | \amino_1^d) = \sum_j R_t(j, \amino_t^d | \amino_1^d) p_{t|1}^d(j | \amino_1^d).
\end{equation}
For the joint space $ \allresidue \in ( \mathbb{R}^3 \times \SO \times \{1, \dots, 20, M\} )^D$ and process defined by the updates in \cref{eq:conditional_multimodal_update}, we also have a joint continuity equation known as the Fokker-Planck-Kolmogorov equation \cite{bect2010unifying},
\begin{align}
    \partial_t p_{t|1}(\allresidue_t | \allresidue_1) =& - \nabla \cdot \left( v_\trans(\trans_t^{1:D} | \trans_1^{1:D}) p_{t|1}(\allresidue_t | \allresidue_1) \right) - \nabla \cdot \left( v_{\rots}(\rots_t^{1:D} | \rots_1^{1:D}) p_{t|1}(\allresidue_t | \allresidue_1) \right) +\\
    &\sum_{j^{1:D}} R_t(j^{1:D}, \amino_t^{1:D} | \amino_1^{1:D} ) p_{t|1}\big( ( \trans_t^{1:D}, \rots_t^{1:D}, j^{1:D} ) | \allresidue_1 \big).
    \label{eq:fokker-planck-kolmogorov}
\end{align}
Our aim is to show that the following choices of $p_{t|1}$, $v_\trans$, $v_\rots$ and $R_t$ corresponding to independent processes within each modality and dimension are consistent under \cref{eq:fokker-planck-kolmogorov} i.e. these choices of $v_\trans$, $v_\rots$ and $R_t$ will actually generate $p_{t|1}$ when simulated using \cref{eq:conditional_multimodal_update} with $\Delta t \rightarrow 0$. The choices are as follows:
\begin{align}
    p_{t|1}(\allresidue_t | \allresidue_1) &= \prod_{d=1}^D p_{t|1}(\trans_t^d | \trans_1^d) p_{t|1}(\rots_t^d | \rots_1^d) p_{t|1}(\amino_t^d | \amino_1^d)\\
    v_\trans(\trans_t^{1:D} | \trans_1^{1:D}) &= \begin{bmatrix}
        v_\trans^1(\trans_t^1 | \trans_1^1)\\
        \vdots \\
        v_\trans^D(\trans_t^D | \trans_1^D)
    \end{bmatrix}\\
    v_\rots(\rots^{1:D} | \rots^{1:D}) &= \begin{bmatrix}
        v_\rots^1(\rots_t^1 | \rots_1^1)\\
        \vdots \\
        v_\rots^D(\rots_t^D | \rots_1^D)
    \end{bmatrix}\\
    R_t(j^{1:D}, \amino_t^{1:D} | \amino_1^{1:D}) &= \sum_{d=1}^D \kdelta{j^{1:D \backslash d}}{\amino_t^{1:D \backslash d}} R_t^d(j^d, \amino_t^d | \amino_1^d)
\end{align}
More discussion regarding the form of $R_t(j^{1:D}, \amino_t^{1:D} | \amino_1^{1:D}) $ can be found in Appendix \ref{sec:factorization}. Under these choices, the LHS of \cref{eq:fokker-planck-kolmogorov} becomes
\begin{align}
    \text{LHS} &= \partial_t p_{t|1}(\allresidue_t | \allresidue_1)\\
    &= \sum_{d=1}^D \partial_t p_{t|1}(\trans_t^d | \trans_1^d)\frac{ p_{t|1}(\allresidue_t | \allresidue_1)}{p_{t|1}(\trans_t^d | \trans_1^d)} + \partial_t p_{t|1}(\rots_t^d | \rots_1^d)\frac{ p_{t|1}(\allresidue_t | \allresidue_1)}{p_{t|1}(\rots_t^d | \rots_1^d)} + \partial_t p_{t|1}(\amino_t^d | \amino_1^d)\frac{ p_{t|1}(\allresidue_t | \allresidue_1)}{p_{t|1}(\amino_t^d | \amino_1^d)}
\end{align}
by the product rule for differentiation. The RHS of \cref{eq:fokker-planck-kolmogorov} becomes
\begin{align}
    \text{RHS} &= \sum_{d=1}^D - \nabla^{(d)} \cdot \left( v_\trans^d (\trans_t^d | \trans_1^d) p_{t|1}(\allresidue_t | \allresidue_1) \right) - \nabla^{(d)} \cdot \left( v_\rots^d (\rots_t^d | \rots_1^d) p_{t|1}(\allresidue_t | \allresidue_1) \right) \\
    & \quad + \sum_{j^{1:D}} \sum_{d=1}^D \kdelta{ j^{1:D \backslash d}}{\amino_t^{1:D \backslash d}} R_t^d(j^d, \amino_t^d | \amino_1^d) p_{t|1}\big( (\trans_t^{1:D}, \rots_t^{1:D}, j^{1:D} | \allresidue_1 \big)\\
    &= \sum_{d=1}^D - \nabla^{(d)} \cdot \left( v_\trans^d (\trans_t^d | \trans_1^d) p_{t|1}(\trans_t^d | \trans_1^d) \right) \frac{p_{t|1}(\allresidue_t | \allresidue_1)}{p_{t|1}(\trans_t^d | \trans_1^d) } - \nabla^{(d)} \cdot \left( v_\rots^d (\rots_t^d | \rots_1^d) p_{t|1}(\rots_t^d | \rots_1^d) \right) \frac{p_{t|1}(\allresidue_t | \allresidue_1)}{p_{t|1}(\rots_t^d | \rots_1^d) } \\
    & \quad + \sum_{d=1}^D \Bigg[ \left\{ \prod_{d' =1}^D p_{t|1}^{d'}(\trans_t^{d'} | \trans_1^{d'}) p_{t|1}^{d'}(\rots_t^{d'} | \rots_1^{d'}) \right\} \sum_{j^d} R_t^d(j^d, \amino_t^d | \amino_1^d) p_{t|1}(j^d | \amino_1^d) \\
    &\hspace{1cm} \sum_{j^{1:D \backslash d}} \kdelta{ j^{1:D \backslash d}}{\amino_t^{1:D \backslash d}} \left\{ \prod_{d'=1 \backslash d}^D p_{t|1}(j^{d'} | \amino_1^{d'}) \right\} \Bigg] \\
    &= \sum_{d=1}^D \partial_t p_{t|1}(\trans_t^d | \trans_1^d) \frac{p_{t|1}(\allresidue_t | \allresidue_1)}{p_{t|1}(\trans_t^d | \trans_1^d)} + \partial_t p_{t|1}(\rots_t^d | \rots_1^d) \frac{ p_{t|1}(\allresidue_t | \allresidue_1)}{p_{t|1}(\rots_t^d | \rots_1^d) } \\
    & \quad + \sum_{d=1}^D \left\{ \prod_{d' =1}^D p_{t|1}^{d'}(\trans_t^{d'} | \trans_1^{d'}) p_{t|1}^{d'}(\rots_t^{d'} | \rots_1^{d'}) \right\} \sum_{j^d} R_t^d(j^d, \amino_t^d | \amino_1^d) p_{t|1}(j^d | \amino_1^d) \left\{ \prod_{d'=1 \backslash d}^D p_{t|1}(\amino_t^{d'} | \amino_1^{d'}) \right\}\\
    &= \sum_{d=1}^D \partial_t p_{t|1}(\trans_t^d | \trans_1^d) \frac{p_{t|1}(\allresidue_t | \allresidue_1)}{p_{t|1}(\trans_t^d | \trans_1^d)} + \partial_t p_{t|1}(\rots_t^d | \rots_1^d) \frac{ p_{t|1}(\allresidue_t | \allresidue_1)}{p_{t|1}(\rots_t^d | \rots_1^d) } + \frac{p_{t|1}(\allresidue_t | \allresidue_1)}{p_{t|1}(\amino_t^d | \amino_1^d) } \sum_{j^d} R_t^d(j^d, \amino_t^d | \amino_1^d) p_{t|1}(j^d | \amino_1^d) \\
    &= \sum_{d=1}^D \partial_t p_{t|1}(\trans_t^d | \trans_1^d) \frac{p_{t|1}(\allresidue_t | \allresidue_1)}{p_{t|1}(\trans_t^d | \trans_1^d)} + \partial_t p_{t|1}(\rots_t^d | \rots_1^d) \frac{ p_{t|1}(\allresidue_t | \allresidue_1)}{p_{t|1}(\rots_t^d | \rots_1^d) } + \partial_t p_{t|1}(\amino_t^d | \amino_1^d) \frac{p_{t|1}(\allresidue_t | \allresidue_1)}{p_{t|1}(\amino_t^d | \amino_1^d) } \\
    &= \text{LHS}
\end{align}
Therefore we have shown that our choices of $v_\trans$, $v_\rots$ and $R_t$ will generate the desired $p_{t|1}$.

\subsection{Proof of Proposition \ref{prop:multimodal_unconditional_flow}}
Our proof will mirror that of Proposition \ref{prop:unconditional_generative_rate} by taking the expectation with respect to $\pdata(\allresidue_1)$ of both sides of the Fokker-Planck-Kolmogorov equation (\cref{eq:fokker-planck-kolmogorov}).
\begin{align}
    \E_{\pdata(\allresidue_1)} \left[ \partial_t p_{t|1}(\allresidue_t | \allresidue_1) \right] =& \E_{\pdata(\allresidue_1)} \Big[ - \nabla \cdot \left( v_\trans(\trans_t^{1:D} | \trans_1^{1:D}) p_{t|1}(\allresidue_t | \allresidue_1) \right) - \nabla \cdot \left( v_{\rots}(\rots_t^{1:D} | \rots_1^{1:D}) p_{t|1}(\allresidue_t | \allresidue_1) \right) +\\
    &\sum_{j^{1:D}} R_t(j^{1:D}, \amino_t^{1:D}) p_{t|1}\big( ( \trans_t^{1:D}, \rots_t^{1:D}, j^{1:D} ) | \allresidue_1 \big) \Big]\\
    \partial_t p_t(\allresidue_t) =& \sum_{d=1}^D \Bigg\{ \E_{\pdata(\allresidue_1)} \left[ - \nabla^{(d)} \cdot \left( v_\trans^d ( \trans_t^d | \trans_1^d) p_{t|1}(\allresidue_t | \allresidue_1) \right) \right] + \\
    & \qquad \E_{\pdata(\allresidue_1)} \left[ - \nabla^{(d)} \cdot \left( v_\rots^d(\rots_t^d | \rots_1^d) p_{t|1}(\allresidue_t | \allresidue_1) \right) \right] +\\
    & \qquad \E_{\pdata(\allresidue_1)} \left[ \frac{p_{t|1}(\allresidue_t | \allresidue_1)}{p_{t|1}(\amino_t^d | \amino_1^d) } \sum_{j^d} R_t^d(j^d, \amino_t^d | \amino_1^d) p_{t|1}(j^d | \amino_1^d) \right] \Bigg\} \label{eq:uncond_fpk_intermediate}
\end{align}
where on the second line on the left hand side we have used the fact that $p_t(\allresidue_t) = \E_{\pdata(\allresidue_1)} \left[ p_{t|1}(\allresidue_t | \allresidue_1) \right]$. We shall first examine the $v_\trans^d$ term on the right hand side in isolation.
\begin{align}
    \E_{\pdata(\allresidue_1)} &\left[ - \nabla^{(d)} \cdot \left( v_\trans^d ( \trans_t^d | \trans_1^d) p_{t|1}(\allresidue_t | \allresidue_1) \right) \right] = - \int_{\trans_1^{1:D}} \int_{\rots_1^{1:D}} \sum_{\amino_1^{1:D}} \pdata(\allresidue_1) \nabla^{(d)} \cdot \left( v_\trans^d(\trans_t^d | \trans_1^d) p_{t|1}(\allresidue_t | \allresidue_1) \right) \dd \trans_1^{1:D} \dd \rots_1^{1:D}\\
    &= - \nabla^{(d)} \cdot \left( \int_{\trans_1^{1:D}} \int_{\rots_1^{1:D}} \sum_{\amino_1^{1:D}} \pdata(\allresidue_1) v_\trans^d(\trans_t^d | \trans_1^d) p_{t|1}(\allresidue_t | \allresidue_1) \dd \trans_1^{1:D} \dd \rots_1^{1:D} \right)\\
    &= - \nabla^{(d)} \cdot \left( \int_{\trans_1^d} v_\trans^d(\trans_t^d | \trans_1^d) \int_{\trans_1^{1:D \backslash d}} \int_{\rots_1^{1:D}} \sum_{\amino_1^{1:D}} \pdata(\allresidue_1) p_{t|1}(\allresidue_t | \allresidue_1) \dd \trans_1^{1:D \backslash d} \dd \rots_1^{1:D} \dd \trans_1^d \right)\\
    &= - \nabla^{(d)} \cdot \left( \int_{\trans_1^d} v_\trans^d(\trans_t^d | \trans_1^d) p_t(\allresidue_t) p_{1|t}(\trans_1^d | \allresidue_t) \right)\\
    & = - \nabla^{(d)} \cdot \left( \E_{p_{1|t}(\trans_1^d | \allresidue_t)} \left[ v_\trans^d ( \trans_t^d | \trans_1^d) \right] p_t(\allresidue_t) \right)
\end{align}
The same argiments follow through for the $v_\rots^d$ term giving
\begin{equation}
    \E_{\pdata(\allresidue_1)} \left[ - \nabla^{(d)} \cdot \left( v_\rots^d(\rots_t^d | \rots_1^d) p_{t|1}(\allresidue_t | \allresidue_1) \right) \right] = - \nabla^{(d)} \cdot \left( \E_{p_{1|t}(\rots_1^d | \allresidue_t)} \left[v_\rots^d(\rots_t^d | \rots_1^d) \right] p_t(\allresidue_t) \right)
\end{equation}
We finally analyse the $R_t^d$ term in isolation. In the following, we will use $\amino_t^{1:D \backslash d} \odot j^d$ to refer to the $D$ dimensional discrete variable with the values $\amino_t^{1:D \backslash d}$ in all dimensions except $d$ and the value $j^d$ in dimension $d$. 
\begin{align}
    \E_{\pdata(\allresidue_1)} \Bigg[ & \frac{p_{t|1}(\allresidue_t | \allresidue_1)}{p_{t|1}(\amino_t^d | \amino_1^d) } \sum_{j^d} R_t^d(j^d, \amino_t^d | \amino_1^d) p_{t|1}(j^d | \amino_1^d) \Bigg] \\
    &= \int_{\trans_1^{1:D}} \int_{\rots_1^{1:D}} \sum_{\amino_1^{1:D}} \pdata(\allresidue_1) \frac{p_{t|1}(\allresidue_t | \allresidue_1)}{p_{t|1}(\amino_t^d | \amino_1^d)} \sum_{j^d} R_t^d(j^d, \amino_t^d | \amino_1^d) p_{t|1}(j^d | \amino_1^d) \dd \trans_1^{1:D} \dd \rots_1^{1:D} \\
    &= \int_{\trans_1^{1:D}} \int_{\rots_1^{1:D}} \sum_{\amino_1^{1:D}} \sum_{j^d} p_{t,1}\left( (\trans_t^{1:D}, \rots_t^{1:D}, \amino_t^{1:D \backslash d} \odot j^d), \allresidue_1 \right) R_t^d(j^d, \amino_t^d | \amino_1^d) \dd \trans_1^{1:D} \dd \rots_1^{1:D}\\
    &= \int_{\trans_1^{1:D}} \int_{\rots_1^{1:D}} \sum_{\amino_1^{1:D}} \sum_{j^d} p_t \left( (\trans_t^{1:D}, \rots_t^{1:D}, \amino_t^{1:D \backslash d} \odot j^d) \right) p_{1|t}\left(\amino_1^d | (\trans_t^{1:D}, \rots_t^{1:D}, \amino_t^{1:D \backslash d} \odot j^d)\right) \\
    & \hspace{3.3cm} p\left( (\trans_1^{1:D}, \rots_1^{1:D}, \amino_1^{1:D \backslash d} ) | \amino_1^d, (\trans_t^{1:D}, \rots_t^{1:D}, \amino_t^{1:D \backslash d} \odot j^d) \right) R_t^d(j^d, \amino_t^d | \amino_1^d) \dd \trans_1^{1:D} \dd \rots_1^{1:D}\\
    &= \sum_{j^d} p_t \left( ( \trans_t^{1:D}, \rots_t^{1:D}, \amino_t^{1:D \backslash d} \odot j^d)  \right) \sum_{\amino_1^d} p_{1|t}\left( \amino_1^d | ( \trans_t^{1:D}, \rots_t^{1:D}, \amino_t^{1:D \backslash d} \odot j^d) \right) R_t^d(j^d, \amino_t^d | \amino_1^d) \\
    & \hspace{1cm} \underbrace{\int_{\trans_1^{1:D}} \int_{\rots_1^{1:D}} \sum_{\amino_1^{1:D \backslash d}} p \left( ( \trans_1^{1:D}, \rots_1^{1:D}, \amino_1^{1:D \backslash d}) | \amino_1^d, (\trans_t^{1:D}, \rots_t^{1:D}, \amino_t^{1:D \backslash d} \odot j^d) \right) \dd \trans_1^{1:D} \rots_1^{1:D}}_{=1}\\
    &= \sum_{j^d} \E_{p_{1|t}(\amino_1^d | (\trans_t^{1:D}, \rots_t^{1:D}, \amino_t^{1:D \backslash d} \odot j^d) } \left[ R_t^d(j^d, \amino_t^d | \amino_1^d )\right] p_t \left( ( \trans_t^{1:D}, \rots_t^{1:D}, \amino_t^{1:D \backslash d} \odot j^d ) \right)
\end{align}
Now we substitute these simplified forms back into \cref{eq:uncond_fpk_intermediate}.
\begin{align}
    \partial_t p_t(\allresidue_t) =& \sum_{d=1}^D \Bigg\{ - \nabla^{(d)} \cdot \left( \E_{p_{1|t}(\trans_1^d | \allresidue_t)} \left[ v_\trans^d(\trans_t^d | \trans_1^d) \right] p_t(\allresidue_t) \right) - \nabla^{(d)} \cdot \left( \E_{p_{1|t}(\rots_1^d | \allresidue_t)} \left[ v_\rots^d(\rots_t^d | \rots_1^d) \right] p_t(\allresidue_t) \right) + \\
    &\sum_{j^d} \E_{p_{1|t}(\amino_1^d | (\trans_t^{1:D}, \rots_t^{1:D}, \amino_t^{1:D \backslash d} \odot j^d) } \left[ R_t^d(j^d, \amino_t^d | \amino_1^d )\right] p_t \left( ( \trans_t^{1:D}, \rots_t^{1:D}, \amino_t^{1:D \backslash d} \odot j^d ) \right) \Bigg\} \label{eq:unconditional_fpk_intermediate2} 
\end{align}
We will now show that \cref{eq:unconditional_fpk_intermediate2} is the Fokker-Planck-Kolmogorov equation for $p_t(\allresidue_t)$ with the following choices for the velocities and rate matrix.
\begin{align}
    v_\trans(\allresidue_t) &= \begin{bmatrix}
        \E_{p_{1|t}(\trans_1^1 | \allresidue_t)} \left[ v_\trans^1(\trans_t^1 | \trans_1^1) \right] \\
        \vdots\\
        \E_{p_{1|t}(\trans_1^D | \allresidue_t)} \left[ v_\trans^D(\trans_t^D | \trans_1^D) \right] \\
    \end{bmatrix} \\
    v_\rots(\allresidue_t) &= \begin{bmatrix}
        \E_{p_{1|t}(\rots_1^1 | \allresidue_t)} \left[ v_\rots^1(\rots_t^1 | \rots_1^1) \right] \\
        \vdots\\
        \E_{p_{1|t}(\rots_1^D | \allresidue_t)} \left[ v_\rots^D(\rots_t^D | \rots_1^D) \right] \\
    \end{bmatrix}\\
    R_t(\allresidue_t, j^{1:D}) &= \sum_{d=1}^D \kdelta{\amino_t^{1:D \backslash d}}{j^{1:D \backslash d}} \E_{p_{1|t}(\amino_1^d | \allresidue_t)} \left[ R_t^d(\amino_t^d, j^d | \amino_1^d) \right]
    \label{eq:multimodal_choices_uncond}
\end{align}
We shall substitute the choices in \cref{eq:multimodal_choices_uncond} into the Fokker-Planck-Kolmogorov equation for $p_t(\allresidue_t)$ and show that this equals \cref{eq:unconditional_fpk_intermediate2}.
\begin{align}
    \partial_t p_t(\allresidue_t) &= - \nabla \cdot \left( v_\trans(\allresidue_t) p_t(\allresidue_t) \right) - \nabla \cdot \left( v_\rots(\allresidue_t) p_t(\allresidue_t) \right) + \sum_{j^{1:D}} R_t\left( (\trans_t^{1:D}, \rots_t^{1:D}, j^{1:D}) , \amino_t^{1:D} \right) p_t \left( (\trans_t^{1:D}, \rots_t^{1:D}, j^{1:D}) \right)\\
    &= \sum_{d=1}^D \Bigg\{ - \nabla^{(d)} \cdot \left( \E_{p_{1|t}(\trans_1^d | \allresidue_t)} \left[ v_\trans^d(\trans_t^d | \trans_1^d) \right] p_t(\allresidue_t) \right) - \nabla^{(d)} \cdot \left( \E_{p_{1|t}(\rots_1^d | \allresidue_t) } \left[ v_\rots^d ( \rots_t^d | \rots_1^d) \right] p_t(\allresidue_t) \right) \Bigg\} + \\
    & \qquad \sum_{j^{1:D}} \sum_{d=1}^D \kdelta{j^{1:D \backslash d}}{\amino_t^{1:D \backslash d}} \E_{p_{1|t}(\amino_1^d | (\trans_t^{1:D}, \rots_t^{1:D}, j^{1:D}))} \left[ R_t^d(j^d, \amino_t^d | \amino_1^d)\right] p_t\left( ( \trans_t^{1:D}, \rots_t^{1:D}, j^{1:D}) \right)\\
    &= \sum_{d=1}^D \Bigg\{ - \nabla^{(d)} \cdot \left( \E_{p_{1|t}(\trans_1^d | \allresidue_t)} \left[ v_\trans^d(\trans_t^d | \trans_1^d) \right] p_t(\allresidue_t) \right) - \nabla^{(d)} \cdot \left( \E_{p_{1|t}(\rots_1^d | \allresidue_t)} \left[ v_\rots^d(\rots_t^d | \rots_1^d) \right] p_t(\allresidue_t) \right) + \\
    &\qquad\sum_{j^d} \E_{p_{1|t}(\amino_1^d | (\trans_t^{1:D}, \rots_t^{1:D}, \amino_t^{1:D \backslash d} \odot j^d) } \left[ R_t^d(j^d, \amino_t^d | \amino_1^d )\right] p_t \left( ( \trans_t^{1:D}, \rots_t^{1:D}, \amino_t^{1:D \backslash d} \odot j^d ) \right)  \Bigg\}
\end{align}
which we see matches \cref{eq:unconditional_fpk_intermediate2}. Therefore, we have shown that the choices for the velocities and rate matrix in \cref{eq:multimodal_choices_uncond} create a process that generates $p_t(\allresidue_t)$ as desired.

\section{Analysis of Training Objective}
\label{sec:apdx_analysis_of_training_objective}

In this section we analyse how our cross entropy objective $\mathcal{L}_\mathrm{ce}$ relates to the log-likelihood of the data under the generative model and to the ELBO used to train classical discrete diffusion models.

Our proof is structured as follows. We first introduce path space measures for CTMCs in Section \ref{sec:apdx_intro_ctmc_path_measures} that we will require for the rest of the derivation. In Section \ref{sec:apdx_derivation_Lelbo} we then derive the standard evidence lower bound, $\mathcal{L}_{\text{ELBO}}$ on the model log likelihood, $\E_{\pdata(\x_1)} \left[ \log p_\theta(\x_1) \right]$. We then decompose $\mathcal{L}_{\text{ELBO}}$ into the cross entropy, a rate regularizer and a KL term in Section \ref{sec:apdx_decomposition_of_Lelbo}. Finally in Section \ref{sec:apdx_Lelbo_for_masking} we show that $\mathcal{L}_{\text{ELBO}}$ corresponds exactly to the weighted cross entropy loss for the masking interpolant case.

\subsection{Introduction to CTMC path measures}
\label{sec:apdx_intro_ctmc_path_measures}
Before beginning the proof, we introduce path space measures for CTMC processes, following the exposition in \cite{del2017stochastic}, Chapter 18. A path of a CTMC is a single trajectory from time $0$ to time $t$. The trajectory is a function $\omega : s \in [0, t] \mapsto \omega_s \in \{1, \dots , \statespace \}$ that is everywhere right continuous and has left limits everywhere (also known as càdlàg paths). Intuitively, it is a function that takes in a time variable and outputs the position of the particle following the trajectory at that time. The càdlàg condition in our case states that at jump time $\tau$ we have $\omega_\tau$ taking the new jumped to value and $\omega_{\tau}^- \vcentcolon = \lim_{s \uparrow \tau} \omega_s$ being the previous value before the jump, see \cref{fig:overview}B.

A trajectory drawn from the CTMC, $W$, can be fully described through its jump times, $T_1, \dots T_n$ and its state values between jumps, $W_0, W_1, \dots, W_{T_n}$ where at jump time $T_k$ the CTMC jumps from state value $W_{k-1}$ to value $W_k$. A path space measure $\mathbb{P}$ is able to assign probabilities to a drawn trajectory $W$ from time $0$ to $t$ in the sense of
\begin{align}
    \mathbb{P}(W \in \dd \omega) \vcentcolon = \mathbb{P} \left( W_0 \in \dd \omega_{0}, (T_1, W_{T_1}) \in \dd (t_1, \omega_{t_1}), \dots (T_n, W_{T_n}) \in \dd (t_n, \omega_{t_n}), T_{n+1} \geq t \right)
\end{align}
where $\dd \omega_{t_n}$ and $\dd t_n$ denote infinitesimal neighborhoods around the points $\omega_{t_n} \in \{1, \dots, \statespace \}$ and $t_n \in [0, t]$. This is the same sense in which a probability density function assigns probabilities to the infinitesimal neighborhood around a continuous valued variable.

To understand the form of $\mathbb{P}(W \in \dd \omega)$ we remind ourselves of the definition of a CTMC with rate matrix $R_t$. The CTMC waits in the current state for an amount of time determined by an exponential random variable with time-inhomogeneous rate $R_t(W_t) \vcentcolon = \sum_{k \neq W_t} R_t(W_t, k)$, see \citet{norris1998markov} and \citet{campbell2022continuous} Appendix A for more details. After the wait time is finished, the CTMC jumps to a next chosen state where the jump distribution is 
\begin{equation}
    \mathbb{P}(W_{t_k} | W_{t_k}^-) = \frac{R_t(W_{t_k}^-, W_{t_k}) \left(1 - \kdelta{W_{t_k}^-}{W_{t_k} } \right)   } { R_t(W_{t_k}^-) }
\end{equation}
For an exponential random variable with time-inhomogeneous rate, the cumulative distribution function is given by
\begin{equation}
    \mathbb{P}(T < t) = 1 - \exp \left( - \int_{s=0}^{s=t} R_s(W_s^-) \dd s \right) 
\end{equation}
Therefore, the probability density function, $p(t) = \frac{\partial}{\partial t} \mathbb{P}(T < t)$, is
\begin{equation}
    p(t) =  \exp \left( - \int_{s=0}^{s=t} R_s(W_s^-) \dd s \right)R_t(W_t^-)
\end{equation}
We finally note that if we wish to know $\mathbb{P}(T_k < t | T_{k-1})$ i.e. the probability that the $k$-th jump time is less than $t$ given we know the $k-1$-th jump time, then this is just an exponential random variable started at time $T_{k-1}$ when the previous jump occurred,
\begin{equation}
    \mathbb{P}(T_k < t | T_{k-1}) = 1 - \exp \left( - \int_{s=T_{k-1}}^{s=t} R_s(W_s^-) \dd s \right)
\end{equation}
In other words, we simply start a new exponential timer once the previous jump occurs and the same equation carries through.

We can now write the form of $\mathbb{P}(W \in \dd \omega)$. We split it into a series of conditional distributions
\begin{align}
    \mathbb{P}(W \in \dd \omega) = &\mathbb{P}(W_0 \in \dd \omega_0) \mathbb{P}((T_1, W_{T_1} \in \dd (t_1, \omega_{t_1}) | W_0) \times \dots \\
    &\times \mathbb{P}((T_n, W_{T_n}) \in \dd (t_n, \omega_{t_n}) | W_0, (T_1, W_{T_1}), \dots, (T_{n-1}, W_{T_{n-1}})) \\
    &\times \mathbb{P}(T_{n+1} \geq t |  W_0, (T_1, W_{T_1}), \dots, (T_{n}, W_{T_{n}}))
\end{align}
\begin{align}
    \mathbb{P}(W \in \dd \omega) &= p_0(W_0) \exp \left(- \int_{s=0}^{s=T_1} R_s(W_s^-) \dd s \right) R_{T_1}(W_{T_1}^-) \mathbb{P}(W_{T_1} | W_{T_1}^- ) \times \dots \\
    &\quad \times \exp \left( - \int_{s=T_{n-1}}^{s=T_n} R_s(W_s^-) \dd s \right) R_{T_n}(W_{T_n}^-) \mathbb{P}(W_{T_n} | W_{T_n}^-) \exp \left( - \int_{s=T_n}^{s=t} R_s(W_s^-) \dd s \right) \\
    \mathbb{P}(W \in \dd \omega) &= p_0(W_0) \exp \left( - \int_{s=0}^{s=t} R_s(W_s^-) \dd s \right) \prod_{s: W_s \neq W_s^-} R_s(W_s^-, W_s)
\end{align}
where $p_0$ is the initial state distribution.

We will also need to understand Girsanov's transformation for CTMCs. Girsanov's transformation can be thought of as `importance sampling' for path space measures. Specifically, if we take an expectation with respect to path measure $\mathbb{P}$, $\E_{\mathbb{P}} \left[f (W) \right]$, then this is equal to $\E_{\mathbb{Q}} \left[f(W) \frac{\dd \mathbb{P}}{\dd \mathbb{Q}}(W) \right]$ where $\mathbb{Q}$ is a different path measure and $\frac{\dd \mathbb{P}}{ \dd \mathbb{Q}}$ is known as the Radon-Nikodym derivative. The path measure $\mathbb{Q}$ will result from considering a CTMC with a different rate matrix to our original measure $\mathbb{P}$. Girsanov's transformation allows us to calculate the expectation which should have been taken with respect to the CTMC with $\mathbb{P}$ rate matrix instead with a CTMC with rate matrix corresponding to $\mathbb{Q}$.

The Radon-Nikodym derivative in our case has a form that is simply the ratio of $\mathbb{P}(W \in \dd \omega)$ and $\mathbb{Q}(W \in \dd \omega)$. Let $R_t$, $p_0$ be the rate matrix and initial distribution defining $\mathbb{P}$ and let $R_t'$, $p'_0$ be the rate matrix and initial distribution defining $\mathbb{Q}$.
\begin{equation}
    \frac{\dd \mathbb{P}}{\dd \mathbb{Q}} (W) = \frac{  p_0(W_0) \exp \left( - \int_{s=0}^{s=t} R_s(W_s^-) \dd s \right) \prod_{s: W_s \neq W_s^-} R_s(W_s^-, W_s) } {  p'_0(W_0) \exp \left( - \int_{s=0}^{s=t} R'_s(W_s^-) \dd s \right) \prod_{s: W_s \neq W_s^-} R'_s(W_s^-, W_s)   }
\end{equation}

\subsubsection{Derivation of $\mathcal{L}_{\text{ELBO}}$   }
\label{sec:apdx_derivation_Lelbo}

In this section we will derive the standard evidence lower bound for the model log-likelihood assigned to the data, $\E_{\pdata(\x_1)} \left[ \log p_\theta(\x_1) \right]$ when using our learned generative process to generate data. The entire structure of this section can be understood intuitively by making analogy to the derivation of the evidence lower bound for VAEs, \cite{kingma2013auto, rezende2014stochastic, huang2021variational}. In a VAE, we have a latent variable model $p_\theta(z, x)$ for observed data $x$. To derive the ELBO, we introduce a second distribution over the latent variables $q(z | x)$ with which we will use to take the expectation. The ELBO derivation proceeds as
\begin{align}
    \log p_\theta(x) &= \log \sum_z p_\theta(z, x) \\
   \log p_\theta(x) &= \log \sum_z q(z | x) \frac{p_\theta(z, x)}{q(z | x)} \quad \text{Girsanov's transformation / Importance sampling}\\
    \log p_\theta(x) &\geq \sum_z q(z | x) \log \left( \frac{p_\theta(z, x)}{q(z | x)} \right) \quad \text{Jensen's inequality} \\
    \E_{\pdata(x)} \left[ p_\theta(x) \right] &\geq \E_{\pdata(x) q(z | x)} \left[ \log p_\theta(z, x) \right] + C
\end{align}
In our case, $x$ corresponds to the final state of the generative process at time $t=1$, $\x_1$. The latent variable $z$ corresponds to all other states of the CTMC $W_t$, $t \in [0, 1)$. Our model $p_\theta(z, x)$ corresponds to our generative CTMC with rate matrix $R_t^\theta(\x_t, j) = \E_{p_\theta(\x_1 | \x_t)} \left[R_t(\x_t, j | \x_1) \right]$ and initial distribution $p_0(\x_0)$. Our latent variable distribution $q(z | x)$ corresponds to the $\x_1$ conditioned CTMC that begins at distribution $p_{0|1}(\x_0 | \x_1)$ and simulates with $\x_1$ conditioned rate matrix $R_t(\x_t, j | \x_1)$. We note here that $R_t(\x_t, j | \x_1)$ can be any rate matrix that generates the desired $\x_1$ conditional flow, $\noisemarg(\x_t | \x_1)$ as we described in the main text.

We now derive $\mathcal{L}_{\text{ELBO}}$ using our path space measures for CTMCs. We will use $\mathbb{P}^\theta$ to denote the path measure corresponding to the CTMC simulating from $p_0(\x_0)$ using the generative rate matrix $R_t^\theta(\x_t, j) = \E_{p_\theta(\x_1 | \x_t)} \left[R_t(\x_t, j | \x_1) \right]$. We will use $\mathbb{Q}^{| \x_1}$ to denote the path measure corresponding to the CTMC simulating from $p_{0|1}(\x_0 | \x_1)$ using the $\x_1$ conditioned rate matrix $R_t(\x_t, j | \x_1)$.

We begin by marginalizing out the latent variables, $W_t$, $t \in [0, 1)$ for our generative CTMC
\begin{equation}
    \log p_\theta(\x_1) = \log \int_{W_1 = \x_1} \mathbb{P}^\theta(\dd \omega)
\end{equation}
We now apply Girsnov's transformation using our $\x_1$ conditioned CTMC
\begin{equation}
   \log p_\theta(\x_1) = \log \int_{W_1 = \x_1} \mathbb{Q}^{| \x_1} (\dd \omega) \frac{ \dd \mathbb{P}^\theta }{\dd \mathbb{Q}^{|\x_1}} (\omega)
\end{equation}
where
\begin{equation}
    \frac{\dd \mathbb{P}^\theta }{\dd \mathbb{Q}^{|\x_1}} (\omega) = \frac{ p_0(W_0) \exp \left( - \int_{t=0}^{t=1} R_t^\theta(W_t^-) \dd t \right) \prod_{t: W_t \neq W_t^-} R_t^\theta(W_t^-, W_t)}{p_{0|1}(W_0 | \x_1) \exp \left( - \int_{t=0}^{t=1} R_t(W_t^- | \x_1) \dd t \right) \prod_{t: W_t \neq W_t^-} R_t(W_t^-, W_t | \x_1) }
\end{equation}
we note at this point that $p_{0|1}(W_0 | \x_1) = p_0(W_0)$ and the two intial distribution terms cancel out. Now, apply Jensen's inequality
\begin{equation}
    \log p_\theta(\x_1) \geq \int_{W_1 = \x_1} \mathbb{Q}^{| \x_1} (\dd \omega ) \log \frac{\dd \mathbb{P}^\theta}{\dd \mathbb{Q}^{|\x_1}}(\omega)
\end{equation}
and take the expectation with respect to the data distribution
\begin{equation}
    \E_{\pdata(\x_1)} \left[ \log p_\theta(\x_1) \right] \geq \int \pdata(\dd \x_1) \mathbb{Q}^{| \x_1} (\dd \omega) \log \frac{\dd \mathbb{P}^\theta}{\dd \mathbb{Q}^{|\x_1}}(\omega)
\end{equation}
Finally, substitute in the form for $\frac{\dd \mathbb{P}^\theta}{\mathbb{Q}^{| \x_1}}$ and take terms that don't depend on $\theta$ out into a constant
\begin{align}
    \E_{\pdata(\x_1)}\left[ \log p_\theta(\x_1) \right] &\geq \int \pdata(\dd \x_1) \mathbb{Q}^{|\x_1}(\dd \omega) \left \{ - \int_{t=0}^{t=1} R_t^\theta(W_t^-) \dd t + \sum_{t : W_t \neq W_t^-} \log R_t^\theta(W_t^-, W_t) \right\} + C \\
    &= \int \pdata(\dd \x_1) \mathbb{Q}^{|\x_1}(\dd \omega) \left \{ - \int_{t=0}^{t=1} R_t^\theta(W_t^-) \dd t + \sum_{t : W_t \neq W_t^-} \log \E_{p_\theta(\tx_1 | W_t^-)} \left[ R_t(W_t^-, W_t | \tx_1) \right] \right\} + C \\
    &= \mathcal{L}_{\text{ELBO}} + C
\end{align}
where
\begin{equation}
    \label{eq:intermediate_form_elbo}
    \mathcal{L}_\text{ELBO} = \int \pdata(\dd \x_1) \mathbb{Q}^{|\x_1}(\dd \omega) \left \{ - \int_{t=0}^{t=1} R_t^\theta(W_t^-) \dd t + \sum_{t : W_t \neq W_t^-} \log \E_{p_\theta(\tx_1 | W_t^-)} \left[ R_t(W_t^-, W_t | \tx_1) \right] \right\}
\end{equation}

\subsection{Decomposition of $\mathcal{L}_{\text{ELBO}}$  }
\label{sec:apdx_decomposition_of_Lelbo}

Consider the term $\log \left( \E_{p_\theta(\tx_1 | W_t^-)} \left[ R_t(W_t^-, W_t | \tx_1) \right] \right)$,
\begin{align}
    \log \left( \E_{p_\theta(\tx_1 | W_t^-)} \left[ R_t(W_t^-, W_t | \tx_1) \right] \right) &= \log \left( \E_{p(\tx_1 | W_t^-)} \left[ \frac{p_\theta(\tx_1 | W_t^-)}{p(\tx_1 | W_t^-)} R_t(W_t^-, W_t | \tx_1) \right] \right) \\
    &= \log \left( \E_{p(\tx_1 | W_t^-)} \left[ \frac{p_\theta(\tx_1 | W_t^-)}{p(\tx_1 | W_t^-)} R_t(W_t^-, W_t | \tx_1) \right] \right) \\
    & \quad + \E_{p(\tx_1 | W_t^-)} \left[ \log \left( \frac{p_\theta(\tx_1 | W_t^-)}{p(\tx_1 | W_t^-) } R_t(W_t^-, W_t | \tx_1) \right) \right] \\
    & \quad - \E_{p(\tx_1 | W_t^-)} \left[ \log \left( \frac{p_\theta(\tx_1 | W_t^-)}{p(\tx_1 | W_t^-) } R_t(W_t^-, W_t | \tx_1) \right) \right] \\
    &= \E_{p(\tx_1 | W_t^-)} \left[ \log p_\theta (\tx_1 | W_t^-) \right] + C \\
    & \quad + \log \left( \E_{p(\tx | W_t^-)} \left[ \frac{p_\theta(\tx_1 | W_t^-)}{p(\tx_1 | W_t^-)} R_t(W_t^-, W_t | \tx_1) \right] \right) \\
    & \quad - \E_{p(\tx_1 | W_t^-)} \left[ \log \left( \frac{p_\theta(\tx_1 | W_t^-)}{p(\tx_1 | W_t^-) } R_t(W_t^-, W_t | \tx_1) \right) \right] \\
    &= \E_{p(\tx_1 | W_t^-)} \left[ \log p_\theta (\tx_1 | W_t^-) \right] + C \\
    & \quad + \log \left( \E_{p(\tx | W_t^-)} \left[ \frac{p_\theta(\tx_1 | W_t^-)}{p(\tx_1 | W_t^-)} \mathbb{P}(W_t | W_t^-, \tx_1) R_t(W_t^- | \tx_1) \right] \right) \\
    & \quad - \E_{p(\tx_1 | W_t^-)} \left[ \log \left( \frac{p_\theta(\tx_1 | W_t^-)}{p(\tx_1 | W_t^-) } \mathbb{P}(W_t | W_t^-, \tx_1) R_t(W_t^- | \tx_1) \right) \right] \\
\end{align}
where we have used our definition of the jump distribution of
\begin{equation}
    \mathbb{P}(W_t | W_t^-, \tx_1) = \frac{R_t(W_t^-, W_t | \tx_1)}{R_t(W_t^- | \tx_1) }
\end{equation}
Now we define two new distributions,
\begin{equation}
    p_\theta(\tx_1 | W_t^-) \mathbb{P}(W_t | W_t^-, \tx_1) = p_\theta(W_t | W_t^-) p_\theta(\tx_1 | W_t^-, W_t)
\end{equation}
where
\begin{equation}
    p_\theta(W_t | W_t^-) \vcentcolon =  \sum_{\tx_1} p_\theta(\tx_1 | W_t^-) \mathbb{P}(W_t | W_t^-, \tx_1)
\end{equation}
and
\begin{equation}
    \label{eq:apdx_model_posterior_dual_condition}
    p_\theta(\tx_1 | W_t^-, W_t) \vcentcolon = \frac{ p_\theta(\tx_1 | W_t^-) \mathbb{P}(W_t | W_t^-, \tx_1) }{ \sum_{\x'_1} p_\theta(\x'_1 | W_t^-) \mathbb{P}(W_t | W_t^-, \x'_1) }
\end{equation}
Substitute in these newly defined distributions into our equation for $ \log \left( \E_{p_\theta(\tx | W_t^-)} \left[ R_t(W_t^-, W_t | \tx_1) \right] \right) $ to get
\begin{align}
    \log \left( \E_{p_\theta(\tx | W_t^-)} \left[ R_t(W_t^-, W_t | \tx_1) \right] \right) &= \E_{p(\tx_1 | W_t^-)} \left[ \log p_\theta (\tx_1 | W_t^-) \right] + C \\
    & \quad + \log \left( \E_{p(\tx | W_t^-)} \left[ \frac{ p_\theta(W_t | W_t^-) p_\theta(\tx_1 | W_t^-, W_t) }{p(\tx_1 | W_t^-)} R_t(W_t^- | \tx_1) \right] \right) \\
    & \quad - \E_{p(\tx_1 | W_t^-)} \left[ \log \left( \frac{ p_\theta(W_t | W_t^-) p_\theta(\tx_1 | W_t^-, W_t) }{p(\tx_1 | W_t^-) } R_t(W_t^- | \tx_1) \right) \right] \\
    &= \E_{p(\tx_1 | W_t^-)} \left[ \log p_\theta (\tx_1 | W_t^-) \right] + C \\
    & \quad + \cancel{\log p_\theta(W_t | W_t^-)} + \log \left( \E_{p(\tx | W_t^-)} \left[ \frac{ p_\theta(\tx_1 | W_t^-, W_t) }{p(\tx_1 | W_t^-)} R_t(W_t^- | \tx_1) \right] \right) \\
    & \quad - \cancel{\log p_\theta(W_t | W_t^-)} - \E_{p(\tx_1 | W_t^-)} \left[ \log \left( \frac{  p_\theta(\tx_1 | W_t^-, W_t) }{p(\tx_1 | W_t^-) } R_t(W_t^- | \tx_1) \right) \right] \\
    &= \E_{p(\tx_1 | W_t^-)} \left[ \log p_\theta (\tx_1 | W_t^-) \right] \\
    & \quad + \log \left( \E_{p_\theta(\tx_1 | W_t^-, W_t) } \left[ R_t(W_t^- | \tx_1) \right] \right) \\
    & \quad + \text{KL} \left( p(\tx_1 | W_t^-) \, || \, p_\theta(\tx_1 | W_t^-, W_t) \right) + C
\end{align}
Substituting this into our form for $\mathcal{L}_{\text{ELBO}}$ given in equation \eqref{eq:intermediate_form_elbo} gives
\begin{align}
    \mathcal{L}_{\text{ELBO}} = \int \pdata(\dd \x_1) \mathbb{Q}^{|\x_1}(\dd \omega) \Bigg \{ - \int_{t=0}^{t=1} R_t^\theta(W_t^-) \dd t + \sum_{t: W_t \neq W_t^-} \Bigg(& \E_{p(\tx_1 | W_t^-)} \left[ \log p_\theta (\tx_1 | W_t^-) \right] + \\
    & \log \left( \E_{p_\theta(\tx_1 | W_t^-, W_t) } \left[ R_t(W_t^- | \tx_1) \right] \right) + \\
    &\text{KL} \left( p(\tx_1 | W_t^-) \, || \, p_\theta(\tx_1 | W_t^-, W_t) \right) \Bigg) \Bigg\}
\end{align}

Substituting this into our original bound on the model log-likelihood gives
\begin{equation}
    \E_{\pdata(\x_1)} \left[ \log p_\theta( \x_1) \right] \geq \mathcal{L}_{\text{ELBO}} + C =  \mathcal{L}_{\text{ce}} + \mathcal{L}_{R} + \mathcal{L}_{\text{KL}} + C
\end{equation}
where
\begin{align}
    \mathcal{L}_{\text{ce}} &= \int \pdata(\dd \x_1) \mathbb{Q}^{|\x_1}(\dd \omega) \sum_{t: W_t^- \neq W_t} \E_{p(\tx_1 | W_t^-)} \left[ \log p_\theta(\tx_1 | W_t^-) \right] \\
    \mathcal{L}_{R} &= \int \pdata(\dd \x_1) \mathbb{Q}^{|\x_1}(\dd \omega) \Bigg \{ - \int_{t=0}^{t=1} R_t^\theta(W_t) \dd t + \sum_{t: W_t^- \neq W_t} \log \left( \E_{p_\theta(\tx_1 | W_t^-, W_t) } \left[  R_t(W_t^- | \tx_1) \right] \right) \Bigg \} \\
    \mathcal{L}_{\text{KL}} &= \int \pdata(\dd \x_1) \mathbb{Q}^{|\x_1}(\dd \omega) \sum_{t: W_t^- \neq W_t} \text{KL} \left(  p(\tx_1 | W_t^-) \, || \, p_\theta(\tx_1 | W_t^-, W_t) \right)
\end{align}
and $C$ is a constant term independent of $\theta$.

In the next stages of the proof, we going to show that $\mathcal{L}_{\text{ce}}$ is the weighted cross-entropy, $\mathcal{L}_{R}$ is a regularizer towards the arbitrarily chosen $\x_1$ conditioned rate matrix that we argue we can ignore and $\mathcal{L}_{\text{KL}}$ is a KL term that we will absorb into the bound on the model log-likelihood.\\

In order to proceed, we will need to make use of Dynkin's formula
\begin{equation}
    \int \pdata(\dd \x_1) \mathbb{Q}^{|\x_1} (\dd \omega) \sum_{t: W_t^- \neq W_t} f( W_t^-, W_t) = \int \pdata(\dd \x_1) \mathbb{Q}^{|\x_1} (\dd \omega) \int_{t=0}^{t=1} \sum_{y \neq W_t} R_t(W_t, y | \x_1) f(W_t, y) \dd t
\end{equation}
where $f(\cdot, \cdot)$ is a two-argument function. This formula can be understood intuitively as allowing us to switch from a sum over the jump times to a full integral over the time interval appropriately weighted by the probability that a jump occurs and the destination to which a jump goes to.\\

\paragraph{Weighted Cross Entropy}
We first show that $\mathcal{L}_{\text{ce}}$ is the weighted cross entropy.
\begin{align}
     \mathcal{L}_{\text{ce}} &= \int \pdata(\dd \x_1) \mathbb{Q}^{|\x_1}(\dd \omega) \sum_{t: W_t^- \neq W_t} \E_{p(\tx_1 | W_t^-)} \left[ \log p_\theta(\tx_1 | W_t^-) \right] \\
     &= \int \pdata(\dd \x_1) \mathbb{Q}^{|\x_1}(\dd \omega) \int_{t=0}^{t=1} \sum_{y \neq W_t} R_t(W_t, y | \x_1) \E_{p(\tx_1 | W_t)} \left[ \log p_\theta(\tx_1 | W_t) \right] \dd t \hspace{0.5cm} \text{Dynkin} \\ 
     &= \int \int_{t=0}^{t=1} \pdata(\dd \x_1) \mathbb{Q}^{|\x_1}(\dd \omega) \E_{p(\tx_1 | W_t)} \left[ \log p_\theta(\tx_1 | W_t) \right] R_t(W_t | \x_1) \dd t\\
     &= \mathbb{E}_{\pdata(\x_1) \mathcal{U}(t; 0, 1) p(\x_t | \x_1) } \left[ R_t(\x_t | \x_1) \E_{p(\tx_1 | \x_t) } \left[ \log p_\theta(\tx_1 | \x_t) \right] \right]\\
     &= \mathbb{E}_{\pdata(\x_1) \mathcal{U}(t; 0, 1) p(\x_t | \x_1) p(\tx_1 | \x_t) } \left[ R_t(\x_t | \x_1) \log p_\theta(\tx_1 | \x_t) \right] \\
     &= \mathbb{E}_{ \mathcal{U}(t; 0, 1)  p(\x_1, \x_t)  p(\tx_1 | \x_t) } \left[ R_t(\x_t | \x_1) \log p_\theta(\tx_1 | \x_t) \right] \\
     &= \mathbb{E}_{ \mathcal{U}(t; 0, 1)  p(\x_t) p(\x_1 | \x_t)  p(\tx_1 | \x_t) } \left[ R_t(\x_t | \x_1) \log p_\theta(\tx_1 | \x_t) \right] \\
     &= \mathbb{E}_{ \mathcal{U}(t; 0, 1)  p(\x_t) p(\tx_1 | \x_t)  p(\x_1 | \x_t) } \left[ R_t(\x_t | \tx_1) \log p_\theta(\x_1 | \x_t) \right] \hspace{1cm} \text{Relabel $\x_1 \leftrightarrow \tx_1 $} \\
     &= \mathbb{E}_{ \mathcal{U}(t; 0, 1)  p(\x_t) p(\x_1 | \x_t) } \left[ \E_{p(\tx_1 | \x_t) } \left[ R_t(\x_t | \tx_1) \right] \log p_\theta(\x_1 | \x_t) \right] \\
     &= \mathbb{E}_{ \mathcal{U}(t; 0, 1)  p(\x_t) p(\x_1 | \x_t) } \left[ \omega_t(\x_t) \log p_\theta(\x_1 | \x_t) \right]
\end{align}
where on the second line we apply Dynkin's formula with $f(W_t^-, W_t) = \E_{p(\tx_1 | W_t^-)} \left[ \log p_\theta(\tx_1 | W_t^-) \right]$ which we note is independent of $W_t$. $\omega_t(\x_t)$ is a weighting function. In diffusion model training it is common for the likelihood based objective to be a weighted form of a recognisable loss e.g. the L2 loss for diffusion models. Here we have a `likelihood weighted' cross entropy. We can then make the same approximation as in diffusion models and set $\omega(\x_t) = 1$ to equally weight all loss levels. This also has the benefit of making our loss independent of the arbitrarily chosen rate matrix $R_t$ that could have been any rate that generates the desired conditional flow.

\paragraph{Rate Forcing Term}
We now analyse the term $\mathcal{L}_{R}$. We will show that it is approximately equal to an objective which at its optimum sets the learned generative rate matrix to have the same overall jump probability as the arbitrarily chosen rate matrix that generates our $\noisemarg(\x_t | \x_1)$ conditional flow.

\begin{align}
    \mathcal{L}_{R} &= \int \pdata(\dd \x_1) \mathbb{Q}^{|\x_1}(\dd \omega) \Bigg \{ - \int_{t=0}^{t=1} R_t^\theta(W_t) \dd t + \sum_{t: W_t^- \neq W_t} \log \left( \E_{p_\theta(\tx_1 | W_t^-, W_t) } \left[  R_t(W_t^- | \tx_1) \right] \right) \Bigg \} \\
    &= \int \pdata(\dd \x_1) \mathbb{Q}^{|\x_1}(\dd \omega) \Bigg \{ - \int_{t=0}^{t=1} R_t^\theta(W_t) \dd t + \int_{t=0}^{t=1} \sum_{y \neq W_t} R_t(W_t, y | \x_1)  \log \left( \E_{p_\theta(\tx_1 | W_t, y)} \left[ R_t(W_t | \tx_1) \right] \right) \dd t \Bigg \}
\end{align}
where on the second line we have applied Dynkin's formula with $f(W_t^-, W_t) = \E_{p_\theta(\tx_1 | W_t^-, W_t)} \left[ R_t(W_t^- | \tx_1) \right]$. To further understand this term, we make the following approximation
\begin{equation}
    \E_{p_\theta(\tx_1 | W_t, y)} \left[ R_t(W_t | \tx_1) \right] \approx \E_{p_\theta(\tx_1 | W_t)} \left[ R_t(W_t | \tx_1) \right]
\end{equation}
$p_\theta(\tx_1 | W_t, y)$ is the Bayesian posterior update given by equation \eqref{eq:apdx_model_posterior_dual_condition} starting with prior $p_\theta(\tx_1 | W_t)$ and with likelihood $\mathbb{P}(y | W_t, \tx_1)$. It is therefore the models prediction of $\tx_1$ updated with the information that the process has jumped to new value $y$. When our CTMC is multi-dimensional then a single jump will change only a single dimension, see Appendix \ref{sec:factorization}, and so when we operate in high-dimensional settings, the Bayesian posterior will be close to the prior.

We will denote the approximate form of $\mathcal{L}_R$ as $\hat{\mathcal{L}}_R$.
\begin{align}
    \hat{\mathcal{L}}_R &= \int \pdata(\dd \x_1) \mathbb{Q}^{|\x_1}(\dd \omega) \Bigg \{ - \int_{t=0}^{t=1} R_t^\theta(W_t) \dd t + \int_{t=0}^{t=1} \sum_{y \neq W_t} R_t(W_t, y | \x_1) \log \left( \E_{p_\theta(\tx_1 | W_t)} \left[ R_t(W_t | \tx_1) \right] \right) \dd t \Bigg \} \\
    &= \int \pdata(\dd \x_1) \mathbb{Q}^{|\x_1}(\dd \omega) \Bigg \{ - \int_{t=0}^{t=1} R_t^\theta(W_t) \dd t + \int_{t=0}^{t=1} \log \left( \E_{p_\theta(\tx_1 | W_t)} \left[ R_t(W_t | \tx_1) \right] \right) R_t(W_t | \x_1) \dd t \Bigg \} \\
    &= \int \int_{t=0}^{t=1} \pdata(\dd \x_1) \mathbb{Q}^{|\x_1}(\dd \omega) \Bigg \{ - R_t^\theta(W_t) + R_t(W_t | \x_1) \log R_t^\theta(W_t) \Bigg \} \dd t \\
    &= \E_{\mathcal{U}(t; 0, 1) \pdata(\x_1) p_t(\x_t | \x_1) } \left[ -R_t^\theta(\x_t) + R_t(\x_t | \x_1) \log R_t^\theta(\x_t) \right] \\
    &= \E_{\mathcal{U}(t; 0, 1) p_t(\x_t)} \left[ -R_t^\theta(\x_t) + \E_{p(\x_1 | \x_t) } \left[ R_t(\x_t | \x_1) \right] \log R_t^\theta(\x_t)  \right]
\end{align}
where on the third line we have used the definition of $R_t^\theta(W_t) = \E_{p_\theta(\tx_1 | W_t) } \left[ R_t(W_t | \tx_1) \right]$. Now consider maximizing $\hat{\mathcal{L}}_R$ with respect to the value of $R_\tau^\theta(z)$ at test input $z$ and test time $\tau$. Differentiating $\hat{\mathcal{L}}_R$ with respect to $R_\tau^\theta(z)$ and setting to $0$ gives
\begin{align}
    \frac{\partial \hat{\mathcal{L}}_R}{\partial R_\tau(z)} &= p_\tau(z) \left( -1 + \E_{p(\x_1 | z)}\left[R_\tau(z | \x_1) \right] \frac{1}{R_\tau^\theta(z) } \right) = 0 \\
    & \implies R_\tau^\theta(z) = \E_{p(\x_1 | z)} \left[R_\tau(z | \x_1) \right] \quad \text{ at stationarity}
\end{align}
Therefore, we have found that maximizing $\hat{\mathcal{L}}_R$ encourages $R_t^\theta(\x_t)$ to equal $\E_{p(\x_1 | \x_t)} \left[ R_t(\x_t | \x_1) \right]$. However, $R_t(\x_t | \x_1)$ is the overall rate of jumps for the arbitrarily chosen rate matrix that generates the $\noisemarg(\x_t | \x_1)$ conditional flow. This rate of jumps is completely dependent on the level of stochasticity chosen for $R_t(\x_t | \x_1)$ which does not have any a priori known correct level. Therefore, we do not want to be encouraging our learned generative rate matrix $R_t^\theta$ to be matching this stochasticity level and so the term $\hat{\mathcal{L}}_R$ is undesirable to have in the objective. The true evidence lower bound includes the term $\mathcal{L}_R$ which we expect to have a similar effect as $\hat{\mathcal{L}}_R$ as we argued previously.

\paragraph{KL Term}
When we maximize the $\mathcal{L}_{\text{ELBO}}$ objective, we would try to maximize the $\mathcal{L}_{\text{KL}}$ term i.e. we try and push $p(\tx_1 | W_t^-)$ and $p_\theta(\tx_1 | W_t^-, W_t)$ as far apart as possible. This makes sense to do as we try and push the posterior over $\tx_1$ given the information contained in both the pre-jump state $W_t^-$ and the post jump state $W_t$ away from the distribution over $\tx_1$ given just the information within $W_t^-$. Digging into this term deeper we see that
\begin{align}
    &\text{KL} \left( p(\tx_1 | W_t^-) \, || \, p_\theta(\tx_1 | W_t^-, W_t) \right) \\
    &= \E_{p(\tx_1 | W_t^-)} \left[ \log \frac{p(\tx_1 | W_t^-)}{p_\theta(\tx_1 | W_t^-, W_t)} \right] \\
    &= \E_{p(\tx_1 | W_t^-)} \left[ - \log \left( p_\theta(\tx_1 | W_t^-) \mathbb{P}(W_t | W_t^-, \tx_1) \right) + \log \left( \sum_{\x'_1} p_\theta(\x'_1 | W_t^-) \mathbb{P}(W_t | W_t^-, \x'_1) \right) \right] + C \\
    &= \E_{p(\tx_1 | W_t^-)} \left[ - \log p_\theta(\tx_1 | W_t^-) +  \log \left( \sum_{\x'_1} p_\theta(\x'_1 | W_t^-) \mathbb{P}(W_t | W_t^-, \x'_1) \right)\right] + C
\end{align}
where we have substituted in our definition of $p_\theta(\tx_1 | W_t^-, W_t)$ given by equation \eqref{eq:apdx_model_posterior_dual_condition}. We see that the first term $ - \log p_\theta(\tx_1 | W_t^-)$ cancels with our cross entropy term. This then makes clear how we have arrived at our cross entropy decomposition of $\mathcal{L}_\text{ELBO}$. $\mathcal{L}_\text{ELBO}$ will usually remove the cross entropy training signal and replace it with the term $ \log \left( \sum_{\x'_1} p_\theta(\x'_1 | W_t^-) \mathbb{P}(W_t | W_t^-, \x'_1) \right) $ which will be used as the training signal for the denoising model $p_\theta(\x_1 | W_t^-)$. The denoising model is encouraged to be such that the expected jump probability assigns high likelihood to the jump observed under the $\x_1$ conditioned process $\mathbb{Q}^{|\x_1}$. This is an indirect training signal for $p_\theta(\x_1 | W_t^-)$ and one that relies on the arbitrary specification of our $\mathbb{Q}^{| \x_1}$ process. We instead show how we can replace this $p_\theta(\x_1 | W_t^-)$ training signal with the cross entropy loss and be left with a KL term showing that the cross entropy is a lower bound on $\mathcal{L}_{\text{ELBO}}$ minus the rate regularizing term. We summarize this argument in the next section.

\paragraph{Summary}
To summarize, we have first derived the standard evidence lower bound on the model log-likelihood when using our specific generative rate matrix, $R_t^\theta(\x_t, j) = \E_{p_\theta(\x_1 | \x_t)} \left[ R_t(\x_t, j | \x_1) \right]$ for some arbitrarily chosen $R_t(\x_t, j | \x_1)$ that generates the $\noisemarg(\x_t | \x_1)$ conditional flow.
\begin{equation}
    \E_{\pdata(\x_1)} \left[ \log p_\theta(\x_1) \right] \geq \mathcal{L}_{\text{ELBO}} + C
\end{equation}
We then split $\mathcal{L}_{\text{ELBO}}$ into three terms $\mathcal{L}_{\text{ce}} + \mathcal{L}_R + \mathcal{L}_{\text{KL}}$. We have seen how the term $\mathcal{L}_{\text{KL}}$ allows us to remove the standard $\mathcal{L}_\text{ELBO}$ training signal for the denoising model $p_\theta(\x_1 | \x_t)$ and replace it with the cross entropy, creating the $\mathcal{L}_{\text{ce}}$ term. This creates a looser bound if we are to train without the $\mathcal{L}_{\text{KL}}$ term,
\begin{equation}
    \E_{\pdata(\x_1)} \left[ \log p_\theta(\x_1) \right] \geq \mathcal{L}_\text{ce} + \mathcal{L}_R + C
\end{equation}
We then argue that $\mathcal{L}_R$ is close to $\hat{\mathcal{L}}_R$ which is an unnecessary forcing term encouraging our generative rate to achieve a similar jump rate to our chosen $R_t(\x_t, j | \x_1)$ even though this $R_t$ matrix is an arbitrary decision and will have a different jump rate depending on which $R_t$ is chosen. We are then left with the standard cross entropy term as our final objective for $p_\theta(\x_1 | \x_t)$ with a final modification to its unweighted form for implementation ease.

\subsubsection{Objective for the Masking Interpolant}
\label{sec:apdx_Lelbo_for_masking}
In this section we will show that $\mathcal{L}_\text{ELBO}$ is exactly the weighted cross entropy for the case when we use the masking form for $\noisemarg(\x_t | \x_1)$. We note that a similar result has been proven by \citet{austin2021structured} for the discrete time diffusion model, and here we verify that this result also holds for our \method model. We will assume multi-dimensional data, $\x_1 \in \{ 1, \dots, \statespace \}^D$. We refer to Appendix \ref{sec:factorization} for the details of the multi-dimensional setting. We will also assume that we use $\relurate_t$ as our rate matrix that generates the $\noisemarg(\x_t | \x_1)$ conditional flow.

Before we manipulate $\mathcal{L}_\text{ELBO}$, we will first find the forms of $\relurate_t(\x_t^{1:D}, j^{1:D} | \x_1^{1:D})$, $R_t^\theta(\x_t^{1:D}, j^{1:D})$ and $R_t^\theta(\x_t^{1:D})$ for the masking case. From Appendix \ref{sec:apdx_masking_example}, equation \eqref{eq:relu_rate_multi_dim_masking} we have,
\begin{equation}
    {\relurate_t}^d(\x_t^d, j^d | \x_1^d) = \frac{1}{1-t} \kdelta{ j^d }{\x_1^d } \kdelta { \x_t^d}{ M }
\end{equation}
and so
\begin{align}
    {\relurate_t}(\x_t^{1:D}, j^{1:D} | \x_1^{1:D}) &= \sum_{d=1}^D \kdelta { \x_t^{\oneDd}} {j^{\oneDd} } {\relurate_t}(\x_t^d, j^d | \x_1^{d}) \\
    &= \sum_{d=1}^D \kdelta { \x_t^{\oneDd}}{ j^{\oneDd} } \kdelta { j^d}{ \x_1^d } \kdelta {\x_t^d}{ M } \frac{1}{1-t}
\end{align}

From Appendix \ref{sec:apdx_masking_example}, equation \eqref{eq:masking_uncond_rate_d} we have that,
\begin{equation}
    R_t^{\theta d} (\x_t^{1:D}, j^d) = \frac{p_\theta(\x_1^d = j^d | \x_t^{1:D})}{1-t} \kdelta { \x_t^d}{ M }
\end{equation}
and therefore, 
\begin{align}
    R_t^\theta(\x_t^{1:D}, j^{1:D}) &= \sum_{d=1}^D \kdelta { \x_t^{\oneDd}}{ j^{\oneDd} } R_t^{\theta d} (\x_t^{1:D}, j^d) \\
    &= \sum_{d=1}^D \kdelta { \x_t^{\oneDd}}{ j^{\oneDd} } \frac{p_\theta(\x_1^d = j^d | \x_t^{1:D})}{1-t} \kdelta { \x_t^d}{ M }
\end{align}
We now find $R_t^\theta(\x_t^{1:D})$
\begin{align}
    R_t^\theta(\x_t^{1:D}) &= \sum_{j^{1:D} \neq \x_t^{1:D}} R_t^\theta(\x_t^{1:D}, j^{1:D})\\
    &=\sum_{j^{1:D}} \left( 1 - \kdelta { j^{1:D}}{ \x_t^{1:D} } \right) \sum_{d=1}^D \kdelta { j^{\oneDd}}{ \x_t^{\oneDd} } \frac{p_\theta(\x_1^d = j^d | \x_t^{1:D})}{1-t} \kdelta { \x_t^d}{ M } \\
    &= \sum_{d=1}^D \sum_{j^{\oneDd}} \kdelta { j^{\oneDd}}{ \x_t^{\oneDd} } \kdelta { \x_t^d}{ M } \frac{1}{1-t} \sum_{j^d} \left( 1 - \kdelta { j^{1:D}}{ \x_t^{1:D} } \right) p_\theta(\x_1^d = j^d | \x_t^{1:D}) \\
    &= \sum_{d=1}^D \kdelta { \x_t^d}{ M } \frac{1}{1-t} \sum_{j^d} \left(1 - \kdelta { j^d }{\x_t^d } \right) p_\theta(\x_1^d = j^d | \x_t^{1:D})\\
    &= \sum_{d=1}^D \kdelta { \x_t^d}{ M } \frac{1}{1-t} \sum_{j^d \neq \x_t^d} p_\theta(\x_1^d = j^d | \x_t^{1:D})\\
    &= \sum_{d=1}^D \kdelta {\x_t^d}{ M } \frac{1}{1-t}
\end{align}
where on the final line we have used the fact that $p_\theta(\x_1^d = M | \x_t^{1:D}) = 0$.

We are now ready to manipulate the form of $\mathcal{L}_\text{ELBO}$. We start with 
\begin{equation}
    \mathcal{L}_\text{ELBO} = \int \pdata(\dd \x_1) \mathbb{Q}^{|\x_1}(\dd \omega) \left( - \int_{t=0}^{t=1} R_t^\theta(W_t) \dd t + \sum_{t: W_t^- \neq W_t} \log \left( R_t^\theta(W_t^-, W_t) \right) \right) + C
\end{equation}
We then apply Dynkin's formula
\begin{equation}
    \mathcal{L}_\text{ELBO} = \int \pdata(\dd \x_1) \mathbb{Q}^{|\x_1}(\dd \omega) \left( \int_{t=0}^{t=1} -R_t^\theta(W_t)  + \sum_{y \neq W_t} \relurate_t(W_t, y | \x_1) \log \left( R_t^\theta(W_t, y) \right) \dd t \right) + C
\end{equation}
We now substitute in the masking forms for $R_t^\theta(W_t)$, $\relurate_t(W_t, y | \x_1)$ and $R_t^\theta(W_t, y)$
\begin{align}
    \mathcal{L}_\text{ELBO} = \int \pdata(\dd \x_1) \mathbb{Q}^{|\x_1}(\dd \omega) \Bigg( \int_{t=0}^{t=1} &\left( - \sum_{d=1}^D \kdelta { W_t^d}{ M } \frac{1}{1-t} \right) +\\
    & \sum_{y^{1:D} \neq W_t^{1:D}} \Bigg\{ \left( \sum_{d=1}^D \kdelta { W_t^{\oneDd}}{ y^{\oneDd} } \kdelta { y^d}{ \x_1^d } \kdelta { W_t^d}{ M } \frac{1}{1-t} \right) \times \\
    & \qquad \log \left( \sum_{d=1}^D \kdelta { W_t^{\oneDd}}{ y^{\oneDd} } \kdelta { W_t^d}{ M } p_\theta(y^d | W_t^{1:D}) \frac{1}{1-t} \right) \Bigg\} \dd t \Bigg) + C
\end{align}
\begin{align}
    \mathcal{L}_\text{ELBO} = \int \pdata(\dd \x_1) \mathbb{Q}^{|\x_1}(\dd \omega) \Bigg( \int_{t=0}^{t=1} \sum_{d=1}^D \sum_{y^d \neq W_t^d} \kdelta { W_t^d }{M } \kdelta { y^d}{ \x_1^d } \frac{1}{1-t} \log \left( p_\theta(y^d | W_t^{1:D}) \right) \dd t \Bigg) + C
\end{align}
where we have moved terms that don't depend on $\theta$ into the constant.
\begin{align}
    \mathcal{L}_\text{ELBO} &= \int \pdata(\dd \x_1) \mathbb{Q}^{|\x_1}(\dd \omega) \Bigg( \int_{t=0}^{t=1} \sum_{d=1}^D \kdelta { W_t^d}{ M } \frac{1}{1-t} \log \left( p_\theta(\x_1^d | W_t^{1:D}) \right) \dd t \Bigg) + C\\
    &= \E_{\mathcal{U}(t; 0, 1) \pdata(\x_1) p_t(\x_t | \x_1)} \left[ \sum_{d=1}^D \kdelta {\x_t^d}{ M } \frac{1}{1-t} \log p_\theta(\x_1^d | \x_t^{1:D}) \right] + C \label{eq:masking_elbo}
\end{align}
where we have arrived at the weighted cross entropy, weighted by $\frac{1}{1-t}$ and only calculated for dimensions that are masked in our corrupted sample $\x_t$.

\section{Discussion of Related Work}
\label{sec:apdx_related_work}
Flow based methods for generative modelling were introduced by \cite{liu2022flow, albergo2022building, lipman2022flow}.
These methods simplify the generative modelling framework over diffusion models \cite{sohl2015deep, ho2020denoising, song2020score} by considering noise-data interpolants rather than considering forward/backward diffusions.
This work brings these benefits to discrete data denoising models which previously have used the diffusion methodology \cite{sohl2015deep, hoogeboom2021argmax, austin2021structured} relying on forward/backward processes defined by Markov transition kernels.
Specifically, prior discrete diffusion works first define a forward noising process with a rate matrix $\tilde{R}_t$. This defines infinitesimal noise additions. To train the model, we need access to the equivalent of $\noisemarg$, i.e. the total amount of noise added simulating from $1$ to $t$. To find this value, the matrix exponential needs to be applied to the forward rate matrix, $\noisemarg = \exp \left( \int_t^1 \tilde{R}_s \dd s \right)$. This means that discrete diffusion models are limited in the choice of forward noising process. The choice of $\tilde{R}_t$ must be such that the matrix exponential is tractable. For \method, we simply write down $\noisemarg$ rather than implicitly defining it through the matrix exponential and then can find a rate matrix to simulate with by differentiating $\noisemarg$ and using $\relurate_t$. Furthermore, the standard ELBO objective used to train discrete diffusion models depends on the initial choice of $\tilde{R}_t$. At sample time, it is then standard to simulate with the time reversal of $\tilde{R}_t$. This needlessly limits the choice of simulation process as we have shown in this work that there are infinitely many valid choices of rate matrices that could be used for sampling.

There have been post-hoc changes to the sampling process made in prior work e.g. corrector steps used by \citet{campbell2022continuous}, however due to the ELBO maximizing the model log-likelihood under the assumption of sampling using the time-reversal, the diffusion framework still revolves around one `canonical' sample time process (the time-reversal) whereas \method makes it clear this choice is arbitrary and the sample process can be chosen at inference time for best performance.

Previous discrete diffusion works have also suggested alternatives to the ELBO.
\citet{sun2022score} introduce a categorical score matching loss that resembles the cross entropy, however, the denoising network is required to make a prediction $\x_0^d$ based only on the other $D-1$ dimensions of the input noisy state, $\x_t^{\oneDd}$. This requires specialized architectures and methods to remain computationally efficient.
\citet{vignac2022digress} propose to learn a diffusion based model solely using the cross-entropy but do not analyse the link between the cross-entropy and the log-likelihood of the model as we do in \cref{sec:apdx_analysis_of_training_objective}.
\citet{meng2022concrete} propose to learn a discrete score model based on data ratios using an L2 based loss which has some undesirable properties such as not penalizing mode dropping as described by \citet{lou2023discrete}.
\citet{lou2023discrete} refine this approach and propose to learn data ratios using the score entropy loss which, like the standard cross entropy, does not depend on the choice of forward rate matrix. However, in order for the score entropy to be a true ELBO, the forward rate matrix needs to be used as a weighting factor.

Multimodal diffusion models have been applied to tabular data \cite{kotelnikov2023tabddpm} where continuous diffusion is used for continuous features and a uniform style of corruption under a discrete diffusion framework is applied to discrete features. This idea was then expanded to molecule generation where the task is to generate a molecules atom types, their positions and their connectivity. \citet{peng2023moldiff} use a masking process for the discrete atom types and bond types with a continuous space process for the atom positions. \citet{vignac2023midi} use a discrete process converging towards the independent marginal distribution in each dimension \citep{vignac2022digress} for atom types, bond types and formal charges of the molecules along with a continuous process for atom positions. \citet{hua2023mudiff} use a uniform discrete process for bond types with a continuous space process applied to atom positions as well as atom features embedded in continuous space.
These works also investigate the importance of the multimodal noise schedule. \citet{peng2023moldiff} find that corrupting the bonds first and then the atom positions improves performance by avoiding unphysical bonds appearing in the corruption process.
\citet{vignac2023midi} have a similar finding that during corruption, the atom types should be corrupted first, then the bond types and finally the atom positions.
We generalize these ideas by using the approach of \citet{albergo2023multimarginal} and learning our model over all relative levels of noise between our modalities. This allows picking the desired path through the multimodal noise landscape at inference time either performing co-generation, inverse folding or forward folding.

Other approaches for discrete data modelling opt to embed the data into a continuous space in order to still use the continuous diffusion framework \cite{li2022diffusion, chen2022analog,richemond2022categorical,gong2022diffuseq, dieleman2022continuous, han2022ssd,      strudel2022self, gulrajani2023likelihood, floto2023diffusion}, however, this loses the discrete structure of the data during generation.
This can be important when the quantity that is represented by the discrete variable as algorithmic importance. For example, \citet{qin2023sparse} perform sparse graph generation where the discrete token represents the existence of an edge. It is then important for the edge to be known to physically exist or not so that sparse graph networks can be applied to the problem. 

General Fokker-Planck equations on discrete state spaces \cite{chow2012fokker} have been used to construct sampling methods for energy functions \cite{sun2023discrete}. Further, in a generative modelling context, the Kolmogorov equation has been used to construct equivalent diffusion processes with fewer transitions \cite{zhang2023formulating} making links to optimal transport. We take this idea further to build a generative modelling paradigm around the flexibility of the Kolmogorov equation.

The construction of discrete diffusion model from a marginal distribution perspective as opposed to a forward corruption process has also been used by \citet{chen2023fast}. Their method defines the marginal distribution at time $t$ as a combination of the data and a noise sample and then finds a process that generates those marginals, for the masking and uniform case. They use this to create a faster sampling algorithm by exploiting the fact that if you have a low stochasticity process, you know there should only be $D$ transitions in the masking case (although this is not the case in the unconditional uniform case). Therefore, when conditioning on these times, only $D$ function evaluations are needed. This approach could also be used with a DFM when $\eta = 0$, however, our general framework also demonstrates the benefits of allowing $\eta > 0$.

The consideration of flows on discrete state spaces has also been used to construct GFlowNet algorithms \cite{bengio2023gflownet} which aim to sample from a given energy function. Here we instead focus on the the generative modeling context where we aim to sample novel datapoints when only given access to some dataset of training examples. GFlowNets also can use the detailed balance equation \cref{eq:detailed_balance} as a training training objective. Detailed balance is also used in Markov Chain Monte Carlo methods \citep{metropolis1953equation, hastings1970monte} to construct a transition probability with the desired energy function that we wish to sample from as its stationary distribution.  In our work, we use the detailed balance condition as a way to increase the inference time flexibility in our framework

\section{Multidimensional Data}
\label{sec:factorization}
In this section we derive how we can efficiently model $D$ dimensional data, $\x_1 \in \{ 1, \dots, \statespace \}^D$ by using factorization assumptions. When we wish to emphasize the multidimensional aspect we can write $\x_1^{1:D}$ and use $\x_1^d \in \{1, \dots, \statespace \}$ to refer to the value in dimension $d$. We use $\oneDd$ to denote all dimensions except $d$. To operate in multidimensional spaces, we will make the following assumptions
\begin{itemize}
    \item \textbf{Assumption 1} \quad $\noisemarg(\x_t^{1:D} | \x_1^{1:D}) = \prod_{d=1}^D \noisemarg(\x_t^d | \x_1^d)$
    \item \textbf{Assumption 2} \quad $\noisemarg(\x_t^d | \x_1^d) = 0 \implies \partial_t \noisemarg(\x_t^d | \x_1^d) = 0, \forall d$
    \item \textbf{Assumption 3} \quad $R_t(\x_t^{1:D}, j^{1:D} | \x_1^{1:D}) = \sum_{d=1}^D \delta \{ \x_t^{\oneDd}, j^{\oneDd} \} R_t^d(\x_t^d, j^d | \x_1^d) $
\end{itemize}
The first assumption creates independent corruption processes in each dimension, similar to the factorization assumptions made in diffusion models where the forward noising processes proceed independently in each dimension. Assumption 2 is the same assumption we made in order to derive $\relurate_t$ in $1$-dimension but now we assume it individually for every dimension. Finally, assumption 3 states that for our data conditional rate matrix, it decomposes into a sum of rate matrices for each dimension and so the rate for transitions that change more than $1$ dimension at a time are 0. This is the same assumption made by \citet{campbell2022continuous} in order to make calculations tractable. We will enable our process to make multiple dimensional changes simultaneously later when we come to derive our sampling algorithm.

Under these assumptions, we will now derive \method for the multidimensional case. We start with the data conditional Kolmogorov equation
\begin{equation}
    \partial_t \noisemarg(\x_t^{1:D} | \x_1^{1:D}) = \sum_{j^{1:D}} R_t(j^{1:D}, \x_t^{1:D} | \x_1^{1:D}) \noisemarg(j^{1:D} | \x_t^{1:D})
    \label{eq:apx_multidim_conditional_continuity}
\end{equation}
We now substitute the form for the rate matrix under Assumption 3 into the RHS of \eqref{eq:apx_multidim_conditional_continuity} to get
\begin{align}
    \text{RHS} &= \sum_{j^{1:D}} \sum_{d=1}^D \delta \{ \x_t^{1:D \backslash d}, j^{1:D \backslash d} \} R_t^d(j^d, \x_t^d | \x_1^d) \noisemarg(j^{1:D} | \x_1^{1:D})\\
    \label{eq:R_factorize_intermediate_continuity}
    &= \sum_{d=1}^D \sum_{j^d} R_t^d(j^d, \x_t^d | \x_1^d) \noisemarg(\x_t^{1:D \backslash d} \odot j^d | \x_1^{1:D})
\end{align}
where we use $\x_t^{1:D \backslash d} \odot j^d$ to denote a vector of dimension $D$ where in the $d$-th dimension it has the value of $j^d$ and in the other dimensions it has values $\x_t^{1:D \backslash d}$. We now verify that the following form for $R_t^d$ satisfies the Kolmogorov equation,
\begin{equation}
    \label{eq:apdx_multidim_relu_rate}
    {\relurate_t}^d(\x_t^d, j^d | \x_1^d) = \begin{cases} \frac{\relu \left( \partial_t \noisemarg(j^d | \x_1^d) - \partial_t \noisemarg(\x_t^d | \x_1^d) \right)}{\mathcal{Z}_t^d \noisemarg(\x_t^d | \x_1^d) } &\text{ for } \noisemarg(\x_t^d | \x_1^d) > 0, \noisemarg(j^d | \x_1^d) > 0\\
    = 0 &\text{ otherwise}
    \end{cases}
\end{equation}
where $\mathcal{Z}_t^d = | \{ j^d : \noisemarg(j^d | \x_1^d) > 0 \} |$ and we only define ${\relurate_t}^d$ for off-diagonal entries, $\x_t^d \neq j^d$ remembering that ${\relurate_t}^d(\x_t^d, \x_t^d | \x_1^d) = -\sum_{j^d \neq \x_t^d} {\relurate_t}^d(\x_t^d, j^d | \x_1^d)$.

We first assume $\noisemarg(\x_t^d | \x_1^d) > 0 \, \forall d$ and substitute in ${\relurate_t}^d$ into equation \eqref{eq:R_factorize_intermediate_continuity}.
\begin{align}
    \text{RHS} = \sum_{d=1}^D \sum_{j^d \neq \x_t^d, \noisemarg(j^d | \x_1^d) > 0} \Bigg(& \frac{\relu \left( \partial_t \noisemarg(\x_t^d | \x_1^d) - \partial_t \noisemarg(j^d | \x_1^d) \right)}{\mathcal{Z}_t^d \noisemarg(j^d | \x_1^d) } \noisemarg(\x_t^{1:D \backslash d} \odot j^d | \x_1^{1:D}) \\
    & - \frac{ \relu \left(\partial_t \noisemarg(j^d | \x_1^d) - \partial_t \noisemarg(\x_t^d | \x_1^d) \right)  }{\mathcal{Z}_t^d \noisemarg(\x_t^d | \x_1^d)} \noisemarg(\x_t^{1:D} | \x_1^{1:D})    \Bigg)
\end{align}
\begin{align}
    \text{RHS} = \sum_{d=1}^D \frac{1}{\mathcal{Z}_t^d} \noisemarg(\x_t^{1:D \backslash d} | \x_1^{1:D})  \sum_{j^d \neq i^d, \noisemarg(j^d | \x_1^d) > 0} \Bigg(& \relu \left( \partial_t \noisemarg(\x_t^d | \x_1^d) - \partial_t \noisemarg(j^d | \x_1^d) \right)\\
    & - \relu \left(\partial_t \noisemarg(j^d | \x_1^d) - \partial_t \noisemarg(\x_t^d | \x_1^d) \right) \Bigg)
\end{align}
\begin{align}
    \text{RHS} &= \sum_{d=1}^D \frac{1}{\mathcal{Z}_t^d} \noisemarg(\x_t^{1:D \backslash d} | \x_1^{1:D})  \sum_{j^d \neq i^d, \noisemarg(j^d | \x_1^d) > 0} \Bigg( \partial_t \noisemarg(\x_t^d | \x_1^d) - \partial_t \noisemarg(j^d | \x_1^d) \Bigg)\\
    &= \sum_{d=1}^D \noisemarg(\x_t^{1:D \backslash d} | \x_1^{1:D}) \partial_t \noisemarg(\x_t^d | \x_1^d)\\
    &= \partial_t \left( \prod_{d=1}^D \noisemarg(\x_t^d | \x_1^d )\right)\\
    &= \text{LHS}
\end{align}
where we have used the fact that $\noisemarg(\x_t^{1:D} | \x_1^{1:D}) = \prod_{d=1}^D \noisemarg(\x_t^d | \x_1^d)$.

For the case that there exists a $d'$ for which $\noisemarg(\x_t^{d'} | \x_1^{d'}) = 0$ we have $\partial_t \noisemarg(\x_t^{d'} | \x_1^{d'}) = 0$ by assumption. We first examine the LHS of equation \eqref{eq:apx_multidim_conditional_continuity} in this case.
\begin{align}
    \text{LHS} &= \partial_t \noisemarg(\x_t^{1:D} | \x_1^{1:D})\\
    &= \partial_t \left( \prod_{d=1}^D \noisemarg(\x_t^d | \x_1^d) \right) \\
    &= \sum_{d=1}^D \noisemarg(\x_t^{1:D \backslash d} | \x_1^{1:D \backslash d}) \partial_t \noisemarg(\x_t^d | \x_1^d) \\
    &= \noisemarg(\x_t^{1:D \backslash d'} | \x_1^{1:D \backslash d'}) \partial_t \noisemarg(\x_t^{d'} | \x_1^{d'}) + \sum_{d=1 \backslash d'}^D \noisemarg(\x_t^{1:D \backslash d} | \x_1^{1:D \backslash d}) \partial_t \noisemarg(\x_t^d | \x_1^d)\\
    &= \noisemarg(\x_t^{1:D \backslash d'} | \x_1^{1:D \backslash d'}) \underbrace{\partial_t \noisemarg(\x_t^{d'} | \x_1^{d'})}_{0} + \sum_{d=1 \backslash d'}^D \underbrace{\noisemarg(\x_t^{d'} | \x_1^{d'})}_{0} \noisemarg(\x_t^{1:D \backslash d,d'} | \x_1^{1:D \backslash d, d'}) \partial_t \noisemarg(\x_t^d | \x_1^d)\\
    &= 0
\end{align}
where we use $1:D \backslash d, d'$ to mean all dimensions except $d$ and $d'$. We now examine the RHS of equation \eqref{eq:apx_multidim_conditional_continuity}.
\begin{align}
   \text{RHS} &=  \sum_{d=1}^D \sum_{j^d} {\relurate_t}^d(j^d, \x_t^d | \x_1^d) \noisemarg(\x_t^{1:D \backslash d} \odot j^d | \x_1^{1:D}) \\
   &= \sum_{j^{d'}} {\relurate_t}^{d'}(j^{d'}, \x_t^{d'} | \x_1^{d'}) \noisemarg(\x_t^{1:D \backslash d'} \odot j^{d'} | \x_1^{1:D}) + \sum_{d=1 \backslash d'}^D \sum_{j^d} {\relurate_t}^d (j^d, \x_t^d | \x_1^d) \noisemarg(\x_t^{1:D \backslash d} \odot j^d | \x_1^{1:D}) \\
   &= \sum_{j^{d'}} \underbrace{{\relurate_t}^{d'}(j^{d'}, \x_t^{d'} | \x_1^{d'})}_{0} \noisemarg(\x_t^{1:D \backslash d'} \odot j^{d'} | \x_1^{1:D}) \\
   & \qquad + \sum_{d=1 \backslash d'}^D \sum_{j^d} {\relurate_t}^d (j^d, \x_t^d | \x_1^d) \underbrace{\noisemarg(\x_t^{d'} | \x_1^{d'})}_{0} \noisemarg(\x_t^{1:D \backslash d, d'} \odot j^d | \x_1^{1:D \backslash d'}) \\
   & = 0 \\
   & = \text{LHS}
\end{align}
where we have used the fact that ${\relurate_t}^{d'}(j^{d'}, \x_t^{d'} | \x_1^{d'})=0$ because $\noisemarg(j^{d'} | \x_1^{d'})=0$
Therefore, for both cases we have ${\relurate_t}$ satisfies the conditional Kolmogorov equation \eqref{eq:apx_multidim_conditional_continuity} and thus we have found a rate matrix that generates our desired conditional flow. The final step is to convert this rate matrix conditioned on $\x_1^{1:D}$ into an unconditional rate matrix that can be used for generative modeling. We first write down the unconditional multi-dimensional Kolmogorov equation
\begin{equation}
    \partial_t p_t(\x_t^{1:D}) = \sum_{j^{1:D}} R_t(j^{1:D}, \x_t^{1:D}) p_t(j^{1:D})
    \label{eq:apdx_unconditional_multidim_continuity}
\end{equation}

We now make the following assumption for the form of the unconditional rate matrix and verify that it indeed satisfies the unconditional multi-dimensional Kolmogorov equation, \eqref{eq:apdx_unconditional_multidim_continuity}.
\begin{equation}
    R_t(\x_t^{1:D}, j^{1:D}) = \sum_{d=1}^D \delta \{ \x_t^{1:D \backslash d}, j^{1:D \backslash d} \} R_t^d(\x_t^{1:D}, j^d)
    \label{eq:apdx_uncond_relu_rate}
\end{equation}
with
\begin{equation}
    R_t^d(\x_t^{1:D}, j^d) = \E_{p(\x_1^d | \x_t^{1:D})} \left[ {\relurate_t}^d(\x_t^d, j^d | \x_1^d) \right]
\end{equation}
with ${\relurate_t}^d(\x_t^d, j^d | \x_1^d)$ being given by \eqref{eq:apdx_multidim_relu_rate}. Substitute this form into \eqref{eq:apdx_unconditional_multidim_continuity}
\begin{align}
    \text{RHS} &= \sum_{j^{1:D}} \sum_{d=1}^D \delta \{ j^{1:D \backslash d}, \x_t^{1:D \backslash d} \} \E_{p(\x_1^d | j^{1:D})} \left[ {\relurate_t}^d(j^d, \x_t^d | \x_1^d) \right] p_t(j^{1:D})\\
    &= \sum_{d=1}^D \sum_{j^d} \E_{p(\x_1^d | \x_t^{1:D \backslash d} \odot j^d)} \left[ {\relurate_t}^d(j^d, \x_t^d | \x_1^d) \right] p_t(\x_t^{1:D \backslash d} \odot j^d)\\
    &= \sum_{d=1}^D \sum_{j^d} \sum_{\x_1^{d}} p(\x_1^d | \x_t^{1:D \backslash d} \odot j^d) {\relurate_t}^d(j^d, \x_t^d | \x_1^d) p_t(\x_t^{1:D \backslash d} \odot j^d)\\
    &= \sum_{d=1}^D \sum_{j^d} \sum_{\x_1^{d}} p(\x_1^d | \x_t^{1:D \backslash d} \odot j^d) {\relurate_t}^d(j^d, \x_t^d | \x_1^d) p_t(\x_t^{1:D \backslash d} \odot j^d) \underbrace{\sum_{\x_1^{1:D \backslash d}} p(\x_1^{1:D \backslash d} | \x_1^d, \x_t^{1:D \backslash d} \odot j^d )}_{=1}\\
    &= \sum_{d=1}^D \sum_{j^d} \sum_{\x_1^{1:D}} p(\x_1^{1:D} | \x_t^{1:D \backslash d} \odot j^d) {\relurate_t}^d(j^d, \x_t^d | \x_1^d) p_t(\x_t^{1:D \backslash d} \odot j^d) \\
    &= \sum_{d=1}^D \sum_{j^d} \sum_{\x_1^{1:D}} \pdata(\x_1^{1:D}) \noisemarg(\x_t^{1:D \backslash d} \odot j^d | \x_1^{1:D}) {\relurate_t}^d(j^d, \x_t^d | \x_1^d)\\
    &= \E_{\pdata(\x_1^{1:D})} \left[ \sum_{d=1}^D \sum_{j^d}  \noisemarg(\x_t^{1:D \backslash d} \odot j^d | \x_1^{1:D}) {\relurate_t}^d(j^d, \x_t^d | \x_1^d) \right] \\
    &= \E_{\pdata(\x_1^{1:D})} \left[ \partial_t \noisemarg(\x_t^{1:D} | \x_1^{1:D}) \right] \quad \text{by \eqref{eq:R_factorize_intermediate_continuity}} \\
    &= \partial_t p_t(\x_t^{1:D})\\
    &= \text{LHS}
\end{align}
where we have used \cref{eq:R_factorize_intermediate_continuity} with the fact that we know ${\relurate_t}$ given by \cref{eq:apdx_multidim_relu_rate} satisfies the conditional Kolmogorov equation \cref{eq:apx_multidim_conditional_continuity}. We have now verified that the rate given by \cref{eq:apdx_uncond_relu_rate} gives us our desired unconditional flow and we can use it for generative modeling.

\subsection{Training}
In order to approximate the true generative rate matrix given by equation \eqref{eq:apdx_uncond_relu_rate}, we need approximations to the denoising distributions in each dimension, $p(\x_1^d | \x_t^{1:D})$, for $d=1, \dots, D$. We can parameterize these conditionally independent $\x_1^d$ distributions through a neural network that outputs logits of shape $D \times S$ when given input $\x_t^{1:D}$ of shape $D$. We then apply a softmax to the logits to obtain approximate denoising probabilities $p_\theta(\x_1^d | \x_t^{1:D})$, $d = 1, \dots, D$ of shape $D \times S$. We learn the parameters of the neural network with the cross entropy loss for each dimension
\begin{equation}
    \mathcal{L}_{\mathrm{ce}} = \E_{\pdata(\x_1^{1:D}) \mathcal{U}(t; 0, 1) \noisemarg(\x_t^{1:D} | \x_1^{1:D})} \left[ \sum_{d=1}^D \log \denoise^\theta(\x_1^d | \x_t^{1:D}) \right]
\end{equation}
\subsection{Sampling}
The standard Euler step transition probability for our CTMC defined through our learned denoising model with time step $\Dt$ is 
\begin{align}
    p_{t+\Dt|t}(j^{1:D} | \x_t^{1:D}) &= \delta \{ \x_t^{1:D}, j^{1:D} \} + R_t^\theta(\x_t^{1:D}, j^{1:D}) \Dt\\
    \label{eq:multi_dim_euler_step}
    &= \delta \{ \x_t^{1:D}, j^{1:D} \} + \sum_{d=1}^D \delta \{ \x_t^{1:D \backslash d}, j^{1:D \backslash d} \} \E_{p_\theta(\x_1^{d} | \x_t^{1:D})} \left[ R_t^d(\x_t^d, j^d | \x_1^d) \right] \Dt
\end{align}
In this form, we would be unable to make transition steps that involve more than $1$ dimension changing at a time due to our factorized form for $R_t^\theta(\x_t^{1:D}, j^{1:D})$. To enable multiple dimensions to transition simultaneously in a single update step we can approximate the standard Euler transition step \eqref{eq:multi_dim_euler_step} with a factorized version $\tilde{p}_{t+\Dt | t}(j^{1:D} | \x_t^{1:D})$ with the following form
\begin{align}
    \tilde{p}_{t+\Dt | t}(j^{1:D} | \x_t^{1:D}) &= \prod_{d=1}^D \tilde{p}_{t+\Dt | t}^d(j^d | \x_t^{1:D}) \\
    &= \prod_{d=1}^D \left \{ \delta \{ \x_t^d, j^d \} + \E_{p_\theta(\x_1^d | \x_t^{1:D})} \left[ R_t^d (\x_t^d, j^d | \x_1^d) \right] \Dt \right\} \\
    &= \delta \{ \x_t^{1:D}, j^{1:D} \} + \sum_{d=1}^D \delta \{ \x_t^{1:D \backslash d }, j^{1:D \backslash d} \} \E_{p_\theta(\x_1^d | \x_t^{1:D}) } \left[ R_t^d(\x_t^d, j^d | \x_1^d) \right] \Dt + O(\Dt^2)
\end{align}
where we can see on the final line that $\tilde{p}_{t+\Dt | t}$ approximates $p_{t+\Dt | t}$ to first order. Sampling from $\tilde{p}_{t+\Dt | t}$ can be seen as taking an Euler step in each dimension independently for each simulation step.

We note this sampling method is similar to the tau-leaping method used in prior CTMC based approaches \cite{gillespie2001approximate, campbell2022continuous} however tau-leaping allows multiple jumps to be made in the same dimensions which is unsuitable for categorical data.

\subsection{Detailed Balance}
In this section we verify that if we achieve detailed balance individually and independently in each dimension, then our full dimensional process will also be in detailed balance.

Consider the multidimensional detailed balance equation
\begin{equation}
    \noisemarg(\x_t^{1:D} | \x_1^{1:D}) R_t(\x_t^{1:D}, j^{1:D} | \x_1^{1:D}) = \noisemarg(j^{1:D} | \x_1^{1:D}) R_t(j^{1:D}, \x_t^{1:D} | \x_1^{1:D})
\end{equation}
Now, substitute in our factorized forms for $R_t(\x_t^{1:D}, j^{1:D} | \x_1^{1:D})$ and $\noisemarg(\x_t^{1:D} | \x_1^{1:D})$
\begin{equation}
    \left( \prod_{d=1}^D \noisemarg(\x_t^d | \x_1^{d}) \right) \left( \sum_{d=1}^D \delta \{ \x_t^{\oneDd} , j^{\oneDd} \} R_t^d(\x_t^d , j^d | \x_1^{d)}) \right) = \left( \prod_{d=1}^D \noisemarg(j^d | \x_1^{d}) \right) \left( \sum_{d=1}^D \delta \{ j^{\oneDd}, \x_t^{\oneDd} \} R_t^d(j^d , \x_t^d | \x_1^{d}) \right)
\end{equation}
Now, both sides are $0$ for when $\x_t$ and $j$ differ in more than one dimension. Consider the case when they differ in exactly one dimension, call it $d$. The detailed balance equation simplifies to
\begin{equation}
\noisemarg(\x_t^d | \x_1^{d}) R_t^d(\x_t^d , j^d | \x_1^{d}) = \noisemarg(j^d | \x_1^{d}) R_t^d(j^d , \x_t^d | \x_1^{d}) 
\end{equation}
which we note is the standard single dimensional detailed balance equation for dimension $d$. Therefore, if our $R_t^d$ matrices are all in detailed balance with their respective $\noisemarg(\x_t^d | \x_1^{d})$ conditional marginals, then the full dimensional rate matrix $R_t(\x_t^{1:D}, j^{1:D} | \x_1^{1:D})$  will also be in detailed balance with the full dimensional conditional marginals $\noisemarg(\x_t^{1:D} | \x_1^{1:D})$.

\section{Implementation Details}
\label{sec:apdx_implementation_details}
In this section we provide concrete derivations of our \method method.
We use a masking process in \cref{sec:apdx_masking_example}, a uniform process in \cref{sec:apdx_uniform_example} and explore the general case for any given $\noisemarg$ in \cref{sec:apdx_implementation_general_case}. We also provide minimal PyTorch implementations for our training and sampling loops in each case. We will assume multi-dimensional data under the factorization assumptions listed in \cref{sec:factorization}.

Notebooks containing these minimal examples can be found at \url{https://github.com/andrew-cr/discrete_flow_models}.

\subsection{Masking Example}
\label{sec:apdx_masking_example}
Here, we assume the masking form for $\noisemarg$. We begin by writing this data conditional flow
\begin{align}
    \noisemarg(\x_t^{1:D} | \x_1^{1:D}) &= \prod_{d=1}^D \noisemarg(\x_t^d | \x_1^{d})\\
    &= \prod_{d=1}^D \left( t \kdelta { \x_t^d}{ \x_1^d } + (1-t) \kdelta { \x_t^d}{ M } \right)
\end{align}
This is the distribution we will use to train our denoising model $\denoise^\theta(\x_1^{1:D} | \x_t^{1:D})$. PyTorch code for the training loop is given in Listing \ref{lst:masking_training}

\hspace{0.8cm}
\begin{minipage}{\linewidth}
\begin{lstlisting}[style=pythonstyle, caption={Masking Training loop}, label={lst:masking_training}, numbers=none, linewidth=0.9\textwidth]
import torch
import torch.nn.functional as F


# Variables, B, D, S for batch size, number of dimensions and state space size respectively
# Assume we have a model that takes as input xt of shape (B, D) and time of shape (B,) and outputs x1 prediction logits of shape (B, D, S-1). We know the clean data contains no masks and hence we only need to output logits over the valid values.

mask_index = S - 1 # Assume the final state is the mask state

for x1 in dataset:
    # x1 has shape (B, D)
    optimizer.zero_grad()
    t = torch.rand((B,))
    xt = x1.clone()
    xt[torch.rand((B,D)) < (1 - t[:, None])] = mask_index
    logits = model(xt, t) # (B, D, S-1)
    x1[xt != mask_index] = -1 # Don't compute the loss on unmasked dimensions
    loss = F.cross_entropy(logits.transpose(1, 2), x1, reduction='mean', ignore_index=-1)
    loss.backward()
    optimizer.step()

\end{lstlisting}
\end{minipage}

We will also derive the form for ${\relurate_t}^d(i^d, j^d | \x_1^d)$. For this we need to find $\partial_t \noisemarg(\x_t^d | \x_1^d)$.
\begin{align}
    \partial_t \noisemarg(\x_t^d | \x_1^d) &= \partial_t \left( t \kdelta {\x_t^d}{\x_1^d} + (1-t) \kdelta{\x_t^d}{M} \right)\\
    &=\kdelta { \x_t^d}{ \x_1^d } - \kdelta { \x_t^d}{ M } 
\end{align}
We can now find ${\relurate_t}^d(\x_t^d, j^d | \x_1^d)$. When working with rate matrices in this section, we will always assume $\x_t^d \neq j^d$ and calculate the diagonal entries as $R_t(i, i) = - \sum_{j \neq i} R_t(i, j)$ later. We note that ${\relurate_t}^d(\x_t^d, j^d | \x_1^d) = 0$ for $\noisemarg(\x_t^d | \x_1^d) = 0$ or $\noisemarg(j^d | \x_1^d) = 0$. Further, our initial distribution $p_0(\x_0^{1:D}) = \prod_{d=1}^D \kdelta { \x_0^d}{ M}$. Therefore, at all points in our CTMC, $\x_t^d$ is only ever $M$ or $\x_1^d$. Furthermore, we only ever have to consider transitions to a $j^d$ that is either $j^d = M$ or $j^d = \x_1^d$. Now, for $\noisemarg(\x_t^d | \x_1^{1:D}) > 0$ and $\noisemarg(j^d | \x_1^{1:D}) > 0$ we have
\begin{align}
    {\relurate_t}^d(\x_t^d, j^d | \x_1^{d}) &= \frac{ \relu \left( \partial_t \noisemarg(j^d | \x_1^{d}) - \partial_t \noisemarg (\x_t^d | \x_1^{d}) \right)}{\mathcal{Z}_t^d \noisemarg(\x_t^d | \x_1^{d}) }\\
    &= \frac { \relu \left( \kdelta { j^d}{ \x_1^d } - \kdelta { j^d}{ M } - \kdelta { \x_t^d}{ \x_1^d } + \kdelta { \x_t^d}{ M } \right)  }{ 2 \left( t \kdelta { \x_t^d}{ \x_1^d } + (1-t) \kdelta { \x_t^d }{M } \right)} \\
    \label{eq:relu_rate_multi_dim_masking}
    &= \frac { 1 } { 1-t} \quad \text{for $ j^d = \x_1^d, \x_t^d = M$ and $0$ otherwise}
\end{align}
We note here that our calculation may not strictly be valid for exactly $t=0$ or $t=1$ but are valid for any $t \in (0, 1)$ and so we can simply ignore these edge cases, see \cref{sec:rate_match_marginal_proof} for further discussion. Now we find our unconditional rate matrix
\begin{align}
    R_t^{\theta d}(\x_t^{1:D}, j^d) &= \E_{\denoise^\theta(\x_1^d | \x_t^{1:D})} \left[ {\relurate_t}^d(\x_t^d, j^d | \x_1^d) \right] \\
    &= \E_{\denoise^\theta (\x_1^d | \x_t^{1:D}) } \left[ \frac{1}{1-t} \kdelta { j^d }{\x_1^d } \kdelta {\x_t^d}{ M } \right] \\
    \label{eq:masking_uncond_rate_d}
    &= \frac{ \denoise^\theta(\x_1^d = j^d | \x_t^{1:D}) }{1-t} \kdelta { \x_t^d}{ M }
\end{align}
Our transition step is then
\begin{equation}
    p_{t+\Dt | t}(j^d | \x_t^{1:D}) = \kdelta { j^d}{ \x_t^d } + R_t^{\theta d}(\x_t^{1:D}, j^d) \Dt
\end{equation}
For $j^d \neq \x_t^d$ this is
\begin{equation}
\label{eq:masking_state_detail_gen_kernel}
    p_{t+\Dt | t}(j^d | \x_t^{1:D}) = \Dt \frac{\denoise^\theta(\x_1^d = j^d | \x_t^{1:D})}{1-t} \kdelta {\x_t^d}{ M }
\end{equation}
For $j^d = \x_t^d$ this is
\begin{align}
    p_{t+\Dt | t}(j^d = \x_t^d | \x_t^{1:D}) &= 1 - \sum_{k \neq \x_t^d} p_{t+\Dt | t}(k | \x_t^{1:D}) \\
    &= 1 - \sum_{k \neq \x_t^d}  \Dt \frac{\denoise^\theta(\x_1^d = k | \x_t^{1:D})}{1-t} \kdelta {\x_t^d}{ M }\\
    &= 1 - \frac{\Dt}{1-t} \kdelta { \x_t^d}{ M }
\end{align}
where on the final line we have used the fact that when $\denoise^\theta(\x_1^d = M | \x_t^{1:D}) = 0$.
Therefore, if $\x_t^d = M$ then we have a $\frac{\dt}{1-t}$ chance of flipping to some unmasked state with the probabilities for the token to unmask to given by $\denoise^\theta(\x_1^d | \x_t^{1:D})$. If $\x_t^d \neq M$ (i.e. it has already been unmasked) then we simply stay in the current unmasked state.

Listing \ref{lst:masking_sample} shows PyTorch code that implements this sampling loop.

\hspace{0.8cm}
\begin{minipage}{\linewidth}
\begin{lstlisting}[style=pythonstyle, caption={Masking Sampling loop}, label={lst:masking_sample}, numbers=none, linewidth=0.9\linewidth]
import torch
import torch.nn.functional as F
from torch.distributions.categorical import Categorical


# Variables, B, D, S for batch size, number of dimensions and state space size respectively
# Assume we have a model that takes as input xt of shape (B, D) and time of shape (B,) and outputs x1 prediction logits of shape (B, D, S-1). We know the clean data contains no masks and hence we only need to output logits over the valid values.
t = 0.0
dt = 0.001
mask_index = S-1

xt = mask_index * torch.ones((B, D), dtype=torch.long)

while t < 1.0:
    logits = model(xt, t * torch.ones((B,))) # (B, D, S-1)
    x1_probs = F.softmax(logits, dim=-1) # (B, D, S-1)
    x1 = Categorical(x1_probs).sample() # (B, D)
    will_unmask = torch.rand((B, D)) < (dt / (1-t)) # (B, D)
    will_unmask = will_unmask * (xt == mask_index) # (B,D) only unmask currently masked positions
    xt[will_unmask] = x1[will_unmask]

    t += dt
\end{lstlisting}
\end{minipage}
\subsubsection{Detailed Balance}
\label{sec:apdx_masking_DB}
In order to expand our family of rate matrices that we can use at sampling time, we want to find a detailed balance rate matrix $\dbrate_t$ that satisfies the detailed balance equation 
\begin{equation}
    \noisemarg(i | \x_1) \dbrate_t(i, j | \x_1) = \noisemarg(j | \x_1) \dbrate_t(j, i | \x_1)
\end{equation}
We now have to make some assumptions on the form for $\dbrate_t$. With this masking noise a process that is in detailed balance will have some rate for transitions going from a mask state towards $\x_1$ and some rate for transitions going from $\x_1$ back towards the mask state. Such a rate would have the following form
\begin{equation}
    \dbrate_t(i, j | \x_1) = a_t \kdelta { i}{ \x_1 } \kdelta { j}{ M } + b_t \kdelta { i }{M } \kdelta {j}{ \x_1 }
\end{equation}
for some constants $a_t$ and $b_t$ that we must find. Substituting this into the detailed balance equation along with the masking interpolation form for $\noisemarg(\x_t | \x_1)$ gives
\begin{align}
    &\left( t \kdelta{ i}{ \x_1 } + (1-t) \kdelta { i}{ M } \right) \left(  a_t \kdelta { i}{ \x_1 } \kdelta { j }{ M } + b_t \kdelta { i }{ M } \kdelta {j } {\x_1 }\right) =\\
    &\left( t \kdelta{ j}{ \x_1 } + (1-t) \kdelta { j }{ M } \right) \left(  a_t \kdelta { j}{ \x_1 } \kdelta { i}{ M } + b_t \kdelta { j}{ M } \kdelta {i}{ \x_1 }  \right)
\end{align}
\begin{align}
    t a_t \kdelta { i}{ \x_1 } \kdelta { j}{ M } + (1-t) b_t \kdelta { i}{ M } \kdelta { j}{ \x_1 } = t a_t\kdelta { j}{ \x_1 } \kdelta { i}{M } + (1-t) b_t \kdelta {j}{M } \kdelta { i}{ \x_1 }
\end{align}
This equation must be true for all $i, j$. Pick $i = \x_1$ and $j = M$ to get
\begin{equation}
    t a_t = (1-t) b_t
\end{equation}
If we pick $ i = M$ and $j = \x_1$ then we would obtain the same equation and if we pick any other values for $i, j$ with $i \neq j$ then we would get $0 = 0$. Note that we will find $\dbrate_t$ for $i \neq j$ and then the value for $\dbrate_t (i, i)$ is simply calculated using $\dbrate_t(i, i) = - \sum_{j \neq i } \dbrate_t(i, j)$. Since we will obtain no more constraints on the values of $a_t$ and $b_t$, we will need to pick a value for one of them. We can simply set $a_t = \noise$ where $\noise$ is our stochasticity parameter since this value sets the rate at which points that are already at $\x_1$ will come off $\x_1$ and travel back to the mask state. This gives $b_t = \frac{\noise t}{1-t}$ and so for $i \neq j$,
\begin{equation}
    \dbrate_t(i, j | \x_1) = \noise \kdelta { i}{ \x_1 } \kdelta { j}{ M } + \frac{\noise t}{1-t} \kdelta { i }{ M } \kdelta { j }{\x_1 }.
\end{equation}
We now combine this rate with ${\relurate_t}^d$ that we calculated previously to find a new unconditional rate matrix with a variable amount of stochasticity.
\begin{align}
    R_t^{\theta d} (\x_t^{1:D}, j^d) &= \E_{\denoise^\theta(\x_1^d | \x_t^{1:D})} \left[ {\relurate_t}^d(\x_t^d, j^d | \x_1^d) + {\dbrate}_t^d(\x_t^d, j^d | \x_1^d) \right] \\
    &= \E_{\denoise^\theta(\x_1^{d} | \x_t^{1:D})} \left[ \frac{1}{1-t} \kdelta { j^d}{ \x_1^d } \kdelta { \x_t^d}{ M } + \noise \kdelta { \x_t^d}{ \x_1^d } \kdelta { j^d}{ M } + \frac{\noise t}{1-t} \kdelta { \x_t^d}{ M } \kdelta { j^d }{ \x_1^d } \right] \\
    &= \frac{\denoise^\theta(\x_1^d = j^d | \x_t^{1:D})}{1-t} \kdelta { \x_t^d }{ M } + \noise \denoise^\theta(\x_1^d = \x_t^d | \x_t^{1:D}) \kdelta { j^d}{ M } + \frac{\noise t}{1-t} \kdelta { \x_t^d}{ M } \denoise^\theta (\x_1^d = j^d | \x_t^{1:D})\\
    &= \frac{1+\noise t}{1-t} \denoise^\theta(\x_1^d = j^d | \x_t^{1:D}) \kdelta { \x_t^d}{ M } + \noise (1-\kdelta { \x_t^d}{ M }) \kdelta { j^d}{ M }
\end{align}
where on the final line we have used the fact that $\denoise^\theta(\x_1^d = \x_t^d | \x_t^{1:D}) = 0$ for $\x_t^d = M$ and $\denoise^\theta(\x_1^d = \x_t^d | \x_t^{1:D}) = 1$ when $\x_t^d \neq M$ because if a dimension is unmasked then it must be the true $\x_1$ value under our definition of $\noisemarg(\x_t | \x_1)$. We now find our transition probabilities
\begin{equation}
    p_{t+\Dt | t}(j^d | \x_t^{1:D}) = \kdelta { j^d}{ \x_t^d } + R_t^{\theta d}(\x_t^{1:D}, j^d) \Dt
\end{equation}
For $j^d \neq \x_t^d$,
\begin{equation}
    p_{t+\Dt | t}(j^d | \x_t^{1:D}) = \Dt \frac{1 + \noise t}{1-t} \denoise^\theta(\x_1^d = j^d | \x_t^{1:D}) \kdelta { \x_t^d}{ M } + \Dt \noise (1 - \kdelta { \x_t^d}{ M }) \kdelta { j^d}{ M }
\end{equation}
and for $j^d = \x_t^d$
\begin{align}
    p_{t+\Dt | t}(j^d &= \x_t^d | \x_t^{1:D}) = 1 - \sum_{k \neq \x_t^d} p_{t+\Dt | t}(k | \x_t^{1:D})\\
    &= 1 - \sum_{k \neq \x_t^d} \left( \Dt \frac{1 + \noise t}{1-t} \denoise^\theta(\x_1^d = k | \x_t^{1:D}) \kdelta { \x_t^d}{ M } + \Dt \noise (1 - \kdelta { \x_t^d}{ M }) \kdelta { k}{ M } \right) \\
    &= 1 - \Dt \frac{1+\noise t}{1-t} \kdelta {\x_t^d}{ M } - \Dt \noise (1-\kdelta { \x_t^d}{ M })
\end{align}
where again we have used the fact that $\denoise^\theta(\x_1^d = M | \x_t^{1:D}) = 0$. Inspecting $p_{t + \Dt | t}(j^d | \x_t^{1:D})$ for $j^d \neq \x_t^d$, we see that if $\x_t^d = M$ then we have an overall probability of unmasking of $\frac{1+\noise t}{1-t} \Dt$ and once we do unmask, the new value is drawn from $\denoise^\theta(\x_1^d | \x_t^{1:D})$. This is like before but now there is a bonus probability of unmasking of $\frac{\noise t}{1-t}$. When $\x_t^d \neq M$ then we have a probability of $\noise  \Dt$ of jumping back to the mask state. This creates a flux of states switching back and forth between masked and unmasked for $\noise  > 0$ hence why these processes are more `stochastic'. However, because when $\noise $ is increased we also increase the rate at which we unmask, the desired conditional flow $\noisemarg(\x_t | \x_1)$ is maintained for any value of $\noise $. Listing \ref{lst:masking_sample_with_noise} shows PyTorch code that implements sampling with this extra stochasticity.

\hspace{0.8cm}
\begin{minipage}{\linewidth}
\begin{lstlisting}[style=pythonstyle, caption={Masking sampling loop with noise}, label={lst:masking_sample_with_noise}, numbers=none, linewidth=0.9\linewidth]
import torch
import torch.nn.functional as F
from torch.distributions.categorical import Categorical


# Variables, B, D, S for batch size, number of dimensions and state space size respectively
# Assume we have a model that takes as input xt of shape (B, D) and time of shape (B,) and outputs x1 prediction logits of shape (B, D, S-1). We know the clean data contains no masks and hence we only need to output logits over the valid values.
t = 0.0
dt = 0.001
mask_index = S-1
N = 10 # Level of stochasticity

xt = mask_index * torch.ones((B, D), dtype=torch.long)

while t < 1.0:
    logits = model(xt, t * torch.ones((B,))) # (B, D, S-1)
    x1_probs = F.softmax(logits, dim=-1) # (B, D, S-1)
    x1 = Categorical(x1_probs).sample() # (B, D)
    will_unmask = torch.rand((B, D)) < (dt * (1 + N * t) / (1-t)) # (B, D)
    will_unmask = will_unmask * (xt == mask_index) # (B,D) only unmask currently masked positions
    will_mask = torch.rand((B, D)) < dt * N # (B, D)
    will_mask = will_mask * (xt != mask_index) # (B, D) only re-mask currently unmasked positions
    xt[will_unmask] = x1[will_unmask]
    t += dt
    if t < 1.0: # Don't re-mask on the final step
        xt[will_mask] = mask_index 

\end{lstlisting}
\end{minipage}

Our method has similarities to other discrete diffusion models when using this form for $\noisemarg$ and we clarify these links in \cref{sec:apdx_absorbing_state_link_to_d3pm}.



\subsubsection{Purity Sampling}
\label{sec:apdx_purity_sampling}
When using the masking form for $\noisemarg$ we can also easily implement a purity sampling scheme \cite{tang2022improved}. This sampling method decides which dimensions to unmask based on an estimate of the model confidence in that dimension's final value. Currently, our sampling method will uniformly at random choose which dimension to unmask. To improve upon this approach, purity sampling will instead rank dimensions based on which dimension has the highest model probability. More specifically, for each dimension we calculate a purity score for dimension $d$ defined as
\begin{equation}
    \text{purity}_d = \underset{\x_1^d}{\text{max}} \quad  \denoise^\theta(\x_1^d | \x_t^{1:D})
\end{equation}
For the next simulation step, we then decide how many dimensions should be unmasked. The number of dimensions to unmask is binomially distributed with probability of success $\frac{\Delta t}{1-t}$ and number of trials equal to the number of dimensions that are currently masked. Once we have sampled a number of dimensions to unmask from this binomial distribution, we then unmask that number of dimensions starting from the dimension with highest purity score, then the dimension with second highest purity score and so on. We only consider dimensions that are currently masked to be eligible for unmasking. When using $\noise >0$, the probability of success in our binomial distribution increases to $\Delta t \frac{1 + \noise t}{1-t}$ and so on average more dimensions get unmasked during each simulation step. At the end of each simulation step, we then remask a sample of randomly chosen dimensions which are uniformly chosen at random each with a probability $\Delta t \noise $ of being chosen.

\subsection{Uniform Example}
\label{sec:apdx_uniform_example}
In this section we walk through the derivation and implementation of \method when using the uniform based interpolation distribution. We start with the data conditional marginal distribution
\begin{align}
    \noisemarg(\x_t^{1:D} | \x_1^{1:D}) &= \prod_{d=1}^D \noisemarg(\x_t^d | \x_1^d) \\
    &= \prod_{d=1}^D \left( t \kdelta { \x_t^d}{ \x_1^d } + (1-t) \frac{1}{\statespace} \right)
\end{align}
This distribution is all that is needed to train the denoising model $\denoise^\theta(\x_1^{1:D} | \x_t^{1:D})$. We give PyTorch code for the training loop with the uniform interpolant in Listing \ref{lst:uniform_training}.

\hspace{0.8cm}
\begin{minipage}{\linewidth}
\begin{lstlisting}[style=pythonstyle, caption={Uniform training loop}, label={lst:uniform_training}, numbers=none, linewidth=0.9\linewidth]
import torch
import torch.nn.functional as F

# Variables, B, D, S for batch size, number of dimensions and state space size respectively
# Assume we have a model that takes as input xt of shape (B, D) and time of shape (B,) and outputs x1 prediction logits of shape (B, D, S). 

for x1 in dataset:
    # x1 has shape (B, D)
    optimizer.zero_grad()
    t = torch.rand((B,))
    xt = x1.clone()
    uniform_noise = torch.randint(0, S, (B, D))
    corrupt_mask = torch.rand((B, D)) < (1 - t[:, None])
    xt[corrupt_mask] = uniform_noise[corrupt_mask]
    logits = model(xt, t) # (B, D, S)
    loss = F.cross_entropy(logits.transpose(1,2), x1, reduction='mean')
    loss.backward()
    optimizer.step()
\end{lstlisting}
\end{minipage}

In order to sample our trained model, we will need to derive ${\relurate_t}^d(i^d, j^d | \x_1^d)$. The first step is to find $\partial_t \noisemarg(\x_t^d | \x_1^d)$,
\begin{align}
    \partial_t \noisemarg(\x_t^d | \x_1^d) &= \partial_t \left( t \kdelta { \x_t^d}{ \x_1^d } + (1-t) \frac{1}{\statespace} \right) \\
    &= \kdelta { \x_t^d}{ \x_1^d } - \frac{1}{\statespace} 
\end{align}
We will now find ${\relurate_t}^d(\x_t^d, j^d | \x_1^d)$. As before we will always assume $\x_t^d \neq j^d$ and calculate diagonal entries as needed using the relation $R_t(i, i) = - \sum_{j \neq i} R_t(i, j)$.
\begin{align}
    {\relurate_t}^d(\x_t^d, j^d | \x_1^d) &= \frac{ \relu \left( \partial_t \noisemarg(j^d | \x_1^d) - \partial_t \noisemarg(\x_t^d | \x_1^d) \right)  }{\mathcal{Z}_t^d \noisemarg(\x_t^d | \x_1^d)}\\
    &= \frac{ \relu \left( \kdelta { j^d}{ \x_1^d } - \frac{1}{\statespace} - \kdelta { \x_t^d}{ \x_1^d } + \frac{1}{\statespace}  \right) }{ \statespace \left( t \kdelta { \x_t^d}{ \x_1^d } + (1-t) \frac{1}{\statespace} \right)   } \\
    &= \frac{ \relu \left( \kdelta { j^d}{ \x_1^d } - \kdelta { \x_t^d}{ \x_1^d } \right) } { S ( t \kdelta { \x_t^d}{ \x_1^d } + (1-t) \frac{1}{S}}
\end{align}
The only non-zero value is when $j^d = \x_1^d$ and $\x_t^d \neq \x_1^d$ and so ${\relurate_t}^d(\x_t^d, j^d | \x_1^d)$ is
\begin{equation}
    {\relurate_t}^d(\x_t^d, j^d | \x_1^d) = \frac{1}{1-t} \kdelta { j^d}{ \x_1^d } (1-\kdelta { \x_t^d}{ \x_1^d })
\end{equation}
We can now find the unconditional rate matrix, still assuming $\x_t^d \neq j^d$
\begin{align}
    R_t^{\theta d} (\x_t^{1:D}, j^d ) &= \E_{\denoise^\theta(\x_1^d | \x_t^{1:D}) } \left[ {\relurate_t}^d(\x_t^d, j^d | \x_1^d) \right] \\
    &= \E_{\denoise^\theta(\x_1^d | \x_t^{1:D})} \left[ \frac{1}{1-t} \kdelta { j^d }{\x_1^d } (1-\kdelta { \x_t^d}{ \x_1^d }) \right] \\
    &= \frac{1}{1-t} \denoise^\theta(\x_1^d = j^d | \x_t^{1:D})
\end{align}
Our transition step is
\begin{equation}
    p_{t+\Dt | t} (j^d | \x_t^{1:D}) = \kdelta { j^d }{ \x_t^d } + R_t^{\theta d}(\x_t^{1:D}, j^d) \Dt
\end{equation}
For $j^d \neq \x_t^d$ this is
\begin{equation}
    p_{t+\Dt | t}(j^d | \x_t^{1:D}) = \frac{\Dt}{1-t} \denoise^\theta(\x_1^d = j^d | \x_t^{1:D}) 
\end{equation}
and for $j^d = \x_t^d$ this is
\begin{align}
    p_{t+\Dt |t }(j^d = \x_t^d | \x_t^{1:D}) &= 1 - \sum_{k \neq \x_t^d} p_{t+\Dt | t}(k | \x_t^{1:D})\\
    &= 1 - \sum_{k \neq \x_t^d} \frac{\Dt}{1-t} \denoise^\theta(\x_1^d = k | \x_t^{1:D})\\
    &= 1 - \frac{\Dt}{1-t} \left( 1 - \denoise^\theta(\x_1^d = \x_t^d | \x_t^{1:D}) \right)
\end{align}

Listing \ref{lst:uniform_sample} shows PyTorch code that implements this sampling loop.

\hspace{0.8cm}
\begin{minipage}{\linewidth}
\begin{lstlisting}[style=pythonstyle, caption={Uniform sampling loop}, label={lst:uniform_sample}, numbers=none, linewidth=0.9\linewidth]
import torch
import torch.nn.functional as F
from torch.distributions.categorical import Categorical


# Variables, B, D, S for batch size, number of dimensions and state space size respectively
# Assume we have a model that takes as input xt of shape (B, D) and time of shape (B,) and outputs x1 prediction logits of shape (B, D, S).
t = 0.0
dt = 0.001

xt = torch.randint(0, S, (B, D))
while t < 1.0:
    logits = model(xt, t * torch.ones((B,))) # (B, D, S)
    x1_probs = F.softmax(logits, dim=-1) # (B, D, S)

    # Calculate the off-diagonal step probabilities
    step_probs = ((dt / (1-t)) * x1_probs).clamp(max=1.0) # (B, D, S)

    # Calculate the on-diagnoal step probabilities
    # 1) Zero out the diagonal entries
    step_probs.scatter_(-1, xt[:, :, None], 0.0)
    # 2) Calculate the diagonal entries such that the probability row sums to 1
    step_probs.scatter_(-1, xt[:, :, None], (1.0 - step_probs.sum(dim=-1, keepdim=True)).clamp(min=0.0)) 

    xt = Categorical(step_probs).sample() # (B, D)

    t += dt
\end{lstlisting}
\end{minipage}

\subsubsection{Detailed Balance}
Here we derive the form of $\dbrate_t$ for the uniform interpolant case which we can use to vary the stochasticity of sampling. $\dbrate_t$ satisfies the detailed balance equation
\begin{equation}
    \noisemarg(i | \x_1) \dbrate_t(i, j | \x_1) = \noisemarg(j | \x_1) \dbrate_t(j, i | \x_1)
\end{equation}
We now make some assumptions for the form of $\dbrate_t$. We will assume there will be some rate of transitions from $\x_1$ back towards a random other state and a rate towards $\x_1$ in order to cancel out this effect and achieve detailed balance. We note there are other choices for detailed balance, some of which we explore in \cref{sec:apdx_cont_time_diff_comparison}. We will again be assuming $i \neq j$ in the following calculations.
\begin{equation}
    \dbrate_t(i, j | \x_1) = a_t \kdelta { i}{ \x_1 } + b_t \kdelta { j}{ \x_1 }
\end{equation}
We have parameterized $\dbrate_t$ with some time-dependent constants $a_t$ and $b_t$. Substituting this into the detailed balance equation gives
\begin{equation}
    \left( t \kdelta { i}{ \x_1 } + (1-t) \frac{1}{\statespace} \right) \left( a_t \kdelta { i}{ \x_1 } + b_t \kdelta { j}{ \x_1 }\right) =  \left( t \kdelta { j}{ \x_1 } + (1-t) \frac{1}{\statespace} \right) \left( a_t \kdelta { j}{ \x_1 } + b_t \kdelta { i}{ \x_1 }\right)
\end{equation}
Now, this equation must be true for any $i \neq j$. Pick $i = \x_1$ and $j \neq \x_1$ to get
\begin{equation}
    \left( t + (1-t) \frac{1}{\statespace} \right) a_t = (1-t) \frac{1}{\statespace} b_t
\end{equation}
\begin{align}
    b_t &= a_t \frac{t + (1-t) \frac{1}{\statespace}}{ (1-t) \frac{1}{\statespace}} \\
    &= a_t \frac{St + 1 - t}{1 -t} \label{eq:uniform_detailed_balance}
\end{align}
We would obtain the same equation if we were to instead pick $i \neq \x_1$ and $j = \x_1$. Therefore we have to fix one of $a_t$ or $b_t$. If we want a stochasticity level of $\noise $ then we can set $a_t = \noise $ which is the rate at which points that are at the clean data come back off the clean datapoint. $b_t$ can then be found from equation \eqref{eq:uniform_detailed_balance}. This gives a form for $\dbrate_t$ of 
\begin{equation}
\dbrate_t(i, j | \x_1) = \noise  \kdelta { i}{ \x_1 } + \noise  \frac{\statespace t + 1 - t}{1-t} \kdelta { j}{ \x_1 }
\end{equation}
This can now be combined with ${\relurate_t}^d$ to create a new unconditional rate matrix with a variable amount of stochasticity.
\begin{align}
    R_t^{\theta d}(\x_t^{1:D}, j^d) &= \E_{\denoise^\theta(\x_1^d | \x_t^{1:D})} \left[ {\relurate_t}^d(\x_t^d, j^d | \x_1^d) + {\dbrate}_t^d(\x_t^d, j^d | \x_1^d) \right] \\
    &= \E_{\denoise^\theta(\x_1^d | \x_t^{1:D}) } \left[ \frac{1}{1-t} \kdelta {j^d}{ \x_1^d } (1-\kdelta { \x_t^d}{ \x_1^d }) + \noise  \kdelta { \x_t^d}{ \x_1^d } + \noise  \frac{\statespace t + 1 - t}{1-t} \kdelta { j^d }{ \x_1^d } \right] \\
    &= \E_{\denoise^\theta(\x_1^d | \x_t^{1:D}) } \left[ \frac{1 + \noise  + \noise  (\statespace - 1) t}{1-t} \kdelta {j^d}{ \x_1^d} (1-\kdelta {\x_t^d} { \x_1^d }) + \noise  \kdelta {\x_t^d}{ \x_1^d }  \right] \\
    &= \frac{1 + \noise  + \noise (\statespace - 1) t}{1-t} \denoise^\theta(\x_1^d = j^d | \x_t^{1:D}) + \noise  \denoise^\theta(\x_1^d = \x_t^d | \x_t^{1:D})
\end{align}
We can interpret this rate, with the first term being the rate at which we should transition to states that are predicted to correspond to the clean data. The second term is a `noise term' which creates transitions away from the current state if it is predicted to correspond to the final clean data. The first term then has additional weighting as $\noise $ is increased to counter act this effect. The effect of the stochasticity is then to create a flux going on and off the predicted final clean state during generation. We now find our transition probabilities
\begin{equation}
    p_{t+\Dt | t}(j^d | \x_t^{1:D}) = \kdelta { j^d }{\x_t^d } + R_t^{\theta d}(\x_t^{1:D}, j^d) \Dt
\end{equation}
For $j^d \neq \x_t^d$,
\begin{equation}
    p_{t+\Dt | t}(j^d | \x_t^{1:D}) = \Dt \frac{1 + \noise  + \noise (\statespace - 1) t}{1 - t} \denoise^\theta(\x_1^d = j^d | \x_t^{1:D}) + \Dt \noise  \denoise^\theta(\x_1^d = \x_t^d | \x_t^{1:D})
\end{equation}
We can find $p_{t+\Dt | t}(j^d | \x_t^{1:D})$ for $j^d = \x_t^d$ programmatically as before by requiring that the probability vector sum to $1$.
Listing \ref{lst:uniform_sample_with_noise} shows the implementation for the uniform interpolant with noise.

\hspace{0.8cm}
\begin{minipage}{\linewidth}
\begin{lstlisting}[style=pythonstyle, caption={Uniform sampling loop with noise}, label={lst:uniform_sample_with_noise}, numbers=none, linewidth=0.9\linewidth]
import torch
import torch.nn.functional as F
import torch.distributions.categorical import Categorical

# Variables, B, D, S for batch size, number of dimensions and state space size respectively
# Assume we have a model that takes as input xt of shape (B, D) and time of shape (B,) and outputs x1 prediction logits of shape (B, D, S).

t = 0.0
dt = 0.001
noise = 1

xt = torch.randint(0, S, (B, D))

while t < 1.0:
    logits = model(xt, t * torch.ones((B,))) # (B, D, S)
    x1_probs = F.softmax(logits, dim=-1) # (B, D, S)
    x1_probs_at_xt = torch.gather(x1_probs, -1, xt[:, :, None]) # (B, D, 1)

    # Don't add noise on the final step
    if t + dt < 1.0:
        N = noise
    else:
        N = 0

    # Calculate the off-diagonal step probabilities
    step_probs = (
        dt * ((1 + N + N * (S - 1) * t ) / (1-t)) * x1_probs + 
        dt * N * x1_probs_at_xt
    ).clamp(max=1.0) # (B, D, S)

    # Calculate the on-diagnoal step probabilities
    # 1) Zero out the diagonal entries
    step_probs.scatter_(-1, xt[:, :, None], 0.0)
    # 2) Calculate the diagonal entries such that the probability row sums to 1
    step_probs.scatter_(-1, xt[:, :, None], (1.0 - step_probs.sum(dim=-1, keepdim=True)).clamp(min=0.0)) 

    xt = Categorical(step_probs).sample() # (B, D)

    t += dt
\end{lstlisting}
\end{minipage}

\subsection{General Case}
\label{sec:apdx_implementation_general_case}
We now describe the training and sampling loop for a general conditional flow $\noisemarg(\x_t | \x_1)$.
We require this interpolant to be factorized, $\noisemarg(\x_t^{1:D} | \x_1^{1:D}) = \prod_{d=1}^D \noisemarg(\x_t^d | \x_1^d)$, be differentiable and have $\noisemarg(j^d | \x_1^d) = 0 \implies \partial_t \noisemarg(j^d | \x_1^d) = 0$. We assume that we have access to functions that can sample from $\noisemarg(\x_t | \x_1)$, evaluate $\noisemarg(\x_t | \x_1)$ and evaluate $\partial_t \noisemarg(\x_t | \x_1)$. Our training loop consists of sampling data, sampling $\x_t \sim \noisemarg(\x_t | \x_1)$ and training with the cross entropy loss, see Listing \ref{lst:general_training}.

\hspace{0.8cm}
\begin{minipage}{\linewidth}
\begin{lstlisting}[style=pythonstyle, caption={General training loop}, label={lst:general_training}, numbers=none, linewidth=0.9\linewidth]
import torch
import torch.nn.functional as F

# Variables, B, D, S for batch size, number of dimensions and state space size respectively
# Assume we have a model that takes as input xt of shape (B, D) and time of shape (B,) and outputs x1 prediction logits of shape (B, D, S). 

def sample_p_xt_g_x1(x1, t):
    # x1 (B, D)
    # t (B,)
    # Returns xt (B, D)

for x1 in dataset:
    # x1 has shape (B, D)
    optimizer.zero_grad()
    t = torch.rand((B,))
    xt = sample_p_xt_g_x1(x1, t)
    logits = model(xt, t) # (B, D, S)
    loss = F.cross_entropy(logits.transpose(1,2), x1, reduction='mean')
    loss.backward()
    optimizer.step()
\end{lstlisting}
\end{minipage}

Now for sampling we can programmatically calculate ${\relurate_t}^d(\x_t^d, j^d | \x_1^d)$ using \cref{eq:apdx_multidim_relu_rate}. It may not be possible to analytically calculate the expectation with respect to $\denoise^\theta(\x_1^{1:D} | \x_t^{1:D})$ but we note that our Euler step is still valid if we instead take a sample from $\denoise^\theta(\x_1^{1:D} | \x_t^{1:D})$ and substitute into $R_t^d(\x_t^d, j^d | \x_1^d)$, see \cref{sec:apdx_ctmc_sampling}. We assume access further to a function that can produce samples from the prior distribution $\pnoise$ corresponding to the chosen $\noisemarg$. We provide the general case sampling loop in Listing \ref{lst:general_sampling}.

\hspace{0.8cm}
\begin{minipage}{\linewidth}
\begin{lstlisting}[style=pythonstyle, caption={General sampling loop}, label={lst:general_sampling}, numbers=none, linewidth=0.9\linewidth]
def dt_p_xt_g_xt(x1, t):
    # x1 (B, D)
    # t float
    # returns (B, D, S) for varying x_t value

def p_xt_g_x1(x1, t):
    # x1 (B, D)
    # t float
    # returns (B, D, S) for varying x_t value

def sample_prior(num_samples, D):
    # num_samples, D both integer
    # returns prior sample of shape (num_samples, D)

t = 0.0
dt = 0.001
num_samples = 1000
xt = sample_prior(num_samples, D)

while t < 1.0:
    logits = model(xt, t * torch.ones((num_samples,))) # (B, D, S)
    x1_probs = F.softmax(logits, dim=-1) # (B, D, S)
    x1 = Categorical(x1_probs).sample() # (B, D)

    # Calculate R_t^*
    # For p(x_t | x_1) > 0 and p(j | x_1) > 0
    # R_t^*(x_t, j | x_1) = Relu( dtp(j | x_1) - dtp(x_t | x_1)) / (Z_t * p(x_t | x_1))
    # For p(x_t | x_1) = 0 or p(j | x_1) = 0 we have R_t^* = 0
    # We will ignore issues with diagnoal entries as later on we will set
    # diagnoal probabilities such that the row sums to one later on.

    dt_p_vals = dt_p_xt_g_xt(x1, t) # (B, D, S)
    dt_p_vals_at_xt = dt_p_vals.gather(-1, xt[:, :, None]).squeeze(-1) # (B, D)

    # Numerator of R_t^*
    R_t_numer = F.relu(dt_p_vals - dt_p_vals_at_xt[:, :, None]) # (B, D, S)

    pt_vals = p_xt_g_x1(x1, t) # (B, D, S)
    Z_t = torch.count_nonzero(pt_vals, dim=-1) # (B, D)
    pt_vals_at_xt = pt_vals.gather(-1, xt[:, :, None]).squeeze(-1) # (B, D)

    # Denominator of R_t^*
    R_t_denom = Z_t * pt_vals_at_xt # (B, D)
    
    R_t = R_t_numer / R_t_denom[:, :, None] # (B, D, S)

    # Set p(x_t | x_1) = 0 or p(j | x_1) = 0 cases to zero
    R_t[ (pt_vals_at_xt == 0.0)[:, :, None].repeat(1, 1, S)] = 0.0
    R_t[ pt_vals == 0.0] = 0.0

    # Calculate the off-diagonal step probabilities
    step_probs = (R_t * dt).clamp(max=1.0) # (B, D, S)

    # Calculate the on-diagnoal step probabilities
    # 1) Zero out the diagonal entries
    step_probs.scatter_(-1, xt[:, :, None], 0.0)
    # 2) Calculate the diagonal entries such that the probability row sums to 1
    step_probs.scatter_(-1, xt[:, :, None], (1.0 - step_probs.sum(dim=-1, keepdim=True)).clamp(min=0.0)) 

    xt = Categorical(step_probs).sample() # (B, D)
    t += dt
\end{lstlisting}
\end{minipage}

\subsubsection{Detailed Balance}
There are many ways one could solve the detailed balance equation for $\dbrate_t$ as the choice will depend on what kinds of noise are desirable to include in the generative process. A baseline example of how you could solve the detailed balance equation for generate $\noisemarg(\x_t | \x_1)$ is to note
\begin{align}
    \dbrate_t(i, j | \x_1) \noisemarg(i | \x_1) &= \dbrate_t(j, i | \x_1) \noisemarg(j | \x_1) \\
    \frac{\dbrate_t(i, j | \x_1)}{\dbrate_t(j, i | \x_1)} &= \frac{\noisemarg(i | \x_1)}{\noisemarg(j | \x_1)} 
\end{align}
which gives a relation between the diagonal elements of $\dbrate_t$. As a first choice we could simply set the upper triangular section of $\dbrate_t$ to $1$ and set the lower triangular part to the ratio $\frac{\noisemarg(i | \x_1)}{\noisemarg(j | \x_1)}$ which would satisfy detailed balance.

\section{CTMC Sampling Methods}
\label{sec:apdx_ctmc_sampling}
In the main text, our sampling algorithm \cref{alg:sampling} first constructs the unconditional rate matrix $R_t^\theta(\x_t, j) = \E_{\denoise^\theta(\x_1 | \x_t)} \left[ R_t(\x_t, j | \x_1) \right]$ and then samples the next state from the Euler step,
\begin{equation}
    \x_{t + \Delta t} \sim \text{Cat} \left( \kdelta {\x_t}{\x_{t+\Delta t}} + R_t^\theta(\x_t, \x_{t+\Delta t}) \Delta t \right).
\end{equation}
The form of this update means that we don't necessarily need to calculate the full expectation over $R_t(\x_t, j | \x_1)$. We can simply sample $\x_1$ from $\denoise^\theta(\x_1 | \x_t)$ and then plug this sample into $R_t(\x_t, j | \x_1)$ which we then use in the Euler update. To see that this strategy still samples from the same distribution over $\x_{t+\Delta t}$, we can write the distribution over $\x_{t+\Delta t}$ as $p_{t+\Delta t|t}$,
\begin{align}
    p_{t+\Dt | t}(\x_{t+\Delta t} | \x_t) &= \kdelta{\x_t}{\x_{t+\Delta t}} + \E_{\denoise^\theta(\x_1 | \x_t)} \left[ R_t (\x_t, \x_{t+\Delta t} | \x_1) \right] \Dt \\
    &= \E_{\denoise^\theta(\x_1 | \x_t) } \left[ \kdelta{\x_t}{ \x_{t+\Delta t}} + R_t(\x_t, \x_{t+\Delta t} | \x_1) \Dt \right] \\
    &= \sum_{\x_1} \denoise^\theta(\x_1 | \x_t) p_{t+\Dt | t}( \x_{t+\Delta t}| \x_1, \x_t)
\end{align}
where
\begin{equation}
    p_{t+\Dt | t}( \x_{t+\Delta t}| \x_1, \x_t) \vcentcolon =  \kdelta{\x_t}{ \x_{t+\Delta t}} + R_t(\x_t, j | \x_1) \Delta t
\end{equation}
and so $p_{t+\Dt | t}( \x_{t+\Delta t}| \x_t)$ can be seen as the marginal of joint distribution $\denoise^\theta(\x_1 | \x_t) p_{t+\Dt | t}( \x_{t+\Delta t} | \x_1, \x_t)$. Therefore, to produce a sample $\x_{t+\Delta t}$ from $p_{t+\Dt | t}( \x_{t+\Delta t}| \x_t)$, we can instead sample $\x_1, \x_{t+\Delta t}$ from the joint distribution $\denoise^\theta(\x_1 | \x_t^{1:D}) p_{t+\Dt | t}( \x_{t+\Delta t} | \x_1, \x_t)$, and take only the $\x_{t+\Delta t}$ part of this joint sample.

Another method to simulate a CTMC is $\tau$-leaping, \cite{gillespie2001approximate, campbell2022continuous} which allows multiple jumps to be made both across dimensions and within each dimension. Multiple jumps within a single dimension does not make sense for categorical data where there is no ordering, however, it can be useful for ordinal data such as a discretized image where the $\tau$-leaping update allows multiple jumps to be applied at once to cover a larger distance. To calculate a $\tau$-leaping update, a Poisson random variable needs to be drawn with the rate matrix giving the rate parameter. Therefore, for this type of update, the full unconditional $R_t^\theta(\x_t, j)$ would need to be calculated.

We finally note that there is a body of work creating CTMC samplers for generative models \citep{sun2022score, lou2023discrete} that may be faster to simulate than the standard Euler step. In this work, we focus on framework simplicity, not optimizing for sampling speed and leave application of these approaches as future work.

\section{Comparison with Discrete Diffusion Models}
\label{sec:apdx_comparison_diff}
In this section we clarify the relationship between \method and classical discrete diffusion models. In \cref{sec:apdx_cont_time_diff_comparison} we compare to continuous time models using the uniform corruption process as an example. In \cref{sec:apdx_absorbing_state_link_to_d3pm} we compare to discrete time models using the masking process as the example.

\subsection{Continuous Time Discrete Diffusion Models}
\label{sec:apdx_cont_time_diff_comparison}
Here we compare to continuous time discrete diffusion models \citep{campbell2022continuous} using the uniform corruption process as an example.
In this section, we will assume $t=0$ is pure noise and $t=1$ is clean data which we note is a flipped definition of time to classical diffusion models to aid in our comparison with DFMs.

For discrete diffusion, we first specify a corruption process and then approximate its time reversal to give us the generative process. Our corruption process will evolve from $t=1$ back to time $t=0$. It will be specified using a rate matrix $R_t$. In order to make calculation of $\noisemarg(\x_t | \x_1)$, $R_t$ needs to be of a special form, namely $R_t = \beta(t) R_b$ where $\beta(t)$ is a time dependent scalar function and $R_b$ is a base rate matrix that can be decomposed using the eigendecomposition $R_b = Q \Lambda Q^{-1}$. For uniform corruption, we can set $R_b = \mathds{1} \mathds{1} ^\top - S \mathbb{I}$ where $\mathds{1}$ is a vector of all $1$'s. We will now assume $S=3$ so we can carry out all calculations explicitly.

We have $R_b = Q \Lambda Q^{-1}$ with
\begin{equation}
    Q = \begin{bmatrix}
        -1 & -1 & 1 \\
        0 & 1 & 1 \\
        1 & 0 & 1
    \end{bmatrix} \quad
    \Lambda = \begin{bmatrix}
        -3 & 0 & 0 \\
        0 & -3 & 0 \\
        0 & 0 & 0
    \end{bmatrix} \quad
    Q^{-1} = \begin{bmatrix}
        -\frac{1}{3} & -\frac{1}{3} & \frac{2}{3} \\
        - \frac{1}{3} & \frac{2}{3} & - \frac{1}{3} \\
        \frac{1}{3} & \frac{1}{3} & \frac{1}{3}
    \end{bmatrix}
\end{equation}
To calculate $\noisemarg(\x_t | \x_1)$ we can use the equation
\begin{equation}
    P_t = Q \exp \left( \Lambda \int_{1}^{t} \beta(s) \dd s \right) Q^{-1}
\end{equation}
where $(P_t)_{ij} = \noisemarg(\x_t = j | \x_1 = i)$ and $\exp$ is the element wise exponential. By the symmetry of the problem, we can infer that $\noisemarg(\x_t = j | \x_1 = i)$ will have only two possible values. Either $j = i$ and we are finding the probability of staying at $i$, or $j \neq i$ and we are finding the probability of having left $i$, and since uniform corruption treats all states equally, these will be same quantities for any starting state and any state $j \neq i$. So to find our schedule we just need to consider one element of the matrix $P_t$. Let us consider an off-diagonal element $i \neq j$ of $P_t$, which will have probability
\begin{equation}
    (P_t)_{ij} = \frac{1}{3} \left( 1 - \exp \left( -3 \int_1^t \beta(s) \dd s \right) \right), \quad i \neq j
\end{equation}
We will try and match this to the simple linear schedule that we have had as our running example in the explanation of \method.
\begin{align}
    &\frac{1}{3} \left( 1 - \exp \left( -3 \int_t^1 \beta(s) \dd s \right) \right) = \frac{1}{3} ( 1 - t) \\
    & \implies \beta(t) = \frac{1}{3t} 
\end{align}
Therefore, we have found that a corruption rate matrix of $R_t = \frac{1}{3t} \left( \mathds{1} \mathds{1}^\top - 3 \mathbb{I} \right)$ gives a conditional flow of $\noisemarg(\x_t | \x_1) = t \kdelta{\x_t}{\x_1} + (1-t) \frac{1}{3}$.

The next step in a discrete diffusion model is to find the time reversed rate matrix $\hat{R}_t$ which gives a CTMC that runs in the opposite direction to $R_t$ and generates novel data from noise. Here $\hat{R}_t$ is running from time $t=0$ at noise towards clean data at $t=1$. From \citet{campbell2022continuous}, we have
\begin{equation}
    \hat{R}_t(i, j) = \sum_{\x_1} R_t(j, i) \frac{\noisemarg(j | \x_1)}{\noisemarg(i | \x_1)} \denoise(\x_1 | i) \quad i \neq j
\end{equation}
\newcommand{\Rdiff}{R^{\text{diff}}}
We notice a similarity to the \method equations, where the generative rate is the expectation of a quantity with respect to $\denoise(\x_1 | i)$. Indeed we now show that $R_t(j, i) \frac{\noisemarg(j | \x_1)}{\noisemarg(i | \x_1)}$ is a $\x_1$ conditioned rate matrix $\Rdiff_t(i, j | \x_1)$ that achieves the conditional flow $\noisemarg(i | \x_1)$. Consider the Kolmogorov equation
\begin{equation}
    \partial_t \noisemarg(i | \x_1) = \sum_{j \neq i} \Rdiff_t(j, i | \x_1) \noisemarg(j | \x_1) - \sum_{j \neq i} \Rdiff_t(i, j | \x_1) \noisemarg(i | \x_1)
\end{equation}
Substitute in our form for $\Rdiff_t$
\begin{align}
    \text{RHS} &= \sum_{j \neq i} R_t(i, j) \frac{\noisemarg(i | \x_1)}{\noisemarg(j | \x_1)} \noisemarg(j | \x_1) - \sum_{j \neq i} R_t(j, i) \frac{\noisemarg(j | \x_1)}{\noisemarg(i | \x_1)} \noisemarg(i | \x_1) \\
    &= \sum_{j \neq i} R_t(i, j) \noisemarg(i | \x_1) - \sum_{j \neq i} R_t(j, i) \noisemarg(j | \x_1)\\
    &= - \left[ \sum_{j \neq i} R_t(j, i) \noisemarg(j | \x_1) - \sum_{j \neq i} R_t(i, j) \noisemarg(i | \x_1) \right] \\
    &= - \left[ - \partial_t \noisemarg(i | \x_1) \right] \\
    &= \text{LHS} 
\end{align}
where on the second to last line we have used the fact that the corruption matrix $R_t(i, j)$ when started at $p_{t=1}(\x_t | \x_1) = \kdelta{\x_t}{\x_1}$ will evolve the marginals according to $\noisemarg(\x_t | \x_1)$ because this is how we derived $\noisemarg(\x_t | \x_1)$ in the first place. Note $R_t$ runs in the reverse direction hence the negative sign.

Therefore, the diffusion framework has made an implicit choice for $R_t(i, j | \x_1) = \Rdiff_t(i, j | \x_1)$ and this choice is made at training time. We now show on our uniform noise example that $\Rdiff_t$ is simply $\relurate_t + \dbrate_t$ for a specific choice of $\dbrate_t$.

Firstly, we write out the explicit form for $\Rdiff_t$ using $\Rdiff_t(i, j | \x_1) = R_t(j, i) \frac{\noisemarg(j | \x_1)}{\noisemarg(i | \x_1)}$, $R_t = \frac{1}{3t} \left(\mathds{1}\mathds{1}^\top - 3 \mathbb{I} \right)$ and $\noisemarg(i | \x_1) = t \kdelta{\x_t}{\x_1} + (1-t) \frac{1}{3}$.
\begin{equation}
    \Rdiff_t = \frac{1}{3t} \begin{bmatrix}
        -1 - \frac{1+2t}{1-t} & \frac{1 + 2t}{1-t} & 1 \\
        \frac{1-t}{1+2t} & -2 \frac{1-t}{1+2t} & \frac{1-t}{1+2t} \\
        1 & \frac{1+2t}{1-t} & -1 - \frac{1+2t}{1-t}
    \end{bmatrix}
\end{equation}
We will now find $\dbrate_t$ such that $\Rdiff_t = \relurate_t + \dbrate_t$. We will need a slightly more general form for $\dbrate_t$ than was previously derived for the uniform noise case. We will have
\begin{equation}
    \dbrate_t(i, j | \x_1) = a_t \kdelta{i}{\x_1} + b_t \kdelta{j}{\x_1} + c_t( 1 - \kdelta{i}{\x_1})(1 - \kdelta{j}{\x_1})
\end{equation}
Using the detailed balance equation, $\noisemarg(i | \x_1) \dbrate_t(i, j | \x_1) = \noisemarg(j | \x_1) \dbrate_t(j, i | \x_1)$, we find that we need
\begin{equation}
    a_t = \frac{(1-t) \frac{1}{3} b_t}{t + (1-t) \frac{1}{3}}
\end{equation}
with $b_t$ and $c_t$ being fully flexible (provided they are positive). Using the form for $\relurate_t(i, j | \x_1) = \frac{1}{1-t} \kdelta{j}{\x_1}(1-\kdelta{i}{\x_1})$ that we derived in Appendix \ref{sec:apdx_uniform_example} we have
\begin{equation}
    \relurate_t + \dbrate_t = \begin{bmatrix}
        -\frac{1}{1-t} - b_t - c_t & \frac{1}{1-t} + b_t & c_t \\
        \frac{(1-t) \frac{1}{3} b_t}{t + (1-t) \frac{1}{3}} & -2 \frac{(1-t) \frac{1}{3} b_t}{t + (1-t) \frac{1}{3}} & \frac{(1-t) \frac{1}{3} b_t}{t + (1-t) \frac{1}{3}} \\
        c_t & \frac{1}{1-t} + b_t & -c_t - \frac{1}{1-t} - b_t
    \end{bmatrix}
\end{equation}
which is equal to $\Rdiff_t$ if we have $b_t = c_t = \frac{1}{3t}$.

In summary, we have found that classical discrete diffusion models make an implicit choice for $R_t(i, j | \x_1)$ which corresponds to a certain level of stochasticity in the CTMC and that the choice is made at training time because the rate matrix is used in the ELBO objective. Further, we have seen it is much harder to derive the noise schedule $\noisemarg(\x_t | \x_1)$ in classical discrete diffusion models due to the need to be able to apply the matrix exponential to $R_t$. In \method, we can simply write down the $\noisemarg(\x_t | \x_1)$ noise schedule we want and we are not restricted in having to pick $R_t$ that are amenable to matrix exponentiation. We also get to choose any $R_t(i, j |\x_1)$ at test time rather than being fixed to the implicit choice of $\Rdiff_t$.

\subsection{Discrete Time Discrete Diffusion Models}
\label{sec:apdx_absorbing_state_link_to_d3pm}
In this section we will clarify the link to the discrete time diffusion method D3PM \citep{austin2021structured} when using the masking process for both methods. Here, we will use the convention from \citet{austin2021structured} of using $t=0$ for clean data and $t=T$ for noise.

We will first summarize the key results from \cite{austin2021structured} when using the absorbing state process which is a different name for a masking type process (the mask is the absorbing state). $t$ can take on any discrete value in $t \in \{0, 1, \dots, T \}$. The diffusion model is first defined using a noising transition kernel
\begin{equation}
    p(\x_{t} | \x_{t-1}) = \begin{cases}
        1 & \text{if }  \x_t = \x_{t-1} = M \\
        1 - \beta_t & \text{if } \x_t = \x_{t-1} \neq M \\
        \beta_t & \text{if } \x_t = M , \x_{t-1} \neq M
    \end{cases}
\end{equation}
From this transition kernel, we can then calculate the noise marginals, $p(\x_t | \x_0)$
\begin{equation}
    p(\x_t | \x_0) = \left( 1 - \prod_{k \leq t} (1 - \beta_k) \right) \kdelta { \x_t}{ M } + \left( \prod_{k \leq t} (1-\beta_k) \right) \kdelta { \x_t}{ \x_0 }
\end{equation}
We then define our generative reverse process as
\begin{equation}
    p_\theta(\x_{t-1} | \x_t) = \sum_{\x_0} p(\x_{t-1} | \x_t, \x_0) p_\theta(\x_0 | \x_t)
\end{equation}
where $p_\theta(\x_0 | \x_t)$ is the learned denoising model. Note how this is similar to our generative process, $R_t^\theta(\x_t, j) = \E_{p_\theta(\x_1 | \x_t)}\left[ R_t(\x_t, j | \x_1) \right]$ where now $p(\x_{t-1} | \x_t, \x_0)$ is the transition kernel for the clean data conditioned process. We then create our generative model by taking the expectation of this conditional kernel with respect to our denoising model.

Continuing with the D3PM example using the absorbing state process, we obtain the following form for $p_\theta(\x_{t-1} | \x_t)$
\begin{equation}
    p_\theta(\x_{t-1} | \x_t) = \begin{cases}
        \frac{1 - \prod_{k \leq t-1} (1 - \beta_k)  }{1 - \prod_{k \leq t} (1 - \beta_k)} & \text{if } \x_t = \x_{t-1} = M \\
        \frac{ \beta_t \prod_{k \leq t-1} (1 - \beta_k)  }{ 1 - \prod_{k \leq t} (1 - \beta_k) } p_\theta(\x_0 = \x_{t-1} | \x_t) & \text{if } \x_t = M, \x_{t-1} \neq M \\
        \kdelta{\x_{t-1}}{\x_t} & \text{if } \x_t \neq M
    \end{cases}
\end{equation}
When we set $\beta_t = \frac{1}{T- t + 1}$, we obtain a linear noise schedule giving
\begin{equation}
    p_\theta(\x_{t-1} | \x_t) = \begin{cases}
        \left( 1 - \frac{1}{t} \right) & \text{if } \x_t = \x_{t-1} = M \\
        \frac{1}{t} p_\theta(\x_0 = \x_{t-1} | \x_t) & \text{if } \x_t = M, \x_{t-1} \neq M \\
        \kdelta{\x_{t-1}}{\x_t} & \text{if } \x_t \neq M
    \end{cases} 
\end{equation}
Now, let us define $\xi \vcentcolon = \frac{t}{T}$ to be the proportion that the process is through the total number of time steps. $\xi \in [0, 1]$ and if we consider it to be an analogue of our continuous time variable, we can see that the original discretization steps of D3PM correspond to a discretization of the $[0, 1]$ interval with timesteps of $\Delta t = \frac{1}{T}$. Substituting these definitions into our update step gives,
\begin{equation}
    p_\theta(\x_{t-1} | \x_t) = \begin{cases}
        \left( 1 - \frac{\Dt}{\xi} \right) & \text{if } \x_t = \x_{t-1} = M \\
        \frac{1}{\xi} \Dt p_\theta(\x_0 = \x_{t-1} | \x_t) & \text{if } \x_t = M, \x_{t-1} \neq M \\
        \kdelta{\x_{t-1}}{\x_t} & \text{if } \x_t \neq M
    \end{cases} 
\end{equation}
Now we can see a clear comparison to \cref{eq:masking_state_detail_gen_kernel} noting the flipped definition of time. With our method we can pick any time discretization at test time because our method has been trained on all possible $t \in [0, 1]$. We also derive $\dbrate_t$ for the masking case which is not included in the prior D3PM framework. For training we note that the ELBO also simplifies down to a weighted cross entropy term for D3PM as noted by \cite{austin2021structured} and is also the case in our framework, see Appendix \ref{sec:apdx_Lelbo_for_masking}.

\section{Text Experiment Details}
\label{sec:apdx_text_experiment_details}
Code for our text experiments can be found at \url{https://github.com/andrew-cr/discrete_flow_models}.

For our denoising network we use the transformer architecture \cite{vaswani2017attention} as implemented in the nanoGPT repository, \url{https://github.com/karpathy/nanoGPT}. We generally follow the smallest GPT2 architecture \cite{radford2019language}.
At the input we have our input tokens $\x_t$ of shape $B, D$ where $B$ is the batch size and $D$ is the number of dimensions i.e. the sequence length, our time $t$ of shape $B$, and, if we are self-conditioning, prior $\x_1$ prediction tokens of shape $B, D$. We embed the $\x_t$ and $\x_1$ tokens using the same learned embedding, and use a model embedding size of $768$ resulting in tensors of shape $B, D, 768$.
We embed the position of each token using a learned embedding for each possible position.
We embed the time $t$, using Transformer sinusoidal embeddings following \cite{ho2020denoising}.
We train all our diffusion models with self-conditioning \cite{chen2022analog}. To input the prior $\x_1$ prediction, we stack the $\x_t$ embedded tensor $B, D, 768$ with the $\x_1$ prior prediction token tensor $B, D, 768$ to obtain a tensor of shape $B, D, 768 \times 2$. We then apply a linear layer to project down to the model embedding dimension resulting in a tensor of shape $B, D, 768$.
Before applying transformer blocks, we add together the $\x_t$ (and $\x_1$) embedding tensor, the position embedding and the time embedding to obtain the final $B, D, 768$ input tensor.

The transformer stack consists of $12$ transformer blocks, each block consisting of a LayerNorm, SelfAttention, LayerNorm, MLP stack. Within our SelfAttention block, we use $12$ heads and apply Qk-layernorm \cite{dehghani2023scaling} to our query and key values as we observed this improved convergence. Our MLP blocks consist of a $768 \rightarrow 768 \times 4$ linear layer, followed by a GELU activation, followed by a $768 \times 4 \rightarrow 768$ linear layer. We do not apply dropout. Our output layer consists of a linear head with output dimension $28$. We use $28$ token categories, $26$ lower case letters, a whitespace character and a mask token. The model outputs logits of shape $B, D, 28$ which we then apply a softmax to, to obtain $p_\theta(\x_1 | \x_t)$ probabilities.

The dataset text8 is $100$MB of text data from English Wikipedia. The text is all converted to lower case letters, i.e. capital letters are converted to lower case and numbers are written as text, i.e. $8$ becomes `eight'.

During training, we use a batch size of $256$ with $8$ gradient accumulation steps. We train on sequences of length $256$. The model is therefore trained on $524,288$ tokens per gradient update.
To train self-conditioning, on $50\%$ of training iterations, we input prior $\x_1$ prediction tokens as all masks so that the model learns to be able to predict $\x_1$ without any prior information. On the other $50\%$ of training iterations, we perform two model forward passes. We first predict $\x_1$ using masks as the prior $\x_1$ tokens to obtain an initial set of $p_\theta(\x_1 | \x_t)$ logits. We then sample from the initial $p_\theta(\x_1 | \x_t)$ distribution to obtain predicted $\x_1$ tokens. We then feed these tokens back into the model through the self-conditioning input and predict the $\x_1$ logits once more. These logits are then used in the loss. We only back propagate through the second forward pass of the model.

When training the D3PM model, we found that the default cross entropy weighting of $1/t$ (with a flipped definition of time) resulted in poor convergence and so we applied an equal weighting of the cross entropy across time to be consistent with the DFM loss.

We train our D3PM and DFM models for $750$k iterations on 4 Nvidia A40 GPUs using a learning rate of $10^{-4}$ and $1000$ linear warm up steps. We use a cosine decay schedule after the initial warm up towards a minimum learning rate of $10^{-5}$ which would be reached at $1$M iterations.
We use the AdamW optimizer \cite{loshchilov2017decoupled} with weight decay parameter $0.1$.
We monitor the validation loss throughout training. Validation loss continues to drop throughout training and we evaluate the final $750k$ model in our experiments.
When training the autoregressive model, we use the same architecture but find that it begins to overfit the data much faster than the diffusion based models. After $3500$ iterations the validation loss begins to increase and so we use the model with minimum validation loss in our evaluations. This is consistent with findings that autoregressive models require much less compute to converge than diffusion based models \cite{gulrajani2023likelihood}.

We use the masking interpolant in our \method with linear interpolant, as described in Appendix \ref{sec:apdx_masking_example}.
For D3PM, we use the absorbing state corruption process, the links to the \method process are described in Appendix \ref{sec:apdx_absorbing_state_link_to_d3pm}.

For the SEDD baseline, we train two models from scratch using the provided code for $750k$ training iterations with an effective batch size of $2048$ to be consistent with the DFM and D3PM training runs. All other parameters are left at their default values with the transformer using the same hidden size, number of blocks and number of layers as our other runs.

For evaluation, we sample the \method with $\Dt = 0.001$. We simulate up to $t = 0.98$ and then for any remaining tokens that are still mask, we set them to the most likely token under the model's denoising distribution, $p_\theta(\x_1 | \x_t)$. We stop simulating at $t=0.98$ to avoid any singularities similar to how diffusion models stop near $t=0$.
For D3PM we train with $1000$ timesteps to match \method.

For each temperature setting applied to the $p_\theta(\x_1 | \x_t)$ logits, we sample $512$ sequences all of length $256$ tokens. We then calculate the negative log-likelihood assigned to each sequence using GPT-J-6B \cite{gpt-j} and the BPE tokenizer \cite{radford2019language}. We then average the negative log-likelihoods over the $512$ sequences. The sample entropy is calculated by first tokenizing with the BPE tokenizer and then calculating the entropy as $\sum_i - p_i \log p_i$ where $p_i$ is the empirical probability of token $i$ estimated using the full set of $512$ samples. Tokens for which $p_i = 0$ are not included in the sum. For reference, the dataset achieves a negative log-likelihood of 4.2 as measured by GPT-J-6B.

\subsection{Stochasticity Sweep}
\begin{figure}
    \centering
    \includegraphics[width=10cm]{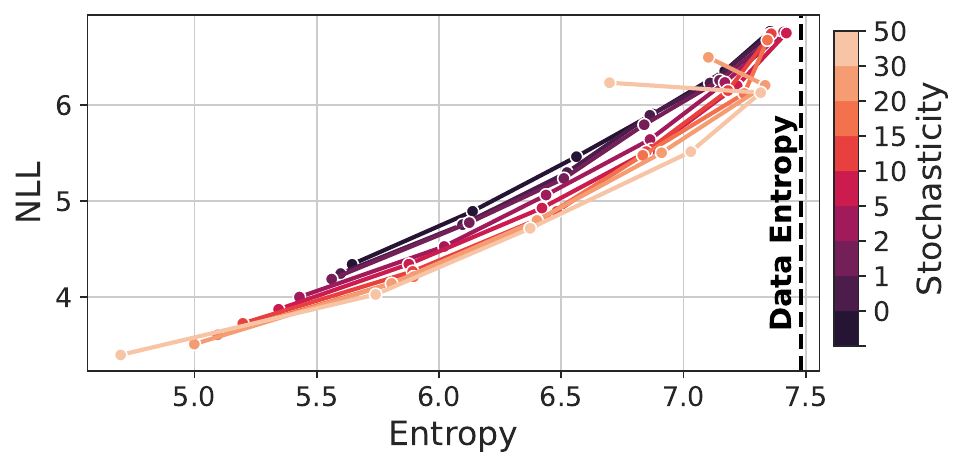}
    \caption{Curves in Entropy-NLL space for varying noise levels used during sampling. For each noise level, the temperature applied to the logits of the $p_\theta(\x_1 | \x_t)$ prediction is varied over values $0.5,0.6,0.7,0.8,0.9,1.0$.}
    \label{fig:text8_noise_sweep}
\end{figure}

Here we examine the effect of the noise level $\noise $ on the sample quality of generations from our \method method. We follow the follow the same procedure as before but vary $\noise $ with values $\noise =0,1,2,5,10,15,20,30,50$. We plot the results in Figure \ref{fig:text8_noise_sweep}. We find that generally, as the noise level increases, we lower our negative log-likelihood. However, we find that if the noise level is increased too much, then degenerate behaviour can occur, for example when $\noise =50$, at high logit temperatures the negative log-likelihood increases and the sample entropy decreases away from the dataset. Observing the samples, we find that the model generates incoherent text at this point. We find that the intermediate noise level $\noise =15$ provides good sample quality whilst avoiding this behaviour.

\subsection{Model Log-Likelihoods}
Here we calculate bounds on the log likelihood $\log p_\theta(\x_1)$ that the model assigns to the test set of text8. We use \cref{eq:masking_elbo} to calculate this bound. We compare our log-likelihoods to other discrete diffusion style methods in terms of bits-per-character in \cref{table:bpc_comparison}, reporting the numbers from \citet{lou2023discrete}. We find that DFM achieves a similar BPC to previous masking style diffusion models with the recent work of \citet{lou2023discrete} achieving the lowest BPC.

\begin{table}[]
\centering
\begin{tabular}{lc}
\toprule
Method & BPC \\
\midrule
DFM $\eta = 0$ & $\leq 1.41$ \\
Multinomial Diffusion \cite{hoogeboom2021argmax} & $\leq 1.72$ \\
MAC \cite{shih2022training} & $\leq 1.40$ \\
BFN \cite{graves2023bayesian} & $\leq 1.41$ \\
D3PM Uniform \cite{austin2021structured} & $\leq 1.61$ \\
D3PM Absorb \cite{austin2021structured} & $\leq 1.45$ \\
SEDD Uniform \cite{lou2023discrete} & $\leq 1.41$ \\
SEDD Absorb \cite{lou2023discrete} & $\leq 1.32$ \\
\bottomrule
\end{tabular}
\caption{Model log-likelihoods computed on the test set of text8 in bits-per-character (BPC).}
\label{table:bpc_comparison}
\end{table}

\subsection{Example Text Generations}
\begin{figure}
    \centering
    \includegraphics[width=10cm]{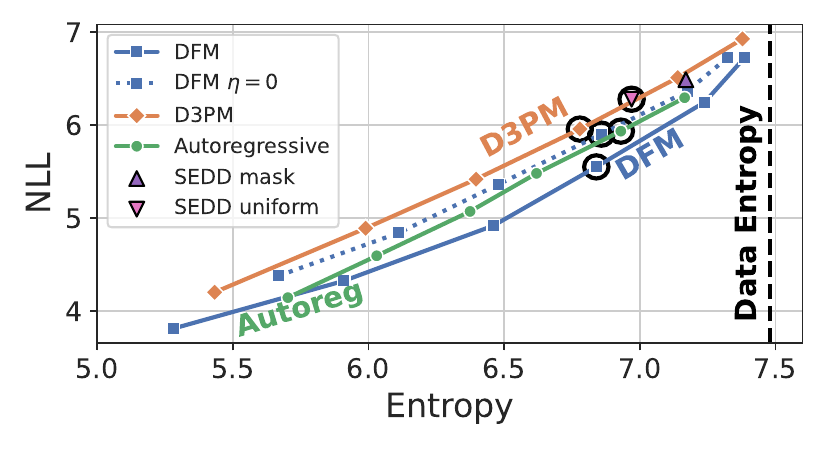}
    \caption{The temperature settings for each model for which we will examine sample generations. The selected temperature setting is highlighted with a black circle.}
    \label{fig:text8_metrics_with_circles}
\end{figure}

In this section we provide non cherry picked generations from the text models. For each model we have swept over the temperature applied to the logits and it would be impractical to include examples for all models for all temperature settings. Instead, we select one temperature setting for each model such that the samples have similar entropy but vary in negative log-likelihood. We show the selected temperature settings in Figure \ref{fig:text8_metrics_with_circles}. For SEDD the method does not have a temperature setting and we select the corruption style that is closest in entropy-NLL space to the other methods.\\

\fbox{\begin{minipage}{0.9\textwidth}
    \textbf{SEDD Uniform}\\
    \textbf{Samples:}
    \begin{quote}
change status regional courses and markets especially canada sport in canada and tennessee in canada the official light offered an newspaper licence named liu beijing world s main neighbouring fan site was sugar the only man with major historical works lic

ions of extension one or at least four subsets of a unique value of one example all these extensions heard of the function is called real line the implementation comes distributed with a continuous input extension to input and two classes the diagram emplo

 of physics the radio atomic institutions independently the eastern united states followed into four six countries norway thus was the father of the university of gloucester but while also the father of germany can we also announce the coexistence of limit
    \end{quote}
\end{minipage}}
\fbox{\begin{minipage}{0.9\textwidth}
    \textbf{D3PM Temperature 0.8}\\
    \textbf{Samples:}
    \begin{quote}
ved as a personal area to form the five counties of the area and a country with their own which is usually called paris gietgothic can also lead an area to work in divisions over a pileur as in the name of man the bears have over the last two years from th

 one five zero zero zero zero press money to present this to a meschasel linear industrial base ulse sudan expanded its economy and accounts for car prices and two eight five more than one zero zero of the largest industrial inventions over the world were 
 
eed alternatively as human being and the anti constitutionalay doctrines a particular example of the concept is one reason for human rights or as in certain regions there is a double constitution more recognized region of europe in this region the glass an
    \end{quote}
\end{minipage}}
\fbox{\begin{minipage}{0.9\textwidth}
    \textbf{\method $\noise =0$ Temperature 0.8}\\
    \textbf{Samples:}
    \begin{quote}
        ed era vol seven one nine one one december one nine six one junju that s one of nine one one country page of love footnote pages charles s feadman history of the red sea corea one nine nine one red sea vol one january one nine nine seven flying profiles ch
        
 allowes the vectores to be composed as systems of data for example no machine is a computer one would do not know where there are undirected storage of other data storage particularly the computer science eve to substitute such a based data that is one of
 
me io the plate n and feminine along the trail to change the amount of naturated information in the start tape selective figurative memory the mind is determined by the second net on the string c with two buttons the tag retes the header when queued the se
    \end{quote}
\end{minipage}}
\fbox{\begin{minipage}{0.9\textwidth}
    \textbf{Autoregressive Temperature 0.9}\\
    \textbf{Samples:}
    \begin{quote}
licklyn american football coach to holy roman emperor and roman stories radio and facilities in the u s civil rights movement the dc circuit collection of the witches leading the transissario times and spinoffs to american cartoonists cartoonist kyle marci

the british one one eight four minamoto minister or al di nortello ministries son of monte oise klepe which chose to give up its character on the go he was known to publish a wade of white performances started in one eight five one kleine married the gigan

mausoleum in one eight one six alabama was engaged by a large scale as we know alabama migration the palace of westminsters and proceeded to father she also learned to speak with the abramic mouth of the space the replica was apparently built de provence g
    \end{quote}
\end{minipage}}
\fbox{\begin{minipage}{0.9\textwidth}
    \textbf{\method $\noise =15$ Temperature 0.8}\\
    \textbf{Samples:}
    \begin{quote}
e curous greek by alexander van hep ven see archaic origin of the word cupola another meaning suggests that the word kupola is the latin word cupei kupolum old german derived from the latin word for the river the name comes from a latin word for tree with 

es so balloonists refine this combination specifically to preserve your own land in the runner both examples of clean steering creating agout like rods that produced successful rods and for the end the first few pistols compact stunt a musical setting mult

by reign over agassi is considered a greatest match by the day he will never play and will continue to be imitated agassi can play determinedly but agassi would always look to the victorious build he should not finish years going up to then that he would b
    \end{quote}
\end{minipage}}

\section{Protein Generation Experiment Details}
\label{sec:apdx_protein_codesign_details}
We present additional experiment details and results for protein generation with \protmodel.

Code for \protmodel and experiments can be found at \url{https://github.com/jasonkyuyim/multiflow}

\subsection{Experimental Details}
\paragraph{Model Architecture.}
We use an architecture modified from the FrameDiff architecture from \citet{yim2023se}. This architecture consists of Invariant Point Attention \citep{jumper2021highly} combined with transformer blocks, we refer to \citet{yim2023se} for in-depth details.
We modify this network architecture by increasing the number of network blocks to 8, increasing the number of transformer layers within each block to 4, decreasing the number of hidden channels used in the IPA calculation to 16, removing skip connections and removing psi-angle prediction. To enable our model to output logits for the discrete $\denoise^\theta(\x_1| \x_t)$ distribution, we add an output 3 layer MLP with the same embedding size as the main trunk. This results in a network with 21.8M parameters.

In \citet{yim2023se}, psi-angle prediction is used to infer the location of oxygen atoms, however, this position can be inferred to high accuracy using prior knowledge of the backbone structure of proteins, following \cite{yim2023fast}.

When training with our $t,\tilde{t}$ objective that enables the model to learn over different relative levels of corruption between structure and sequence, 10\% of the time we set $t=1$ and draw $\tilde{t} \sim \mathcal{U}(0, 1)$ and 10\% of the time we set $\tilde{t}=1$ and draw $t \sim \mathcal{U}(0, 1)$. The remaining 80\% of the time we draw both $t$ and $\tilde{t}$ independently from $t, \tilde{t} \sim \mathcal{U}(0,1)$.

\subsection{Additional \protmodel Results}
\label{sec:additional_cogen_results}
We show results of \protmodel across more lengths than done in \cref{sec:sec:codesign} and show that using the ESMFold oracle for data distillation still gives improved performance when we switch the evaluation oracle to AlphaFold2.

\paragraph{Main metrics with standard error.}
\cref{tab:metrics_with_err} presents results of \cref{table:designability} with standard error. We see our interpretation of the results do not change.

\begin{sidewaystable}[h!]
    \centering
    \begin{tabular}{lccccccccc}
        \toprule
        \textbf{Method} & \multicolumn{3}{c}{\textbf{Co-design 1}} & \multicolumn{3}{c}{\textbf{PMPNN 8}} & \multicolumn{3}{c}{\textbf{PMPNN 1}} \\
        \cmidrule(lr){2-4} \cmidrule(lr){5-7} \cmidrule(lr){8-10}
        & \textbf{Des.} & \textbf{Div.} & \textbf{Nov.} & \textbf{Des.} & \textbf{Div.} & \textbf{Nov.} & \textbf{Des.} & \textbf{Div.} & \textbf{Nov.} \\
        \midrule
        Protpardelle & 0.04 (0.01) & 5 (0.3) & 0.69 (0.01) & 0.9 (0.01) & 47 (0.7) & 0.59 (0.01) & 0.63 (0.01) & 38 (2.7) & 0.60 (0.01) \\
        ProteinGenerator & 0.37 (0.01) & 35 (2.6) & 0.69 (0.01) & 0.89 (0.00) & 75 (1.5) & 0.65 (0.02) & 0.78 (0.02) & 64 (4.6) & 0.66 (0.01) \\
        RFdiffusion & & & & 0.87 (0.02) & 158 (3.7) & 0.63 (0.01) & 0.66 (0.02) & 111 (5.9) & 0.64 (0.01) \\
        Cogen & 0.86 (0.01) & 143 (6.4) & 0.61 (0.02) & 0.99 (0.00) & 156 (7.7) & 0.61 (0.01) & 0.88 (0.01) & 143 (6.4) & 0.61 (0.02) \\
        Cogen w/o distillation & 0.42 (0.02) & 72 (2.4) & 0.62 (0.01) & 0.89 (0.00) & 126 (1.0) & 0.62 (0.02) & 0.71 (0.02) & 101 (7.1) & 0.63 (0.02) \\
        Cogen w/o sequence & & & & 0.99 (0.00) & 116 (1.0) & 0.63 (0.01) & 0.86 (0.01) & 97 (1.5) & 0.62 (0.02) \\
        \bottomrule
    \end{tabular}
    \caption{Performance metrics for various methods across different configurations.}
    \label{tab:metrics_with_err}
\end{sidewaystable}

\begin{figure}[h]
    \centering
    \includegraphics[width=15cm]{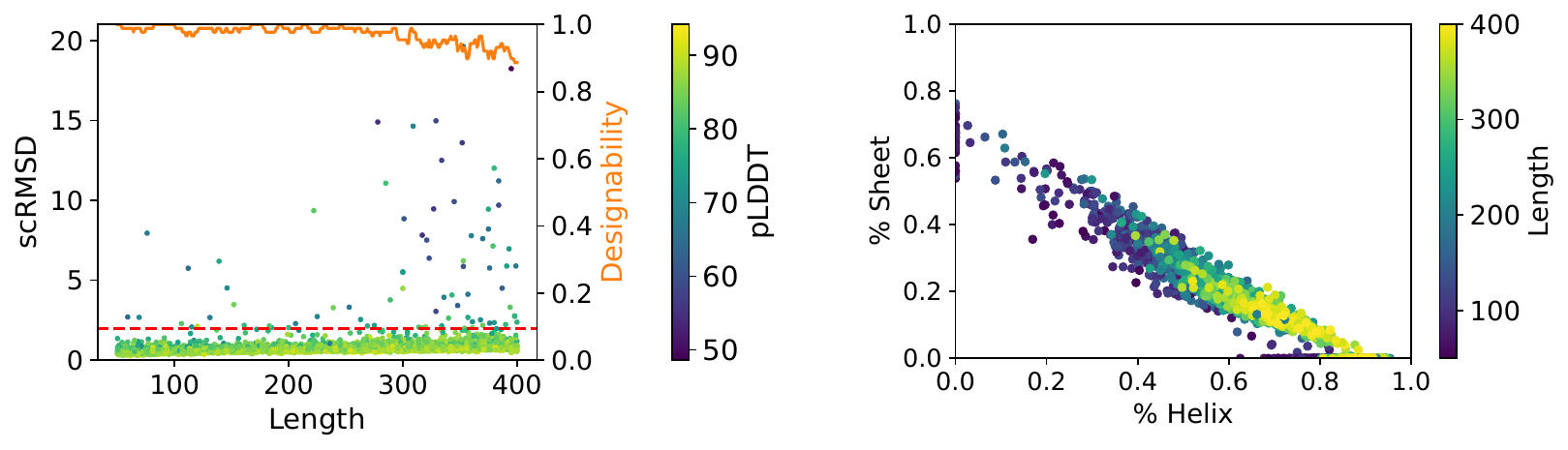}
    \vspace{-0.3cm}
    \caption{
    \textbf{\protmodel results on Protpardelle benchmark.}
    (\textbf{Left}) PMPNN 8 scRMSD and designability versus length. 
    Designability is computed as the proportion of samples that have $\text{scRMSD} < 2 \text{\AA{}}$ within a sliding window of size 11.
    Average pLDDT as computed by ESMFold for each sample is plotted as the colour of the scatter point.
    (\textbf{Right}) Secondary structure distribution. For each sample the proportion of residues as part of an alpha helix or beta strand is measured giving an xy scatter point coordinate.
    }
    \label{fig:protpardelle_metrics}
\end{figure}

\begin{figure}[h]
    \centering
    \includegraphics[width=12cm]{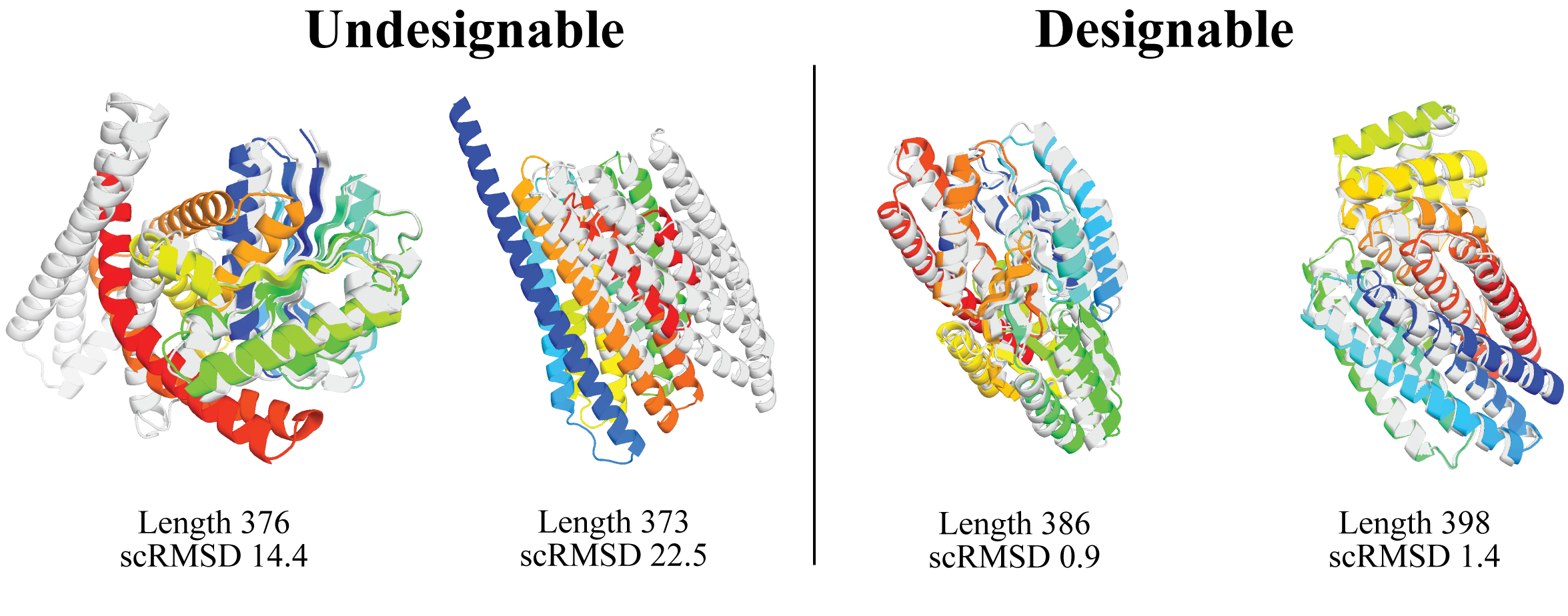}
    \vspace{-0.3cm}
    \caption{
    \textbf{\protmodel samples.}
    (\textbf{Left}) 2 undesignable \protmodel samples with the highest scRMSD from the benchmark.
    (\textbf{Right}) 2 designable \protmodel samples with the lowest scRMSD from the benchmark.
    }
    \label{fig:protpardelle_samples}
\end{figure}

\paragraph{Larger length range.}
Our results in \cref{sec:sec:codesign} only evaluated 4 lengths (70, 100, 200, 300) to match the benchmark in RFdiffusion.
However, other works have evaluated designability across all the lengths the method was trained on.
We follow Protpardelle \citep{chu2023all} to use \protmodel in generating 8 samples per length in the range $\{$50, 51, $\dots$, 400$\}$.
\cref{fig:protpardelle_metrics} shows the results in the same format as Figure 2B in Protpardelle.
We see \protmodel achieves near perfect designability up to around length 350 at which point designability starts to drop.
This is expected since \protmodel was only trained on lengths up to 384, but also demonstrates the ability to generalize beyond the lengths it was trained on.
We see \protmodel also achieves a desirable spread of secondary structure.
We show samples above length 370 with the highest and lowest Co-design 1 RMSD in \cref{fig:protpardelle_samples}.

\textbf{AlphaFold2 evaluation oracle.}
In \cref{sec:sec:codesign}, we presented a distillation technique of filtering out training examples that did not pass the designability criterion.
This also involved adding more proteins to the training set after sampling structures with \protmodel and filtering with designability using ProteinMPNN and ESMFold.
A potential risk of distillation is our model may overfit to  ESMFold since this model is used to filter training data and also for evaluation.
We show this is not the case in \cref{tab:oracle_results} by presenting the Co-design 1 results using AlphaFold2 (AF2) as an alternative oracle.
Our main results do not use AF2 since it is very slow and cumbersome to run and evaluate all our baselines.
We evaluated \protmodel with and without distillation to test \emph{if distillation with ESMFold provides an improvement regardless of the oracle used at evaluation}.
Overall designability numbers are lower with AF2; however, in both columns we see there is a two fold improvement regardless of the evaluation oracle.
This demonstrates distillation is not overfitting to the oracle used at evalution.

\begin{table}[H]
  \centering
  \caption{Co-design 1 designability results based on oracle.}
  \begin{tabular}{l|c|c}
    \toprule
     & Designability with ESMFold & Designability with  AF2 \\
    \midrule
    \protmodel w/o distillation & 0.41 & 0.38 \\
    \protmodel w/ distillation & 0.88 & 0.83 \\
    \hdashline
    Net improvement & \textbf{+0.47} & \textbf{+0.45} \\
    \bottomrule
  \end{tabular}
  \label{tab:oracle_results}
\end{table}

\subsection{Uniform Conditional Flow Ablation}
\label{sec:apdx_uniform_cogen}
\begin{figure}[h]
    \centering
    \includegraphics[width=15cm]{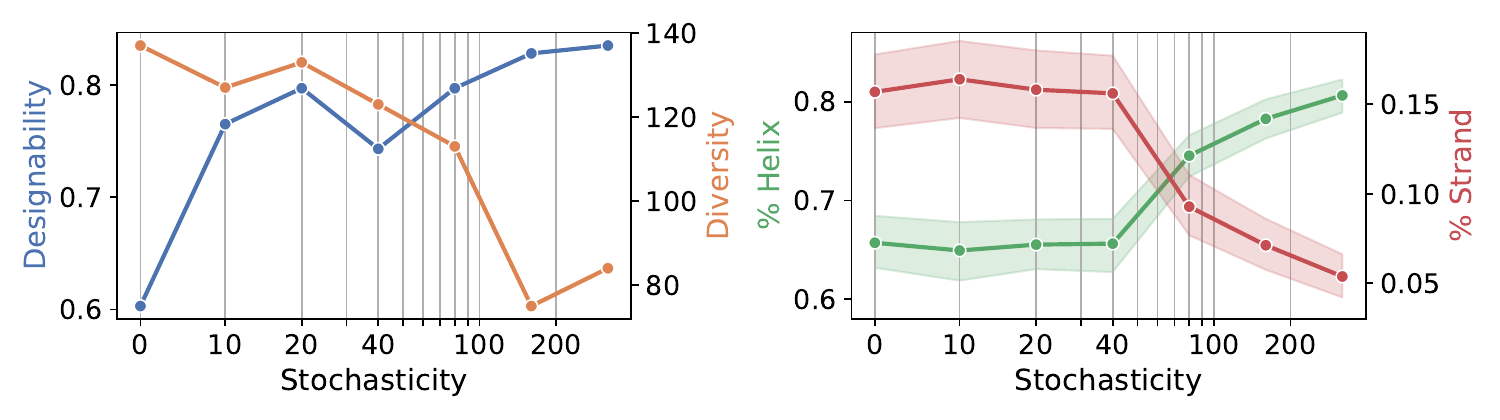}
    \vspace{-0.3cm}
    \caption{Sample metrics for \protmodel trained with the uniform interpolant on the discrete sequence modality. \textbf{(Left)} Co-Design 1 designability and diversity versus stochasticity level used when simulating the discrete CTMC. Higher is better for both designability and diversity. \textbf{(Right)} Average proportion of residues that are part of an alpha helix or beta strand versus the stochasticity level used to simulate the CTMC. Each point corresponds to the mean over 400 samples, 100 samples each for lengths 70, 100, 200, 300. Error bars show the standard error of the mean.}
    \label{fig:cogen_uniform_results}
\end{figure}

We ablate our use of the masking conditional flow and train a version of our \protmodel model using the uniform conditional flow ( see \cref{sec:apdx_uniform_example}).
We assessed the model's co-design performance by measuring the Co-Design 1 designability and diversity versus stochasticity level used at inference time.
We also measure the secondary structure composition of the generated samples versus stochasticity level.
Our results are given in \cref{fig:cogen_uniform_results}.
We find that in general, the Co-Design 1 designability increases with increasing stochasticity whilst the diversity as measured by the number of structural clusters decreases.
We can see the reason when examining the secondary structure statistics versus stochasticity. We see that at high stochasticity levels, the model heavily favours generating alpha helices at the expense of beta strands thus reducing the overall structural diversity. This will be due to interactions between errors in the model and the `churn' induced by extra stochasticity. It may be counter-intuitive that extra stochasticity reduces model diversity however we hypothesize that this is linked to the stochasticity inducing the model to converge on local optima in the likelihood landscape. When the model is generating a sample that it is confidence about, extra stochasticity will not shift it away from continuing down this simulation trajectory. However, when the model is exploring lower likelihood regions, the stochasticity can shift the models trajectory until it becomes stuck in a local optima again.

We find an overall worse trade-off between diversity and designability when using the uniform interpolant and so opt to use the masking interpolant in our main models.

\subsection{Forward and Inverse Folding Experiments}
The goal of our work is to develop the missing piece for a general-purpose framework for protein generation -- namely \method to integrate discrete data generation with a flow model.
We combined \method and FrameFlow to develop \protmodel where we have flexibility at inference time to choose which modality to provide and which to generate.
The task we focus on in this work is co-generation where the structure and sequence are jointly sampled rather than one after the other as done in prior works.
The other useful tasks in protein modeling are forward and inverse folding.
The two tasks are briefly described as follows; more in-depth description can be found in \citet{gao2020deep}.
\begin{enumerate}

    \item \textbf{Forward folding}: the task is to take the sequence as input and predicts the most thermodynamically plausible structure of the sequence.
    During evaluation, the ground truth structure is known, so we calculate the aligned structure erorr between the prediction and the ground truth.
    Several metrics exist to compute structure error, such as the Global Distance Test (GDT) commonly used in biophysical modeling \citep{pereira2021high}.
    We choose to use the aligned backbone RMSD error to keep our analysis simple and intuitive.
    The most well-used methods are AlphaFold2 \citep{jumper2021highly}, RosettaFold \citep{baek2021accurate}, and ESMFold \citep{lin2023evolutionary}.
    AlphaFold2 and RosettaFold rely on using evolutionary information which our model does not have access to (though can be extended to use).
    We compare against ESMFold, which does not use explicit evolutionary information, and due to its speed.

    \item \textbf{Inverse folding}: the task is to use the structure as input and predict the most likely sequence that would \emph{forward fold} into the structure.
    By this definition, the most sensible metric is the designability metric also used for co-generation.
    Specifically, the inverse folding model generates a sequence and we use ESMFold to predict the structure given this generated sequence.
    We call the self-consistency RMSD (scRMSD) as the RMSD between the structure predicted by ESMFold and the original input structure \citep{trippe2022diffusion}.
    The objective is to minimize scRMSD.
    The de facto method for inverse folding is ProteinMPNN \citep{dauparas2022robust}.
    Hence we compare against ProteinMPNN.

\end{enumerate}
It is important to emphasize that different deep learning models have been \emph{specifically} developed for forward and inverse folding, but no method can accomplish both tasks nor co-generate both sequence and structure.
\protmodel is unique in this regard to be able to perform co-generation, forward folding, and inverse folding.
We leave improving forward and inverse folding performance as a future work.
\textbf{Our aim is to demonstrate baseline performance of using a co-generation method to perform forward and inverse folding.}
We hope others can aid in advancing general purpose protein generative models.

\paragraph{Test set.}
ESMFold and ProteinMPNN have their own training and test sets which makes rigorous comparison impossible.
Re-training ESMFold and ProteinMPNN with the same training set of \protmodel is beyond the scope of our work.
Our results are a initial baseline of how \protmodel generally fares to specialized models on forward and inverse folding.

Our test set is based on a time-based split of the PDB.
We downloaded structures and sequences from the PDB that were released between 1st September 2021 and 28th December 2023. \emph{This time-based split ensures that none of the test set proteins are present in the training data for \protmodel, ProteinMPNN or ESMFold.}
We then select all single chain monomeric proteins with length between 50 and 400 inclusive.
We further filter out proteins that are more than 50\% coil residues and proteins that have a radius of gyration in the 96th percentile of the original dataset or above.
We also filter out structures that have missing residues.
We cluster proteins using the 30\% sequence identity MMSeqs2 clustering provided by RCSB.org.
We take a single protein from each cluster that matches our filtering criteria.
This gives us a test set of 449 proteins with minimum length 51 and maximum length 398.

\subsubsection{Forward Folding Results}
As described in \cref{tab:flexible_t_examples}, forward folding with \protmodel is performed by fixing the sequence time to $\tilde{t}=1$, providing the ground truth sequence as input, and running \method from $t=0$ to $t=1$.

\begin{figure}[h]
    \centering
    \includegraphics[width=\textwidth]{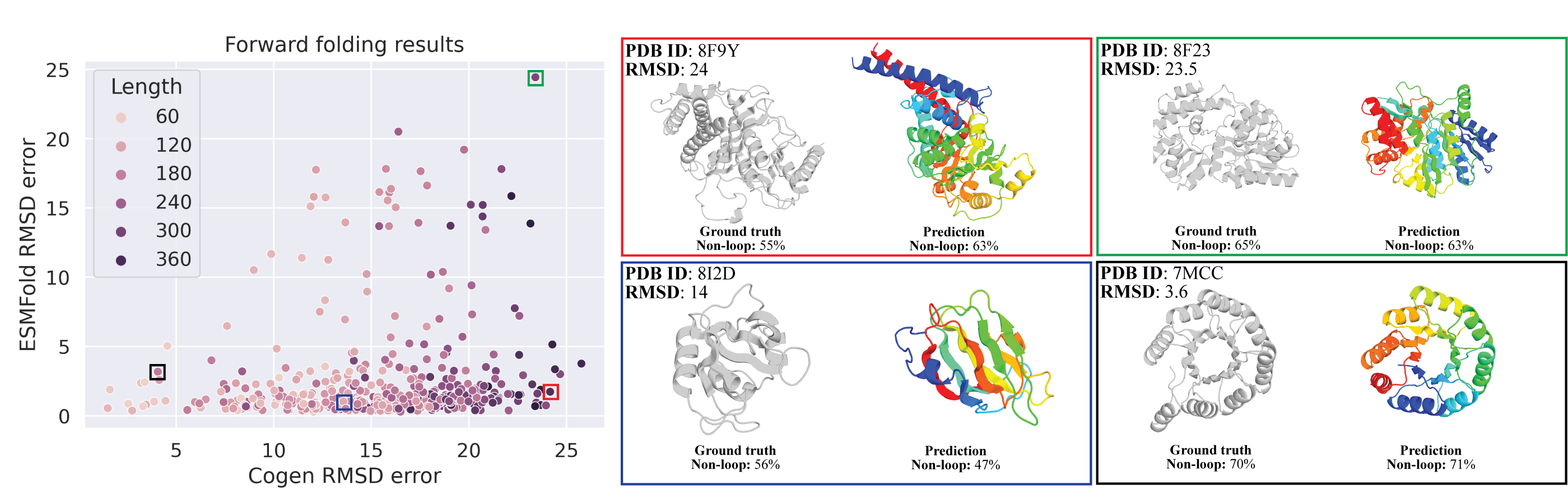}
    \caption{Forward folding RMSD metrics \textbf{(Left)} RMSD error between ground truth and predicted structures for \protmodel along the x-axis versus RMSD error for ESMFold on the y-axis. Each dot represents a protein in the test set. The shading of each point represents the length of the protein. \textbf{(Right)} Visualizations of ground truth structure (left) in grey and predicted structure (right) in color for 4 salient examples highlighted on the RMSD error plot. For each, the \protmodel RMSD error is given along with the proportion of non-loop residues for both the ground truth and prediction.}
    \label{fig:forward_folding}
\end{figure}

In \cref{fig:forward_folding} we examine the distribution of errors on our test set for both ESMFold and \protmodel. We find that generally \protmodel can have some success with proteins of smaller length but struggles with longer proteins. We investigate salient test examples from the plot to understand success and failure modes of our model. \protmodel is generally able to predict realistic protein structures with often similar secondary structure distributions as to the ground truth example seen by having similar proportions of non-loop residues between the ground truth and predicted structure. However, \protmodel often fails to predict the exact folded structure with high accuracy.

\begin{figure}[h]
    \centering
    \includegraphics[width=15cm]{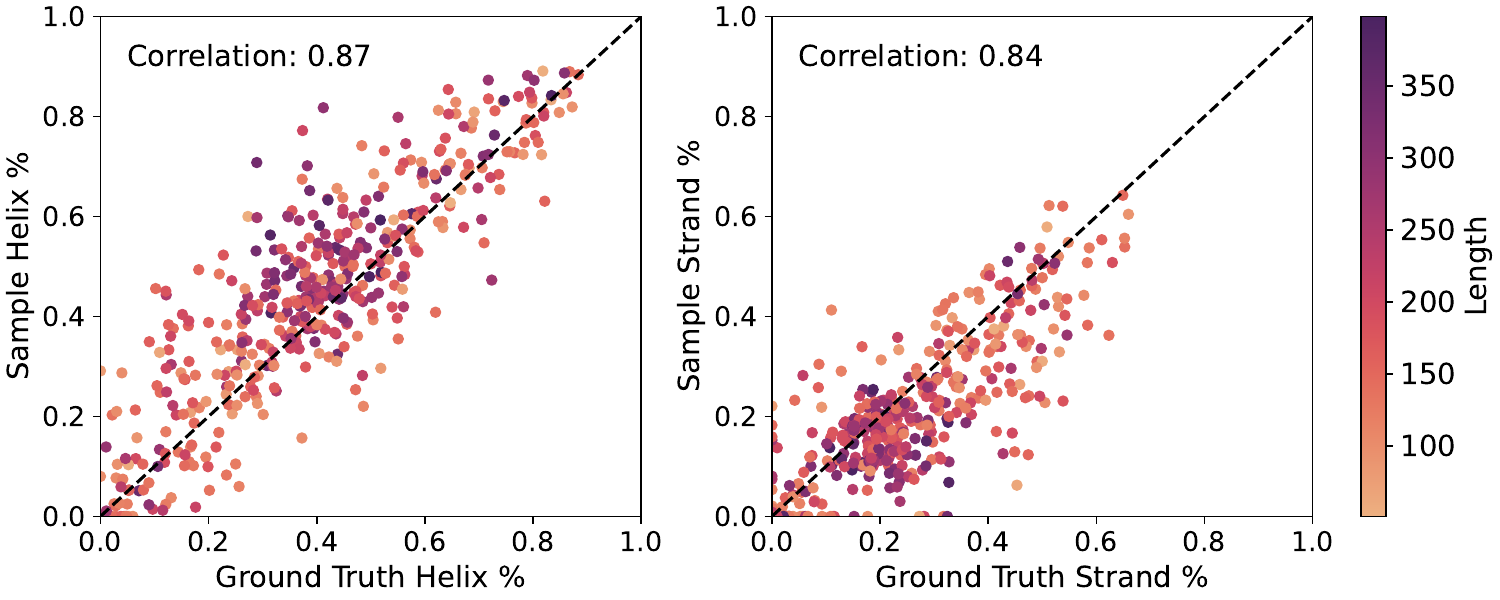}
    \caption{Proportion of residues that are part of secondary structure elements for both the \protmodel predicted structure and the ground truther structure. We plot the ground truth proportion of residues in a secondary structure element along the x-axis and the proportion of residues in the predicted structure on the y-axis. The left plot examines alpha helices whilst the right plot examines strand elements. Each scatter point represents a test set protein with the colour indicating the length. The perfect result of exactly matching proportion with the ground truth is plotted as a dashed diagonal line. We also report the correlation coefficient for each plot.}
    \label{fig:forward_secondary}
\end{figure}

We quantify the secondary structure prediction accuracy in \cref{fig:forward_secondary} by comparing the secondary structure present in the ground truth versus the structure predicted by \protmodel. We find good correlation between the predicted secondary structure and ground truth highlighting that \protmodel is able to use information present within the given sequence to generate structures.

\subsubsection{Inverse Folding Results}

Similarly to forward folding, inverse folding with \protmodel is performed by fixing the structure time to $t =1$, providing the ground truth structure and running the sequence flow from $\tilde{t}=0$ to $\tilde{t}=1$.

\begin{figure}
    \centering
    \includegraphics[width=10cm]{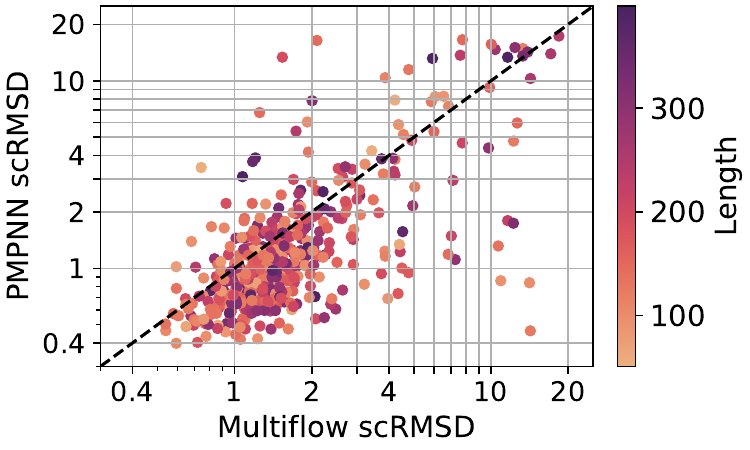}
    \caption{\protmodel scRMSD versus PMPNN scRMSD on our test set. Each scatter point represents a protein with the shading giving the length. We also plot the dividing line of equal scRMSD for the two models for ease of comparison.}
    \label{fig:inverse_folding_rmsds}
\end{figure}

We plot our results in \cref{fig:inverse_folding_rmsds}. We find that \protmodel performs competitively with PMPNN across a wide range of protein lengths with PMPNN achieving slightly lower scRMSD values on average. For both models, scRMSD tends to cluster around 1 to 2 scRMSD. There are test proteins for which PMPNN achieves a lower scRMSD and also cases protein for which \protmodel acheives the lower scRMSD.

\end{document}